\documentclass[10pt, a4paper]{article}

\usepackage[final]{lrec2026} 
\usepackage{graphicx}
\usepackage{multirow}
\usepackage{booktabs}
\usepackage[table,dvipsnames]{xcolor}
\usepackage{colortbl}
\usepackage{subcaption}
\usepackage{adjustbox}
\usepackage{enumitem}
\usepackage[most]{tcolorbox}
\usepackage{listings}
\newcommand{\uzh}{1}
\newcommand{\unibe}{2}

\title{Taming CATS: Controllable Automatic Text Simplification through Instruction Fine-Tuning with Control Tokens}

\name{$\textbf{Hanna Hubarava}^{\uzh, \unibe}$ \quad  $\text{Yingqiang Gao*}^{\uzh}$} 

\address{$^{\uzh}$Department of Computational Linguistics, University of Zurich, Switzerland \\
$^{\unibe}$Department of Clinical Research, University of Bern, Switzerland \\
hanna.hubarava@unibe.ch, yingqiang.gao@cl.uzh.ch}

\abstract{
Controllable Automatic Text Simplification (CATS) produces user-tailored outputs, yet controllability is often treated as a decoding problem and evaluated with metrics that are not reflective to the measure of control.
We observe that controllability in ATS is significantly constrained by \emph{data and evaluation}. To this end, we introduce a domain-agnostic CATS framework based on instruction fine-tuning with discrete control tokens, steering open-source models to target readability levels and compression rates.
Across three model families with different model sizes (Llama, Mistral, Qwen; 1--14B) and four domains (medicine, public administration, news, encyclopedic text), we find that smaller models (1--3B) can be competitive, but reliable controllability strongly depends on whether the training data encodes sufficient variation in the target attribute.
Readability control (FKGL, ARI, Dale-Chall) is learned consistently, whereas compression control underperforms due to limited signal variability in the existing corpora.
We further show that standard simplification and similarity metrics are insufficient for measuring control, motivating error-based measures for target-output alignment.
Finally, our sampling and stratification experiments demonstrate that naive splits can introduce distributional mismatch that undermines both training and evaluation.
\\ \newline
\Keywords{Automatic Text Simplification, Controllable Text Generation, Instruction Fine-Tuning.}
}

\begin{document}



\makeatletter
\maketitleabstract
\begingroup
\renewcommand\thefootnote{}
\renewcommand\@makefnmark{}
\long\def\@makefntext#1{%
  \noindent\parbox{\dimexpr\columnwidth\relax}{#1}%
}
\footnotetext{%
*Corresponding author.\hspace{1em}%
\includegraphics[height=0.32cm]{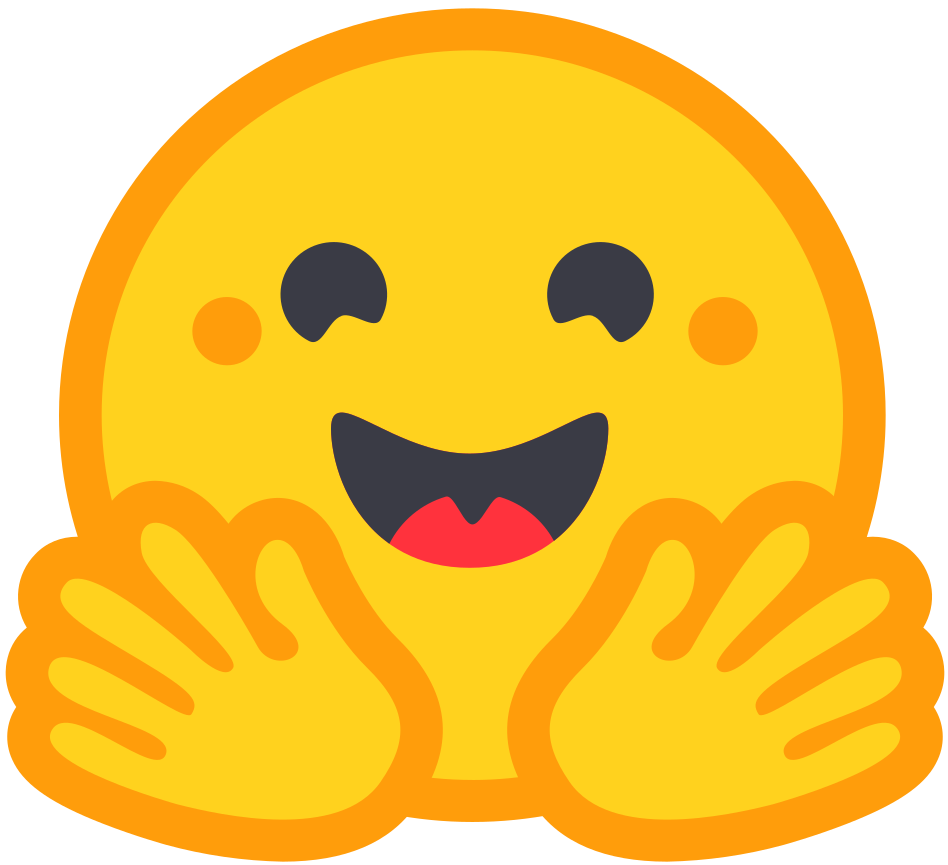}\,
\href{https://huggingface.co/datasets/shtosti/CATS}{Data}\quad
\includegraphics[height=0.32cm]{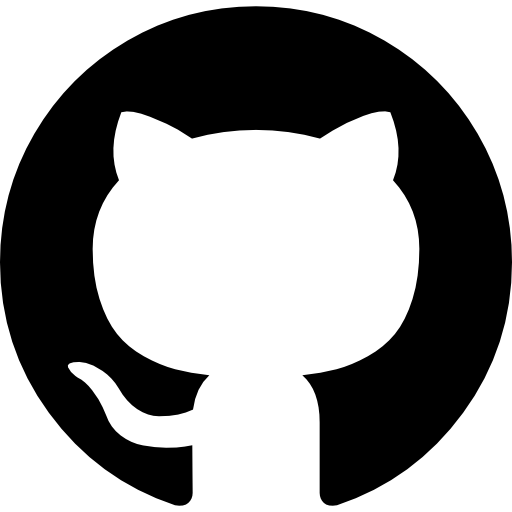}\,
\href{https://github.com/shtosti/taming-CATS}{Code}
}
\endgroup
\makeatother

\section{Introduction} \label{sec:intro}
Automatic Text Simplification (ATS) aims to reduce linguistic complexity while preserving meaning in order to enable accessibility of information for diverse purposes and audiences \citep{Shardlow2014ASO, grabar-saggion-2022-evaluation, espinosa-zaragoza-etal-2023-review}. Recent advances in large language models (LLMs) have rekindled interest in \emph{controllable} simplification \citep{Kew2023BLESSBL, tran-etal-2025-readctrl}, where systems are expected to adapt outputs to user-specified complexity levels rather than produce a static simplified variant. Controllability in ATS tends to be treated as a decoding problem \citep{martin_controllable_2020}, with a focus on conditioning mechanisms such as model prompting or inference configurations, while largely assuming that datasets, splits, and evaluation metrics reflect the intended notion of control.
Yet, ATS datasets vary widely in how complexity is encoded, many exhibit minimal variation along key attributes (e.g., compression). Training LLMs on datasets of limited variation \citep{vasquez-rodriguez_investigating_2021} would thus fail to learn distinguish fine-grained controllability needs.

In this work, we present our approach to CATS: an instruction fine-tuning (IFT) framework using discrete control tokens with open-source decoder-only LLMs. We compared control effectiveness across five readability attributes (FKGL \citep{Kincaid_Navy}; ARI \citep{ARI_smith}; Dale-Chall \citep{Dale_chall_1948, Chall_1995_ReadabilityR} and two compression levels (word and character), on four domain-specific datasets (\textsc{Med-EASi} \citep{Basu2023MedEASiFA}; \textsc{SimPA} \citep{scarton-etal-2018-simpa}; \textsc{WikiLarge} \citep{zhang-lapata-2017-sentence}; \textsc{Newsela} \citep{xu_problems_2015}), by testing various model families (Llama \citep{dubey2024llama}; Mistral \citep{jiang2023mistral}; Qwen \citep{yang2025qwen3}) and sizes (1-14B). 

We argue that \emph{data and evaluation} are equally important for CATS: our results show that LLMs can learn to target absolute readability levels through fine-tuning, but only when the training data contains sufficient and well-distributed learning signal. We also conducted extensive data experiments on sampling and stratification, and observed that use of native dataset splits or naively randomized partitioning can magnify distributional mismatch between training and evaluation sets. We further emphasize that metrics which take into account the error between target (reference simplification) and prediction (model output) values are indispensable for measuring controllability, since traditional simplification and similarity metrics are agnostic to target complexity deviations. With these findings, we show that for effective CATS solutions, robustness, reproducibility, and evaluation design are as critical as architectural choices.

\begin{figure}[t]
    \centering

    \begin{subfigure}[c]{0.48\columnwidth}
        \centering
        \includegraphics[width=\linewidth]{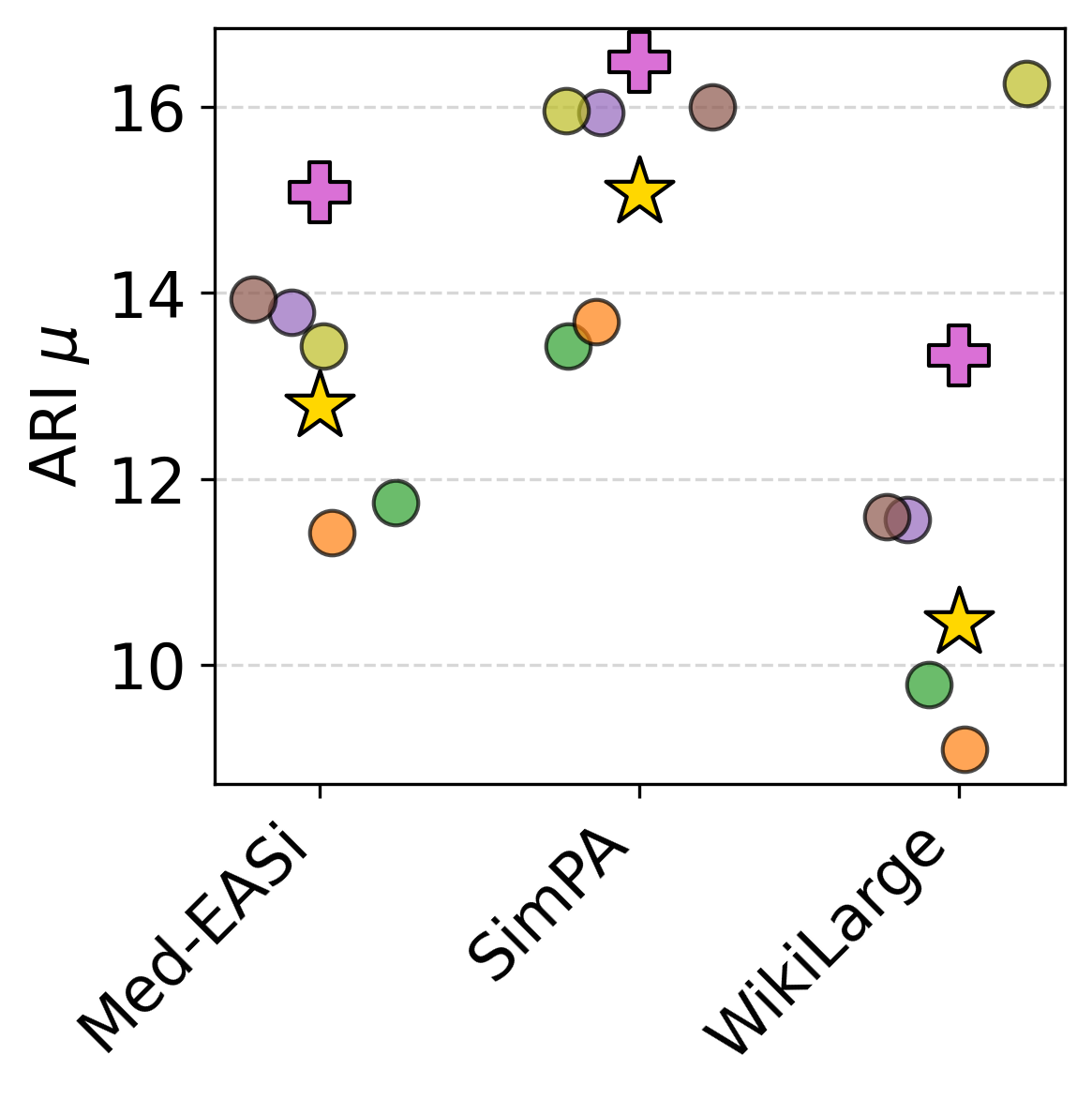}
    \end{subfigure}
    \hfill
    \begin{subfigure}[c]{0.48\columnwidth}
        \centering
        \includegraphics[width=\linewidth]{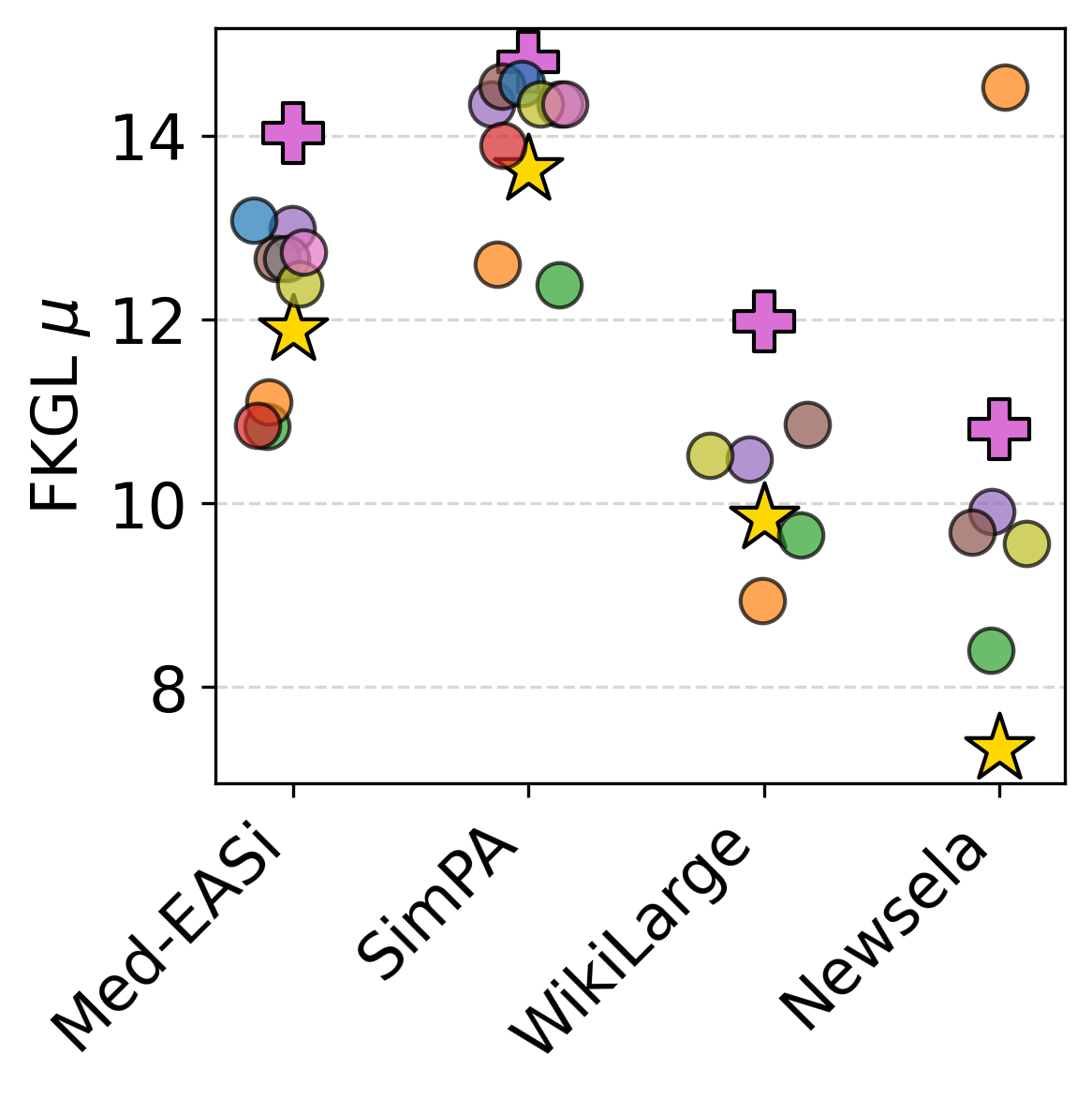}
    \end{subfigure}


    \begin{subfigure}[c]{0.48\columnwidth}
        \centering
        \includegraphics[width=\linewidth]{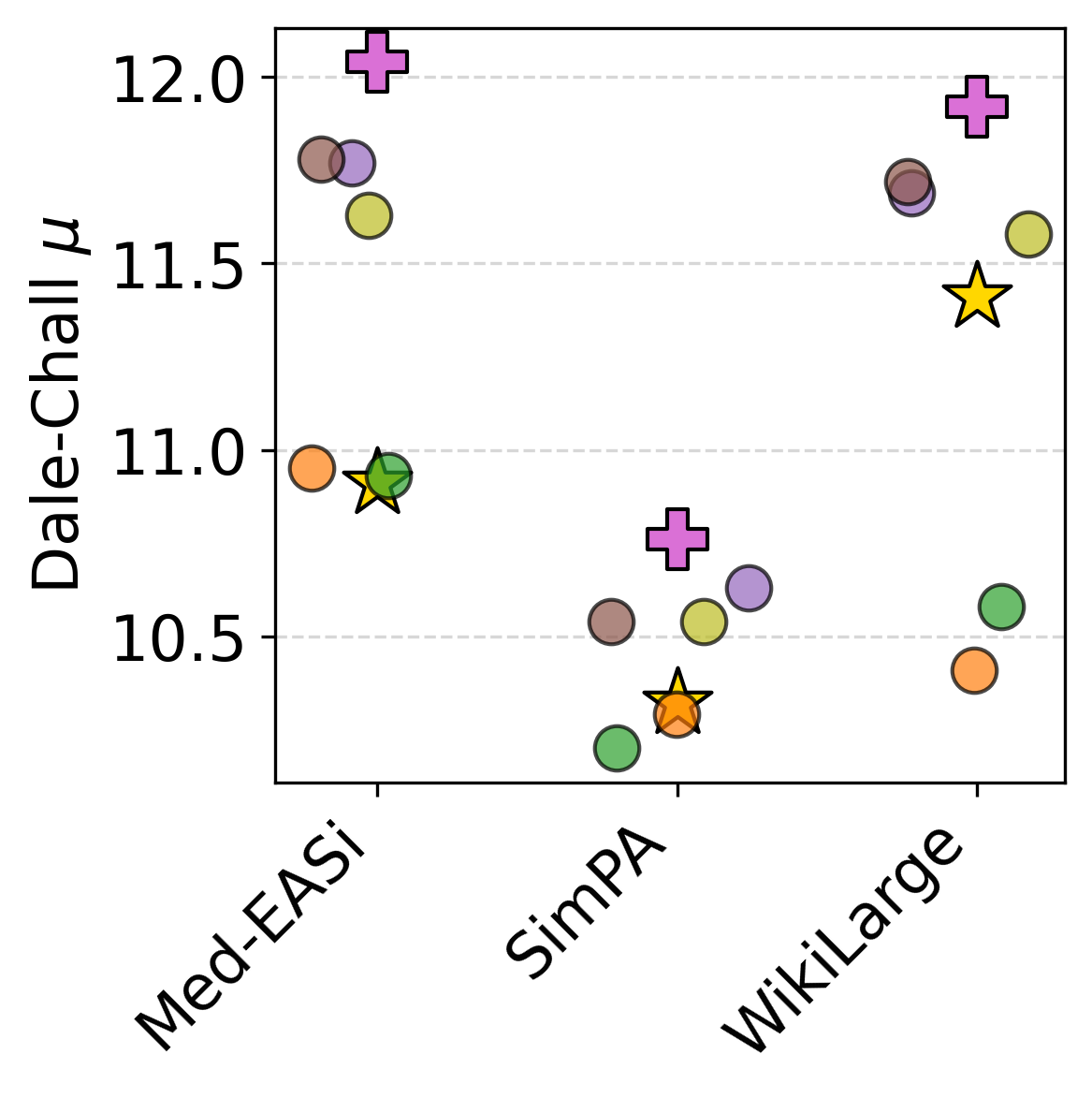}
    \end{subfigure}
    \hfill
    \begin{subfigure}[c]{0.48\columnwidth}
        \centering
        \includegraphics[width=\linewidth]{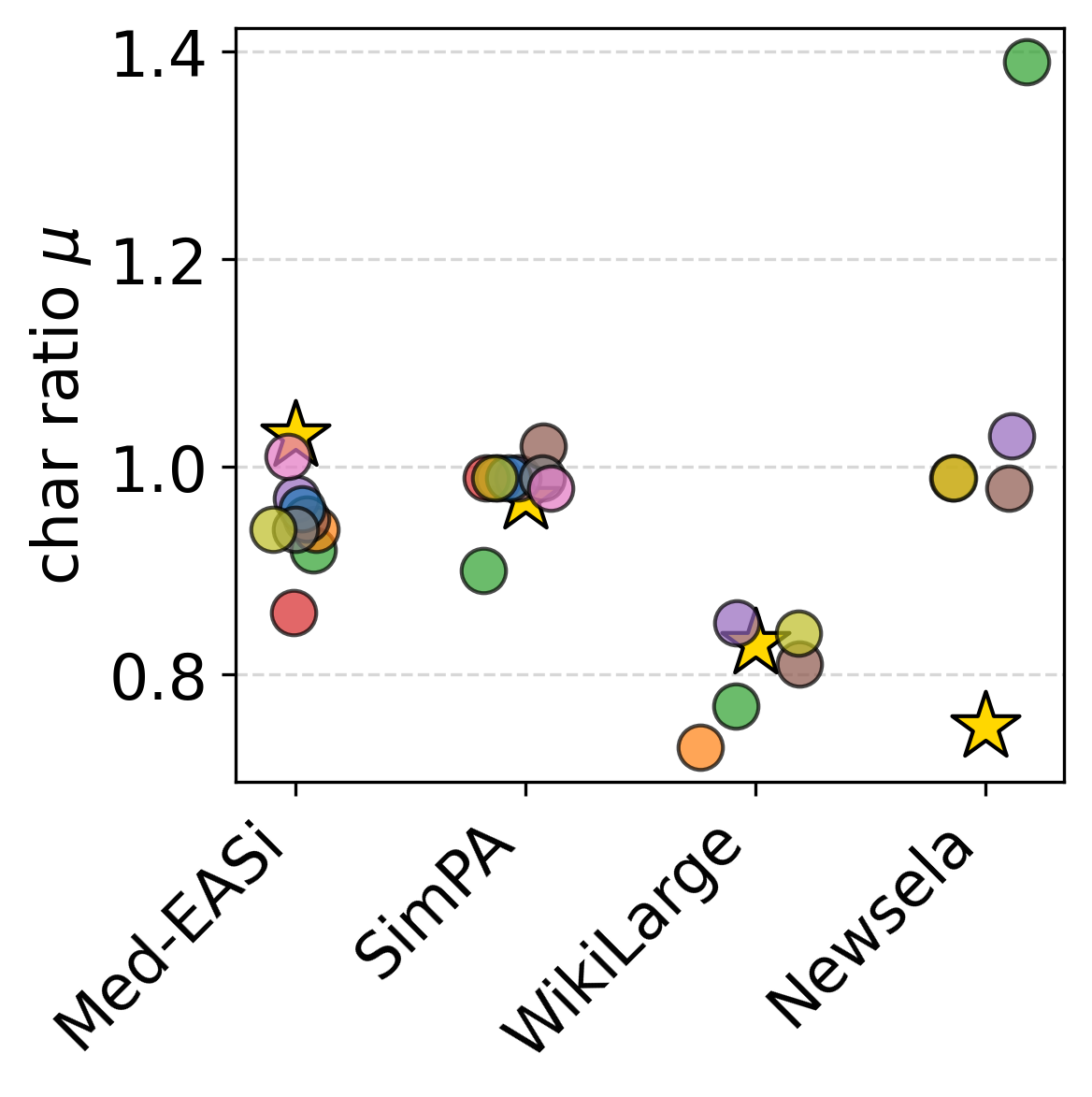}
    \end{subfigure}


    \begin{subfigure}[c]{0.48\columnwidth}
        \centering
        \includegraphics[width=\linewidth]{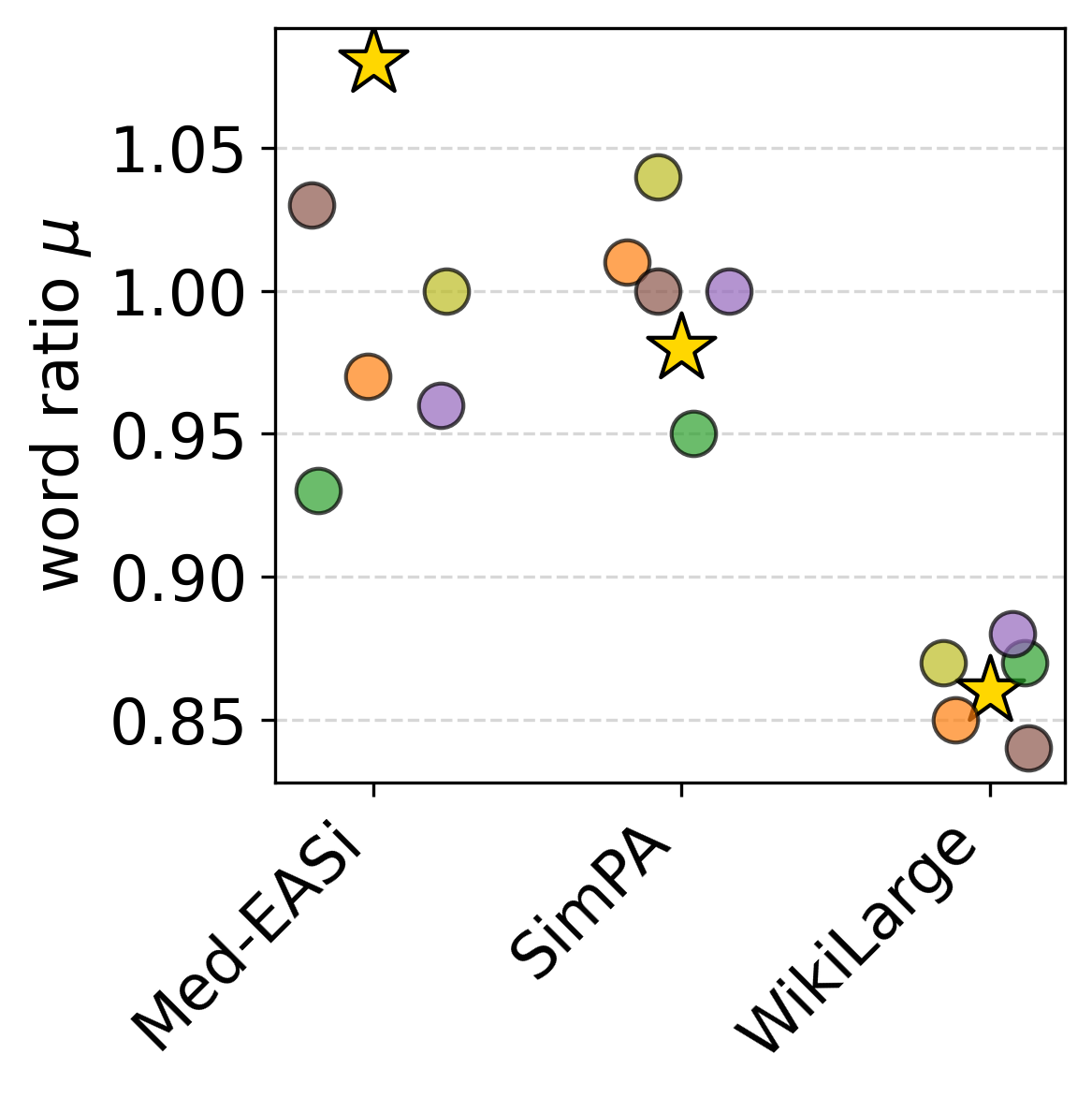}
    \end{subfigure}
    \hfill
    \begin{subfigure}[c]{0.48\columnwidth}
        \centering
        \includegraphics[width=\linewidth]{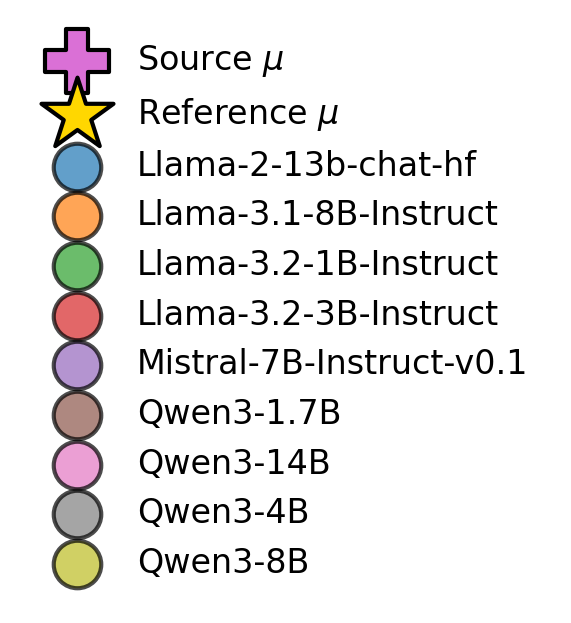}
    \end{subfigure}

    \caption{Mean control-attribute values across models and datasets. Fine-tuned models tend to generate text that is simpler than the source but more complex than the target. Sentence-aligned datasets (\textsc{Med-EASi, SimPA, WikiLarge}) show very little compression signal (around 1.0).}
    \label{fig:mean-ctrl-attr}

\end{figure}

\section{Related Work} \label{sec:related-work}
ATS is a specialized form of controllable text generation, traditionally modeled as monolingual machine translation \citep{wubben_sentence_2012, simplification-as-translation, sheang_controllable_2021}. While it partly shares goals with text summarization \citep{alva-manchego_data-driven_2020}, ATS is distinct in its focus on readability and complexity reduction; simplifications may actually increase text length through explanatory paraphrasing or addition of cohesive markers \citep{alva-manchego_data-driven_2020}. Over the years, ATS research has transitioned from rule- and dictionary-based as well as statistical systems \citep{Chandrasekar1997AutomaticIO, espinosa-zaragoza-etal-2023-review} to data-driven neural approaches \citep{wang2016experimental, nisioi-etal-2017-exploring, zhang-lapata-2017-sentence}, but the scarcity of high-quality parallel corpora remains a key bottleneck \citep{vasquez-rodriguez_investigating_2021, agrawal_text_2024}.

Controllability in the ATS is guided by specific transformation attributes, which can be indicated in the instruction prompt or encoded in the control token. ACCESS framework \citep{martin_controllable_2020} pioneered conditioning models on lexical complexity, length, and syntactic markers. Subsequent work has expanded this to proficiency levels (e.g., CEFR tokens \citep{spring_exploring_2021}) and the mitigation of ``copying behavior'' through explicit labels \citep{sheang_controllable_2021}. To bypass data scarcity, unsupervised approaches like MUSS \citep{martin_muss_2022} leverage mined paraphrases and unsupervised pre-training. Recently, controlled decoding methods such as FUDGE \citep{yang-klein-2021-fudge, kew_target-level_2022} have been combined with paraphrase models to nudge outputs toward target complexity levels without requiring massive in-domain parallel data.

Faced with the inherent subjectivity of simplification quality evaluation \citep{grabar-saggion-2022-evaluation}, ATS evaluation typically considers adequacy (meaning preservation), fluency, and simplicity. Early work relied on machine translation metrics such as BLEU \citep{papineni-etal-2002-bleu}, which are insensitive to structural transformations. SARI \citep{xu_optimizing_2016_sari}, now a standard metric for simplification, explicitly models lexical edit operations but has been shown to correlate weakly with human judgments in cases involving structural changes such as sentence splitting or merging \citep{alva-manchego_data-driven_2020}. Learned metrics such as BERTScore \citep{Zhang2019BERTScoreET} improve the assessment of semantic adequacy \citep{alva-manchego_suitability_2021}, but remain general-purpose and do not explicitly account for simplicity. More recently, LENS \citep{maddela-etal-2023-lens} has been introduced as a simplification-specific learned metric that holistically models human judgments of simplification quality, reflecting adequacy, fluency, and simplicity.

\section{Methods} \label{sec:methods}

\subsection{Control Attributes}
We measured controllability in terms of model's ability to generate a text simplification of a specific control attribute value.  We operated with five such numerical attributes: three readability attributes include FKGL \citep{Kincaid_Navy}, ARI \citep{ARI_smith}, Dale-Chall \cite{Dale_chall_1948, Chall_1995_ReadabilityR}, and two compression ratios in terms of character and word count.

\subsection{Datasets} \label{subsec:methods-datasets}

\newcommand{\histpairheight}{0.135\textheight}

\begin{figure}[t]
  \centering

  \begin{minipage}[t]{0.4\columnwidth}
    \centering
    \textsc{SimPA}\\[2pt]

    \includegraphics[width=\linewidth,height=\histpairheight,keepaspectratio]{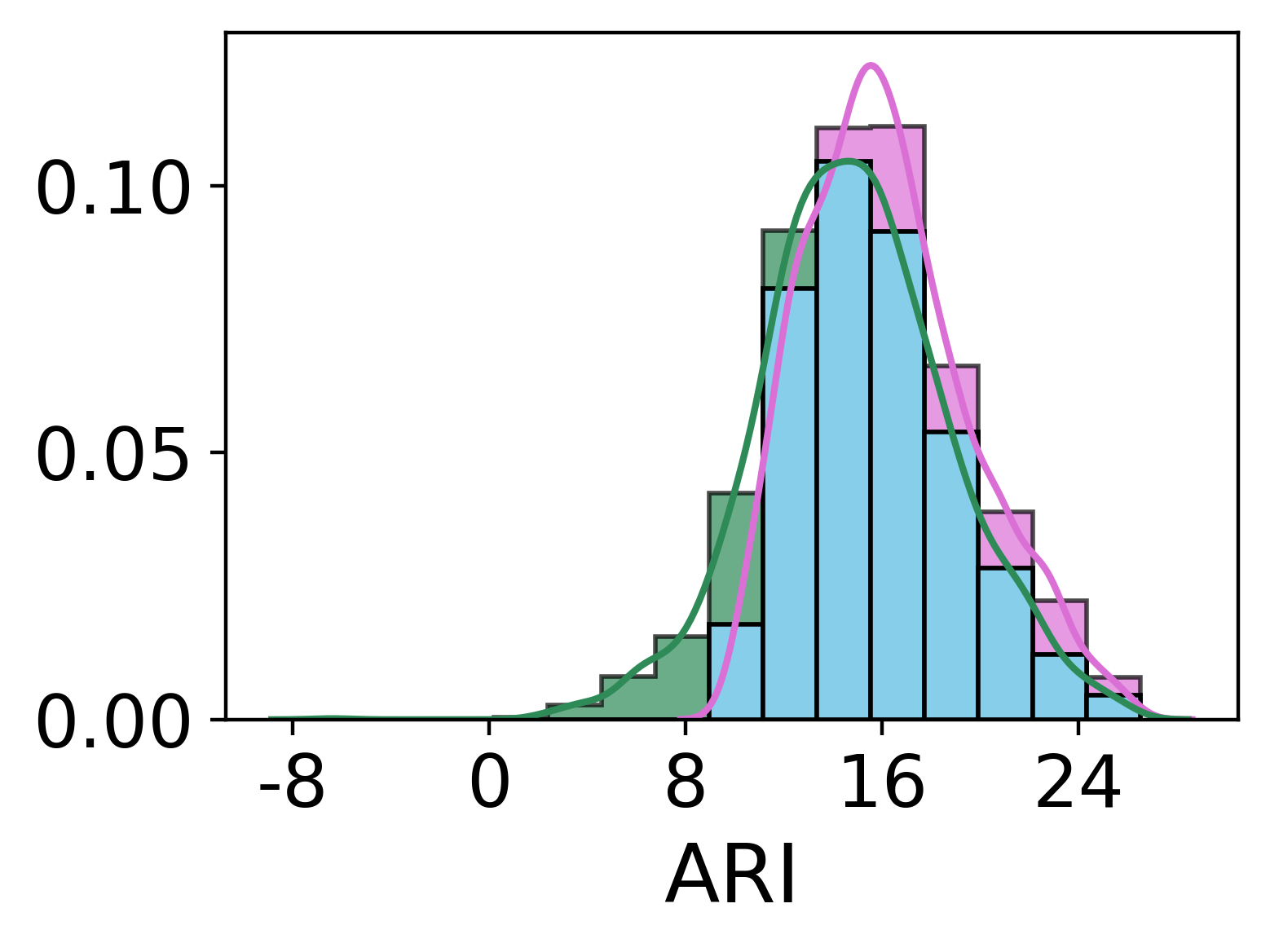}\par

    \vspace{2pt}
    \includegraphics[width=\linewidth,height=\histpairheight,keepaspectratio]{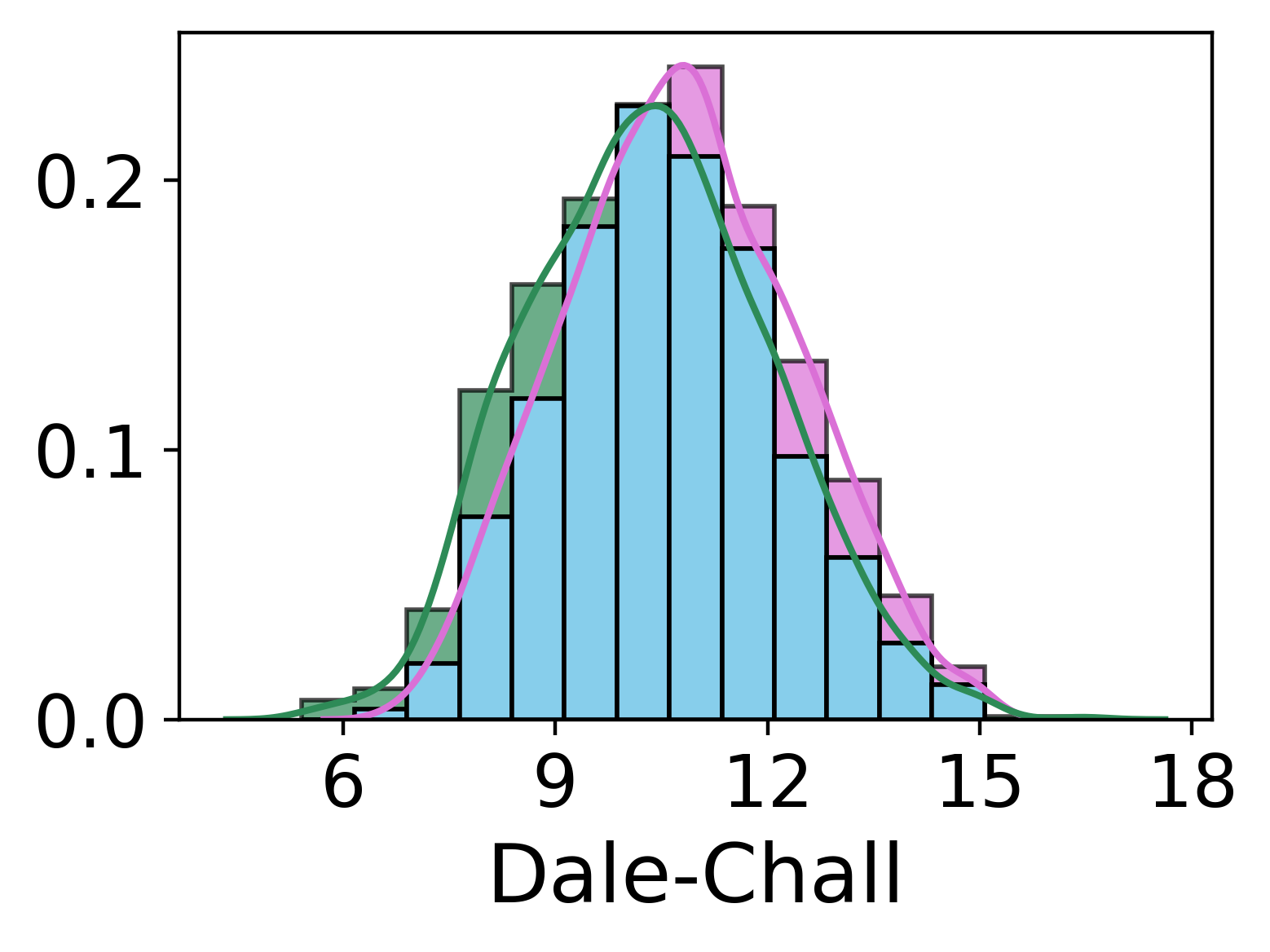}\par

    \vspace{2pt}
    \includegraphics[width=\linewidth,height=\histpairheight,keepaspectratio]{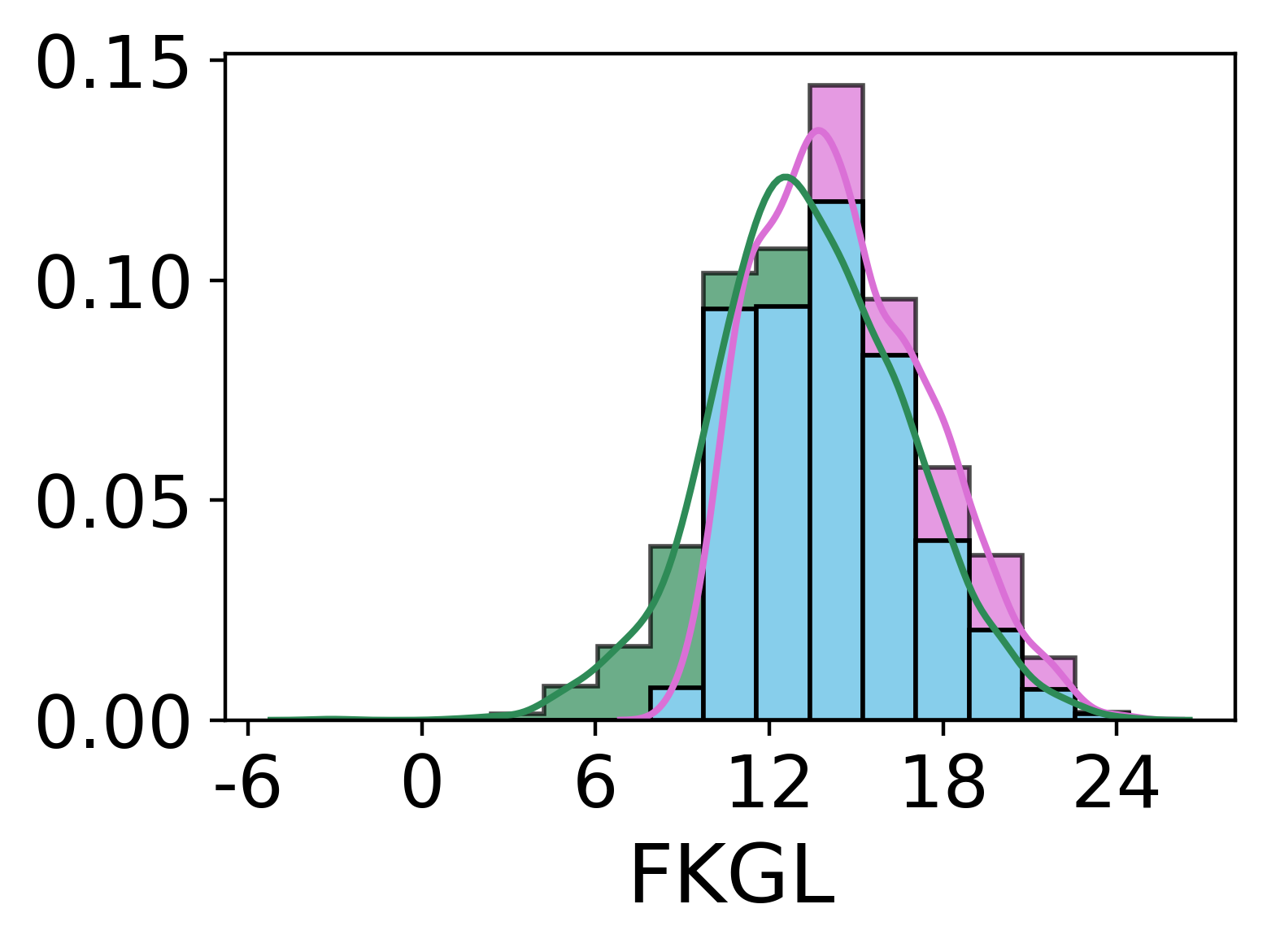}\par

    \vspace{2pt}
    \includegraphics[width=\linewidth,height=\histpairheight,keepaspectratio]{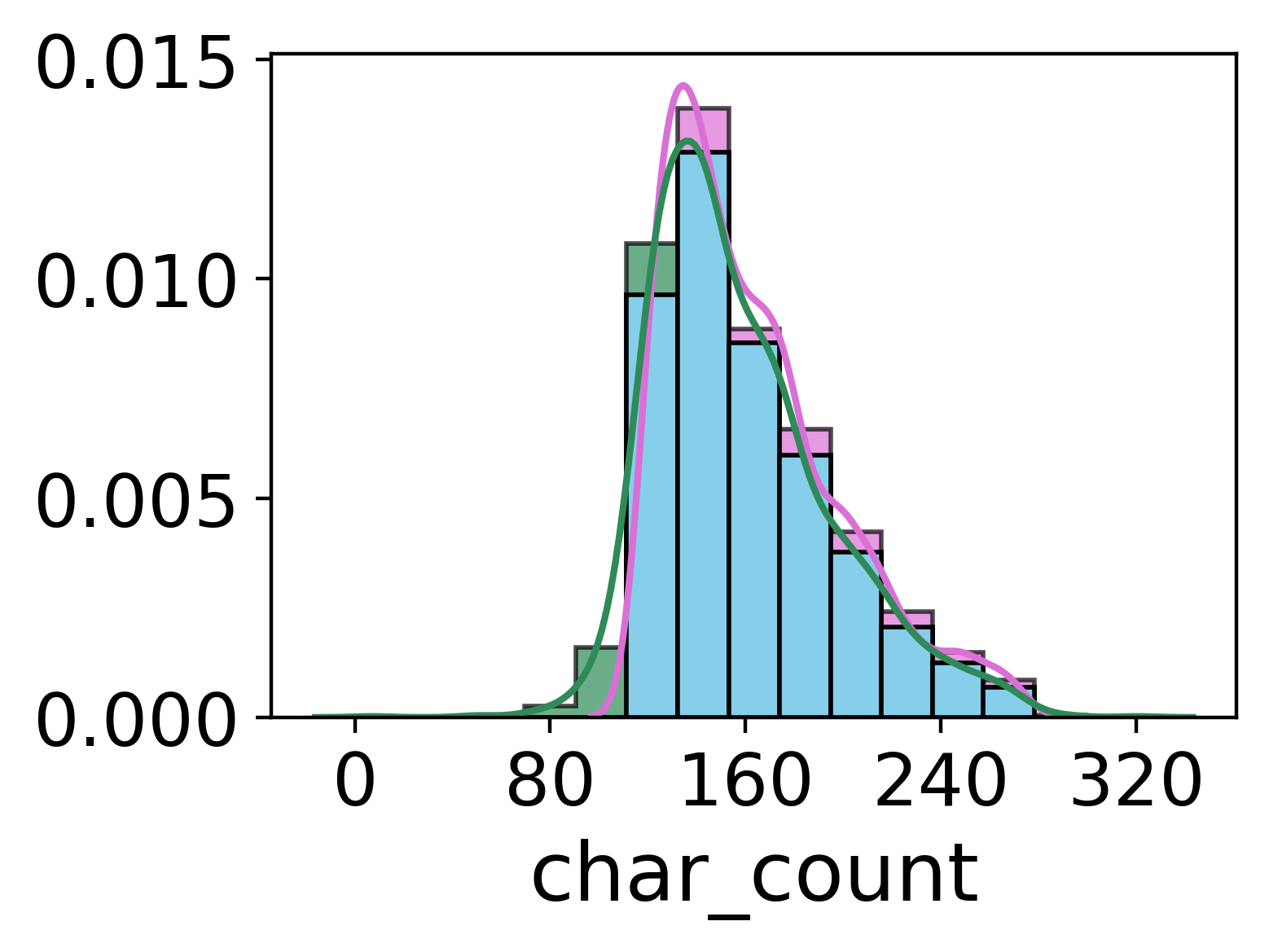}\par

    \vspace{2pt}
    \includegraphics[width=\linewidth,height=\histpairheight,keepaspectratio]{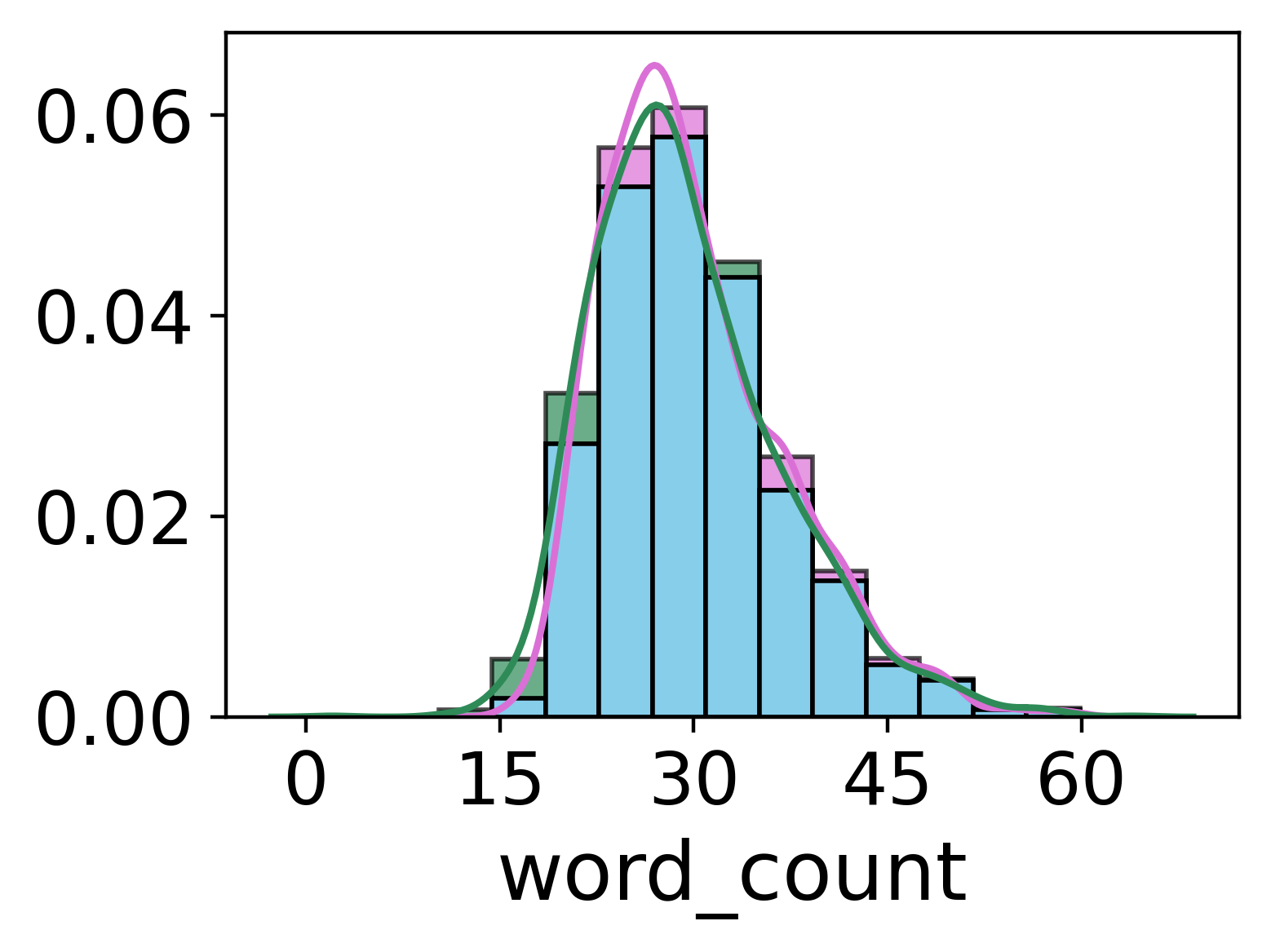}\par
  \end{minipage}
  \hfill
  \begin{minipage}[t]{0.4\columnwidth}
    \centering
    \textsc{Newsela}\\[2pt]

    \includegraphics[width=\linewidth,height=\histpairheight,keepaspectratio]{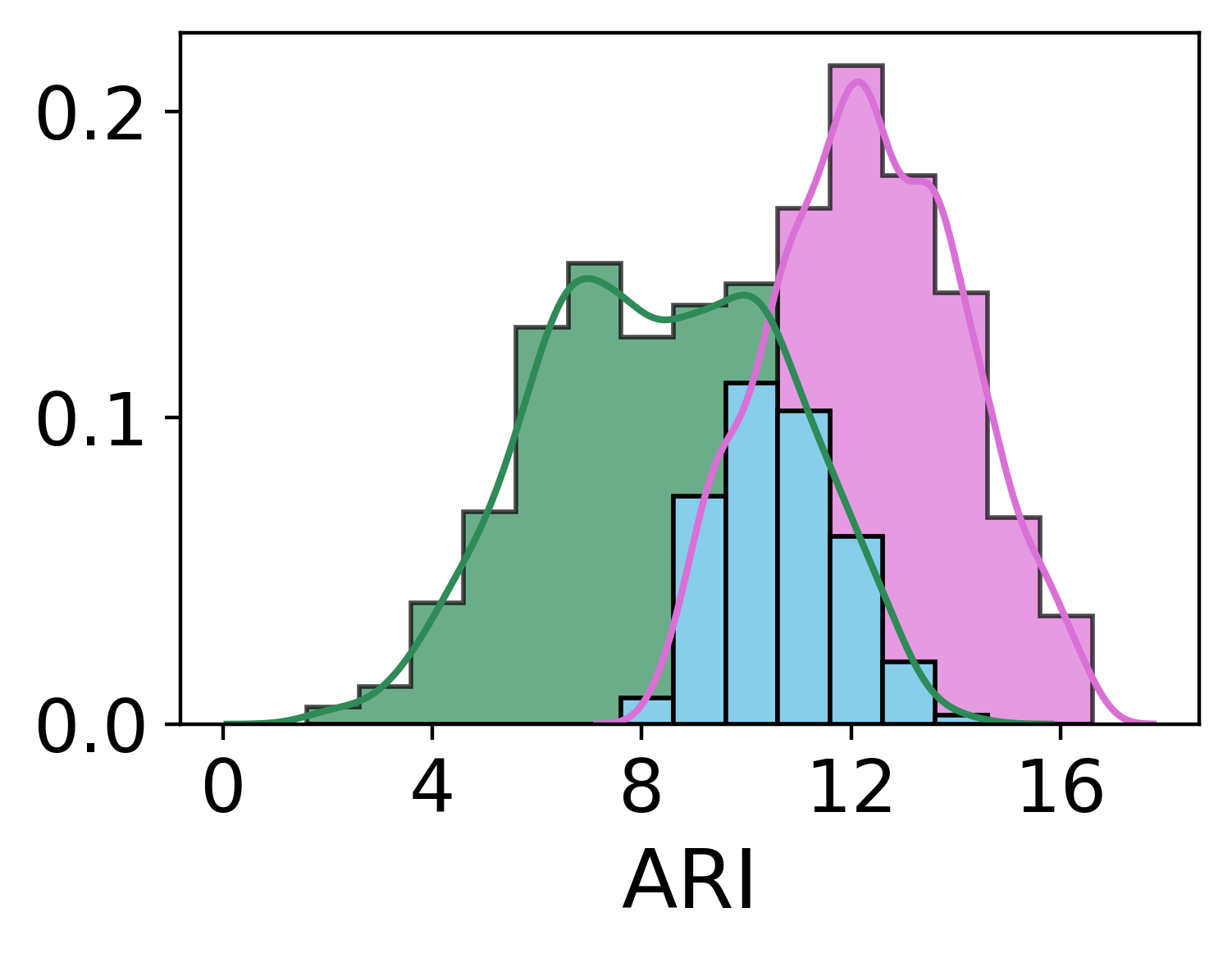}\par
    \includegraphics[width=\linewidth,height=\histpairheight,keepaspectratio]{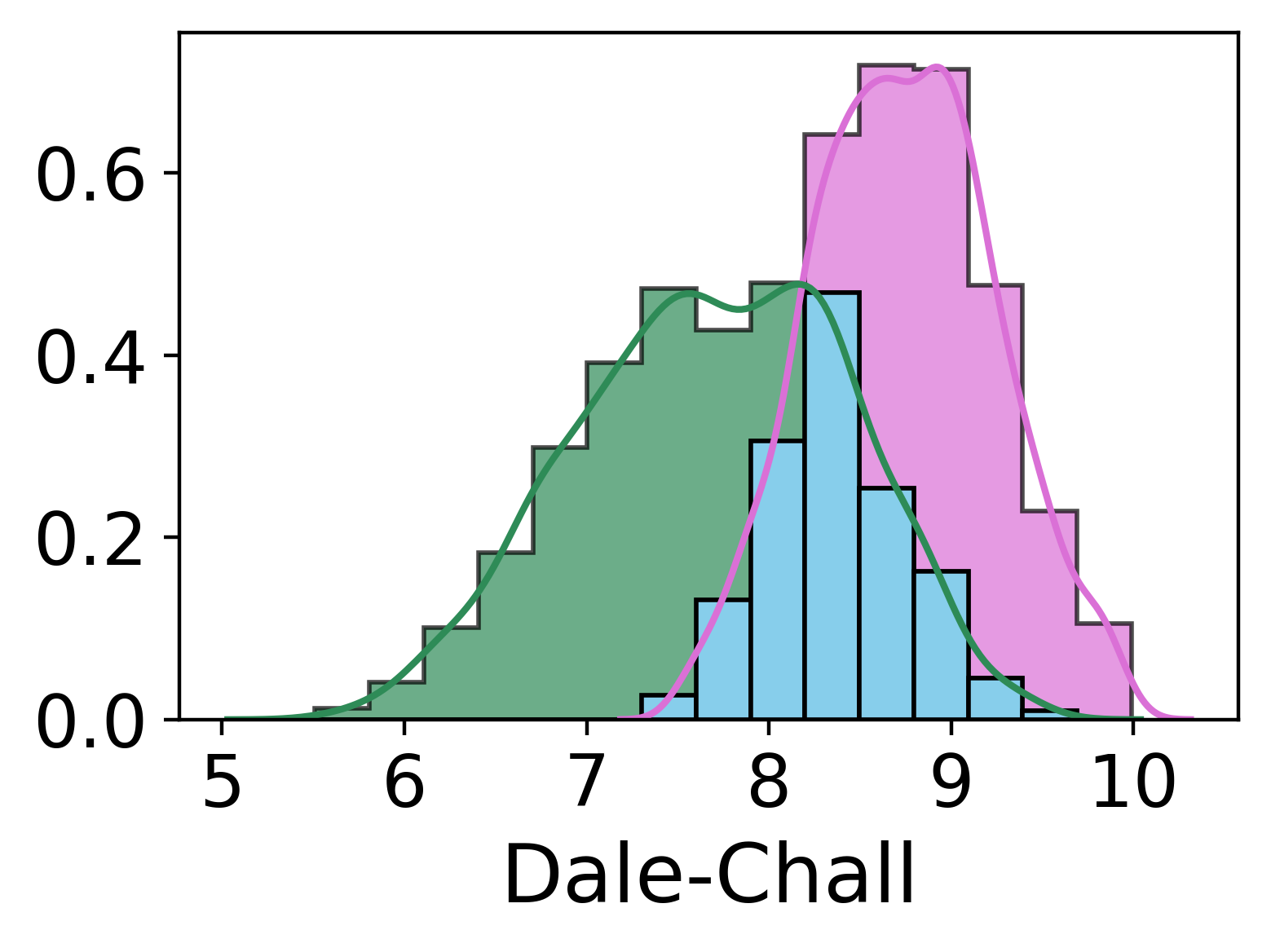}\par
    \includegraphics[width=\linewidth,height=\histpairheight,keepaspectratio]{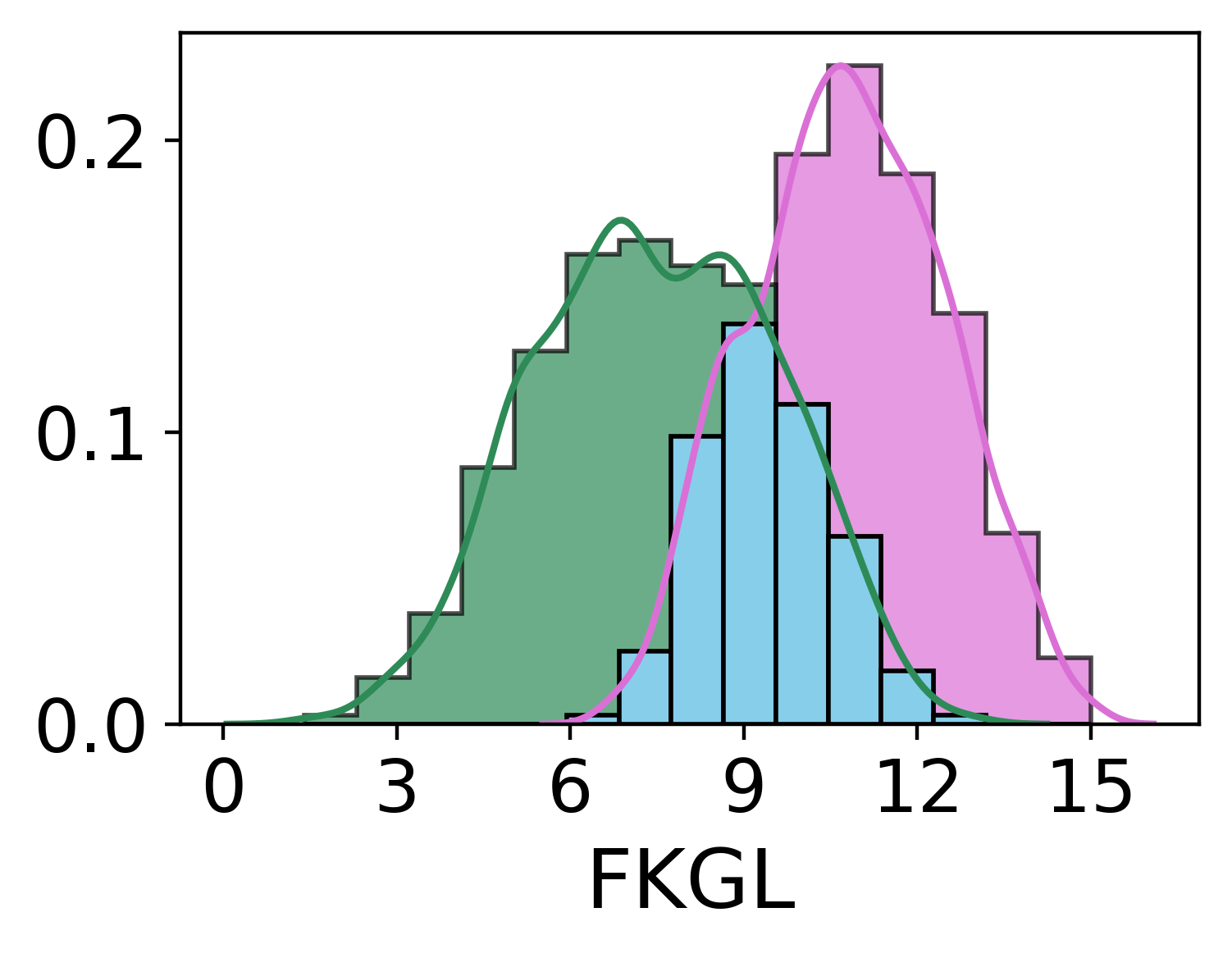}\par
    \includegraphics[width=\linewidth,height=\histpairheight,keepaspectratio]{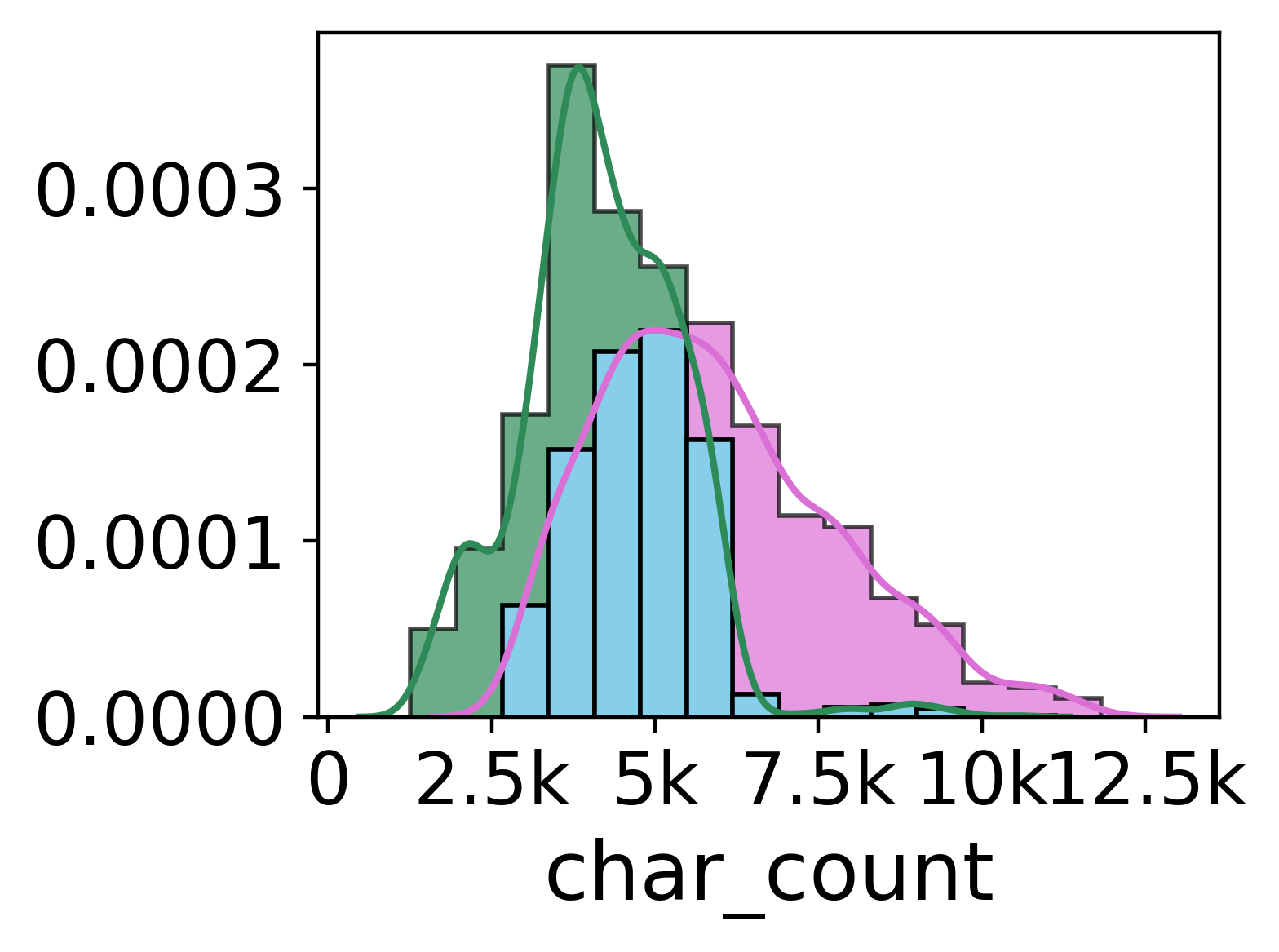}\par
    \includegraphics[width=\linewidth,height=\histpairheight,keepaspectratio]{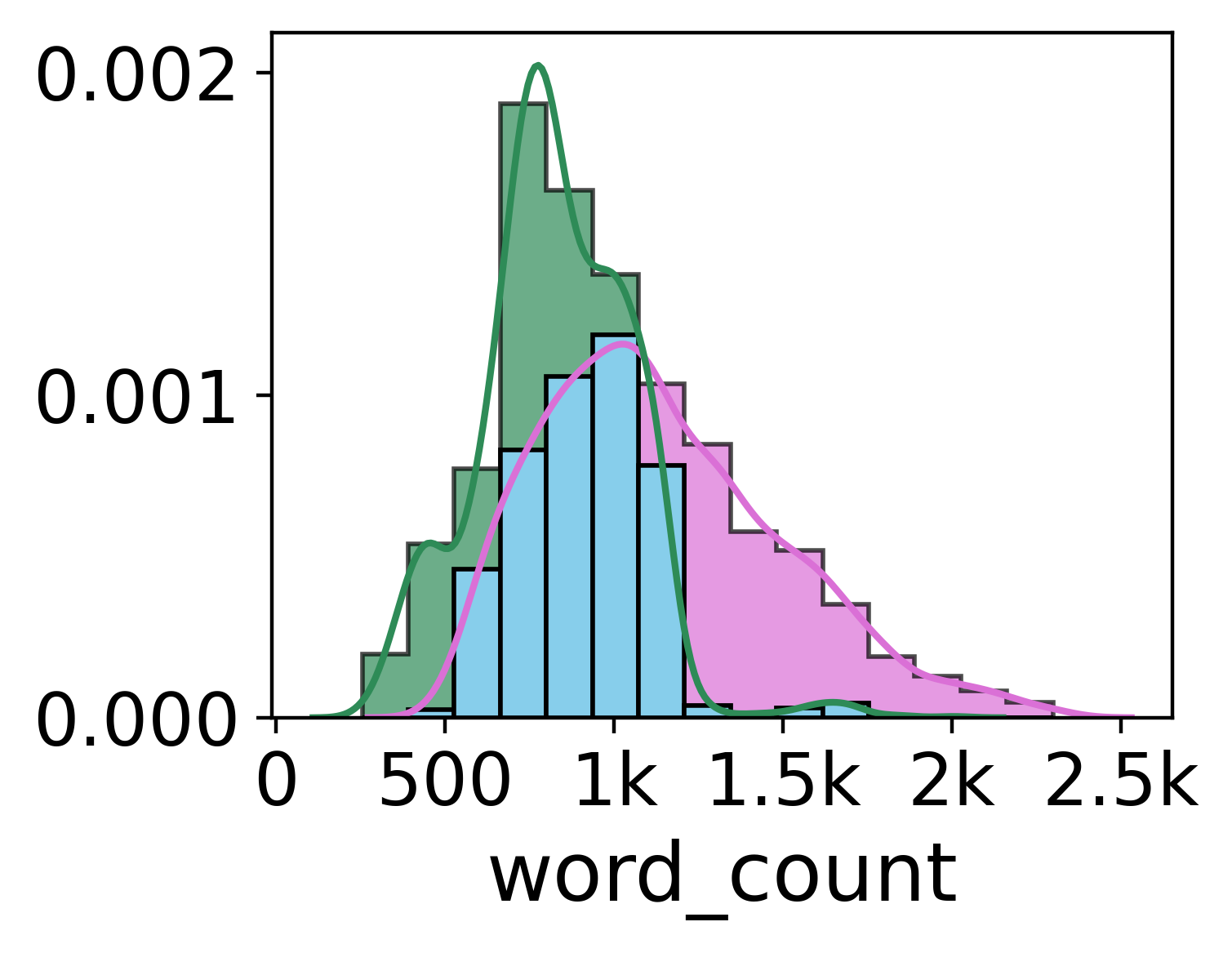}\par
  \end{minipage}

  \vspace{0.4em}
  \includegraphics[width=\columnwidth]{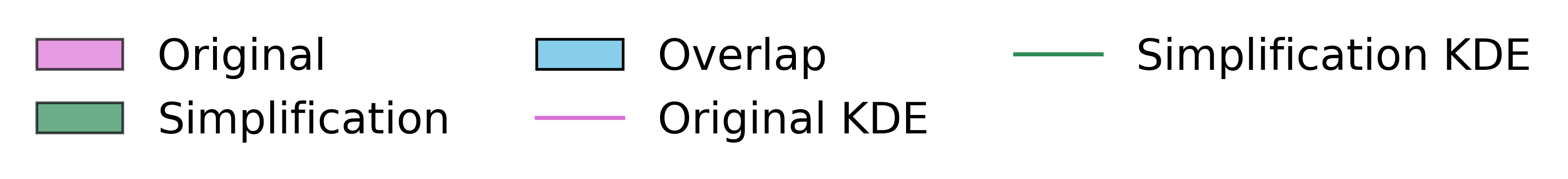}

  \caption{Both \textsc{SimPA} and \textsc{Newsela} show larger distribution shifts by readability than by length transformations. Text-aligned \textsc{Newsela} shows greater distribution shift than the sentence-aligned \textsc{SimPA.}}
  \label{fig:simpa-newsela-hist-grid}
\end{figure}

We curated a multi-domain corpus to evaluate the effectiveness of instruction fine-tuning for controllable text simplification in medicine, public administration, news, and encyclopedic knowledge, representing different text styles and linguistic characteristics. We used the following sources:

\begin{itemize}[left=0pt]
\item \textsc{Med-EASi}~\citep{Basu2023MedEASiFA}. \textbf{Domain}: medical. \textbf{Level}: sentence. \textbf{Creation:} human annotations by medical experts and lay crowd-workers (Toloka), with AI-assisted workflow. \textbf{Mapping:} 1-to-1 complex$\rightarrow$simple. \textbf{Simplification strategies:} token-level spans for \emph{elaboration}, \emph{replacement}, \emph{deletion}, \emph{insertion}.

\item \textsc{SimPA}~\citep{scarton-etal-2018-simpa}. \textbf{Domain}: public administration, \textbf{Level}: sentence. \textbf{Creation:} human simplification in two stages. \textbf{Mapping:} lexical subset is 1-to-3 (three lexical simplifications per complex source); syntactic subset is 1-to-1 (one lexically simplified version further simplified syntactically). \textbf{Simplification strategies:} explicit \emph{lexical} vs.\ \emph{syntactic} simplification.

\item \textsc{WikiLarge}~\citep{zhang-lapata-2017-sentence}. \textbf{Domain}: general/encyclopedic, \textbf{Level}: sentence. \textbf{Creation:} automatic alignment between English Wikipedia and Simple English Wikipedia using similarity heuristics and filtering/cleaning. \textbf{Mapping:} predominantly 1-to-1; the native test set (not used by us) provides multiple references (8 per complex source). \textbf{Simplification strategies:} not explicitly annotated; transformations arise from mined revision/alignment pairs.

\item \textsc{Newsela}~\citep{xu_problems_2015}. \textbf{Domain}: news. \textbf{Level}: document. \textbf{Creation:} professional editors rewrite articles for children at multiple grade levels. \textbf{Mapping:} 1-to-N (up to 4 simplifications per complex source, ordered by decreasing complexity). \textbf{Simplification strategies:} Edits involve content rewrites and frequent sentence splitting.

\end{itemize}

To better understand the learning signal available for different control attributes, we analyze their distributions across datasets (Fig.~\ref{fig:simpa-newsela-hist-grid}). We observe that the document-aligned Newsela dataset shows greater distributional spread across all control attributes than the sentence-aligned datasets.

\subsection{Data Preprocessing Pipeline} \label{subsec:data-preprocessing-pipe}
\paragraph{Harmonization and Metric Calculation.}
All datasets were converted into a unified JSON Lines (JSONL) schema that enables consistent downstream processing. Each entry in the standardized format contains global metadata, source metrics and simplifications array. Global metadata includes: instance id, source text, dataset name, domain and language, annotation type, alignment level, and native split (if applicable). Source metrics include readability values, as well as character and word count. The simplifications array contains one or more simplifications each with the following information (if available, otherwise -1): simplification text, version, compression rates, target control-attribute values and similarity metric values.

The hierarchical JSONL format (one complex source text with multiple text simplifications) was flattened to create individual training instances. For datasets containing multiple reference simplifications for a single source, we converted these into individual complex-simple pairs.

Our approach is based on absolute control (``simplify to FKGL 5 level''), as opposed to relative control (``reduce complexity by X points''). Control attributes are automatically extracted from the dataset. For readability metrics, we computed source (complex) and reference (simplification) values rounded to the nearest integer (<FKGL=5>). For structural attributes, we computed reference/source length ratios in terms of the number of characters and words. The ratios were rounded to one decimal place (<WORD\_COMPRESSION=0.5>).

\paragraph{Subsampling and Stratification.}
To ensure representative train/validation/test splits (80/10/10) across varying text complexities, we employed careful stratified sampling. We removed extreme outliers and filtered out texts falling below the 1st or above the 99th percentile for FKGL, ARI, Dale-Chall, and character length. To determine the optimal split, we utilized the Kolmogorov–Smirnov (KS) goodness-of-fit test. By selecting the partitioning strategy that minimized the KS distance, we ensured that the validation and test sets are statistically representative of the training data. FKGL is our primary stratification feature selected as described in Section~\ref{subsec:setup-dataset-experiments}. We applied stratified sampling by FKGL to keep 3k complex-simple pairs for \textsc{Newsela} and 2k for \textsc{WikiLarge}. Due to the mixed lexical-syntactic strategy employed in \textsc{SimPA}, we merged these subsets by: (1) retaining all unique source (complex) sentences from each subset, and (2) for overlapping sentences, randomly assigning them to either the lexical or syntactic subset with 50-50 probability. This ensures no duplicate complex texts while preserving both simplification types. 

\paragraph{Data filtering.}
To create the filtered subsets, we applied metric-based filtering that retains only instances where all readability values decrease from source (complex) to reference (simple). Instances where complexity increases or remains unchanged were excluded, as they do not reflect the monotonic readability assumption typically associated with simplification. This procedure creates monotonic (mono) variants of each split, with removal statistics logged per dataset and metric. Removal rates range from 10--30\%, depending on dataset characteristics, with automatically aligned corpora (e.g., \textsc{WikiLarge}) exhibiting higher removal rates than manually curated datasets.


\section{Fine-tuning Pipeline}

We fine-tuned the model such that it learns to generate a simplification whose absolute readability value (FKGL, ARI, Dale-Chall) or relative length ratio closely match a control value. Dataset simplifications served both as ground-truth references and sources of the target control attribute value.

\subsection{Prompt Construction}

\begin{figure}[ht]
    \centering
    \hrule
    \vspace{4pt}
    \begin{minipage}{1.0\columnwidth}
        \scriptsize\ttfamily\raggedright
        \setlength{\parskip}{0pt}
        
        \textcolor{blue}{<|begin\_of\_text|>} \\
        \textcolor{blue}{<|start\_header\_id|>system<|end\_header\_id|>} \\
        \vspace{4pt}
        
        \textcolor{ForestGreen}{You are a helpful expert in text simplification. You generate a simplified version of the text input by the user. You simplify the text according to the instructions given by the user. When asked to simplify a text, generate only the requested simplification, without any additional comments, notes or explanations.} \\
        \vspace{4pt}
        \textcolor{blue}{<|eot\_id|>}\\
        \textcolor{blue}{<|start\_header\_id|>user<|end\_header\_id|>} \\
        
        \vspace{4pt}
        
        \textbf{INSTRUCTION}: Simplify the following text such that its Flesch-Kincaid Grade Level (FKGL) score is approximately equal to that specified in the control token prepended to your generated simplification. The control token has the following format: <METRIC=VALUE>. \\
        \vspace{4pt}
        \textbf{SOURCE TEXT}: \textcolor{purple}{\textit{\textbf{<FKGL=4.8>} No cure for the common cold exists, but the symptoms can be treated.}} \\
        \vspace{4pt}
        \textbf{EXPLANATION}: The \textbf{<FKGL=4.0>} token specifies that the target Flesch-Kincaid Grade Level should be approximately 4.0. Lower values indicate simpler text. \\
        \vspace{4pt}
        \textcolor{blue}{<|eot\_id|>} \\
        
        \vspace{4pt}
        
        \textcolor{blue}{<|start\_header\_id|>assistant\\
        <|end\_header\_id|>} \\
        \textcolor{Plum}{\textbf{<FKGL=4.0>}}\textcolor{blue}{<|eot\_id|>}
    \end{minipage}
    \vspace{4pt}
    \hrule
    \vspace{4pt}
    \caption{\textcolor{blue}{Blue}: model-native automatic prompt formatting. \textcolor{ForestGreen}{Green}: system prompt. \textcolor{purple}{Purple}: source text with its control attribute value. \textcolor{Plum}{Plum}: target (reference) control attribute value.}
    \label{fig:llama-prompt-exact}
\end{figure}

\begin{table}[t]
\centering

\hrule
\vspace{4pt}
\begin{minipage}{1.0\columnwidth}
  \scriptsize\ttfamily\raggedright
  \setlength{\parskip}{0pt}

  \vspace{3pt}

  \textcolor{purple}{\textbf{<FKGL=12.3>}} \hfill \textbf{\textit{\textcolor{blue}{Source}}} \\
  Under optimal conditions, it can destroy an entire orchard in a single growing season. \\ \vspace{2pt}
  \textcolor{purple}{\textbf{<FKGL=9.5>}} \hfill \textbf{\textit{\textcolor{blue}{Reference}}} \\
  If the conditions are right, it can destroy an entire orchard in a single growing season. \\ \vspace{3pt}
  \hrule
  \vspace{3pt}

\textcolor{purple}{\textbf{<FKGL=7.6>}} \hfill \textit{\textbf{\textcolor{blue}{Qwen3-8B}}}\\
It can destroy an entire orchard in a single growing season.\\

\textcolor{purple}{\textbf{<FKGL=8.4>}} \hfill \textit{\textbf{\textcolor{blue}{Ministral-3B-Instruct}}}\\
In the perfect conditions, it can destroy an entire orchard in one season. \\ \vspace{4pt}

\textcolor{purple}{\textbf{<FKGL=9.2>}} \hfill \textit{\textbf{\textcolor{blue}{Llama-3.2-1B-Instruct}}}\\
The fungus can destroy an entire orchard in a single growing season.\\ \vspace{4pt}

\textcolor{purple}{\textbf{<FKGL=9.2>}} \hfill \textit{\textbf{\textcolor{blue}{Llama-3.1-8B-Instruct}}}\\
The fungus can destroy an entire orchard in a single growing season. \\ \vspace{4pt}

\textcolor{purple}{\textbf{<FKGL=12.3>}} \hfill \textit{\textbf{\textcolor{blue}{Mistral-7B-Instruct-v0.1}}}\\
Under optimal conditions, it can destroy an entire orchard in a single growing season.\\ \vspace{4pt}

\textcolor{purple}{\textbf{<FKGL=12.3>}} \hfill \textit{\textbf{\textcolor{blue}{Qwen3-1.7B}}}\\
Under optimal conditions, it can destroy an entire orchard in a single growing season.\\ \vspace{4pt}
\end{minipage}
\vspace{4pt}
\hrule
\vspace{4pt}
\caption{Dataset: \textsc{MedEASi}. Control attribute: FKGL. Some models copy the source sentence, resulting in no simplification, while others successfully generate simplifications matching the target FKGL. Even minimal lexical changes can result in significant readability value shifts. }
\label{tab:medeasi-fkgl-outputs}

\end{table}

 To reduce over-reliance on a single phrasing, we utilized six manually curated system prompt variants, with randomly selection during fine-tuning. As shown in Fig.~\ref{fig:llama-prompt-exact}, the instruction prompt contains the complex source text and the reference control attribute. We employed a dynamic prompting strategy to integrate control attributes into the fine-tuning process. We used embedded control tokens in the format <METRIC=VALUE> prepended to the assistant's response. Values were precomputed based on reference simplifications and rounded to one decimal place. To maintain consistency with the models' pre-training, all prompts were formatted using model-native \textbf{chat template} via Hugging Face’s \texttt{tokenizer.apply\_chat\_template()}. See Appendix~\ref{sec:appendix-prompt} for technical details.

\subsection{Models}
We conducted experiments across three open-source model families spanning the 1B–14B parameter range. The model lineup includes instruct models from three families: Llama \citep{dubey2024llama} (3.2-1B, 3.2-3B, 3.1-8B, 2-13B), Mistral \citep{jiang2023mistral} (Ministral-3B, Mistral-7B-v0.3), and Qwen \citep{yang2025qwen3} (Qwen3-1.7B, -4B, -7B, -14B). For models exceeding 4B parameters, we used LoRA~\citep{LoRA} to manage computational constraints.

\subsection{Training Objective}
 By including the control token at the start of the assistant's turn, the model learns the conditional relationship between the specified metric value and the linguistic features of the generated simplification. We use the standard cross-entropy loss for training and validation, refraining from additional signals (e.g. based on a prediction-reference error) and keep training as a pure causal language modeling task. We mask the prompt tokens, and only the completion contributes to the loss.

\subsection{Evaluation Metrics}
We evaluate the models across the three distinct dimensions of controllability, simplification quality, and textual similarity. Controllability is measured using Mean Absolute Error (MAE) between the target and prediction. Simplification quality is assessed via SARI \citep{xu_optimizing_2016_sari} and LENS \citep{maddela-etal-2023-lens}, additionally reporting COMET \citep{rei-etal-2020-comet} as a general-purpose semantic similarity metric. Textual similarity is evaluated using BLEU \citep{papineni-etal-2002-bleu} and BERTScore \citep{Zhang2019BERTScoreET} against both the complex and reference texts.

\subsection{Robust Inference}
To mitigate the inherent non-determinism of LLMs, we did multiple independent inference runs (five for smaller models, three for larger models fine-tuned with LoRA \citep{LoRA}). The results were aggregated to report the mean and standard deviation for each metric. We use greedy decoding to select the most probable token at each step.


\section{Experimental Setup} \label{sec:experiment-setup}
\subsection{Dataset Experiments} \label{subsec:setup-dataset-experiments}
\paragraph{Stratified Partitioning.}
To ensure that the splits are representative of the overall dataset, we carry out stratified partitioning experiments. We evaluated multiple stratification strategies by sampling based on readability level metrics (FKGL, ARI, Dale-Chall) and length (word and char count) of the complex source text. To evaluate the preservation of the original distribution, we utilized the Kolmogorov–Smirnov (KS) goodness-of-fit test using \texttt{SciPy} package \citep{2020SciPy-NMeth} to compare the splits against the full dataset, repeated over 10 random seeds for robustness. The winning approach yielding the lowest KS distance was chosen to generate the final 80/10/10 splits.

\paragraph{Sampling from \textsc{WikiLarge}.}

\begin{figure}[h]
    \centering

    \begin{subfigure}[c]{0.5\linewidth}
        \centering
        \includegraphics[width=\linewidth]{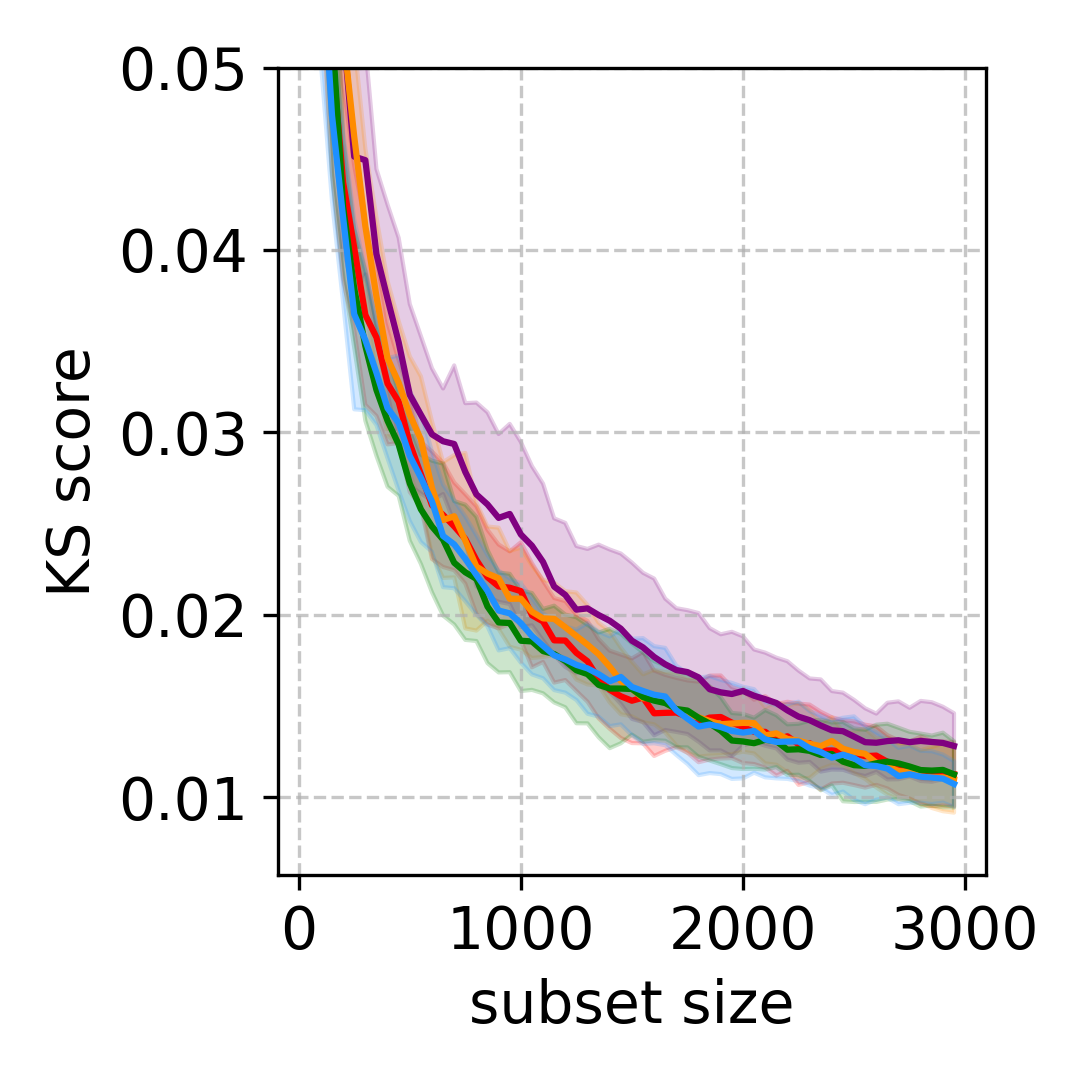}
    \end{subfigure}%
    \hfill
    \begin{subfigure}[c]{0.5\linewidth}
        \centering
        \includegraphics[width=\linewidth]{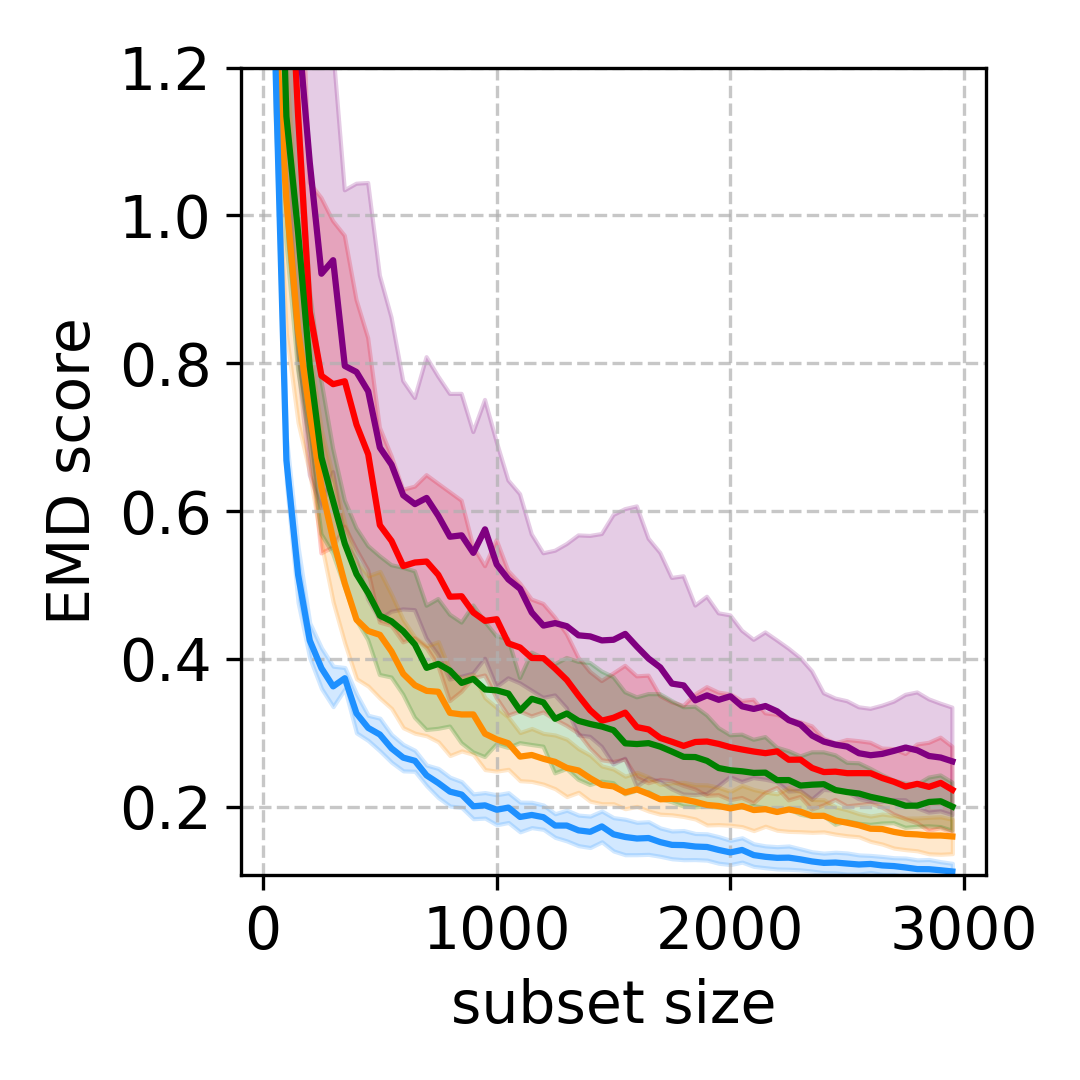}
    \end{subfigure}%

    \begin{subfigure}[c]{0.5\linewidth}
        \centering
        \includegraphics[width=\linewidth]{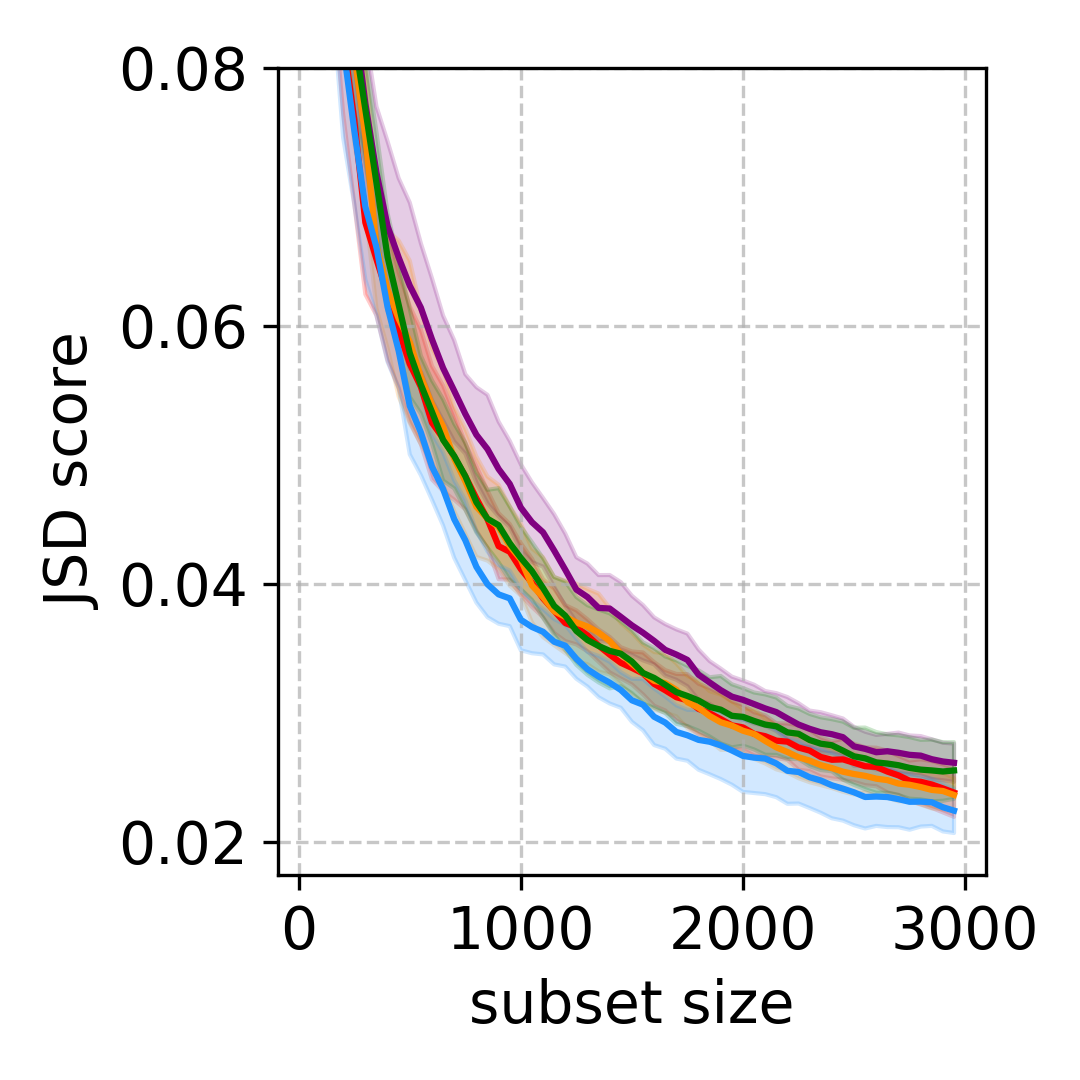}
    \end{subfigure}
    \hfill
    \begin{subfigure}[c]{0.3\linewidth}
        \centering
        \includegraphics[width=\linewidth]{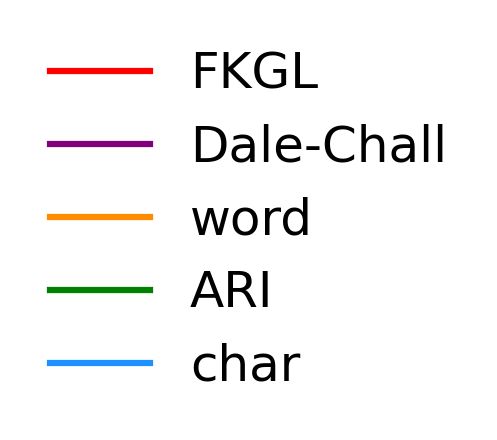}
    \end{subfigure}
    \caption{\textsc{WikiLarge}. \textit{Global} sampling with stratification by readability and length, measured in terms of KS, EMD and JSD. Stratification by $N$ chars shows smallest divergence across all metrics.}
    \label{fig:sampling-wiki}

    \begin{subfigure}[c]{0.5\linewidth}
        \centering
        \includegraphics[width=\linewidth]{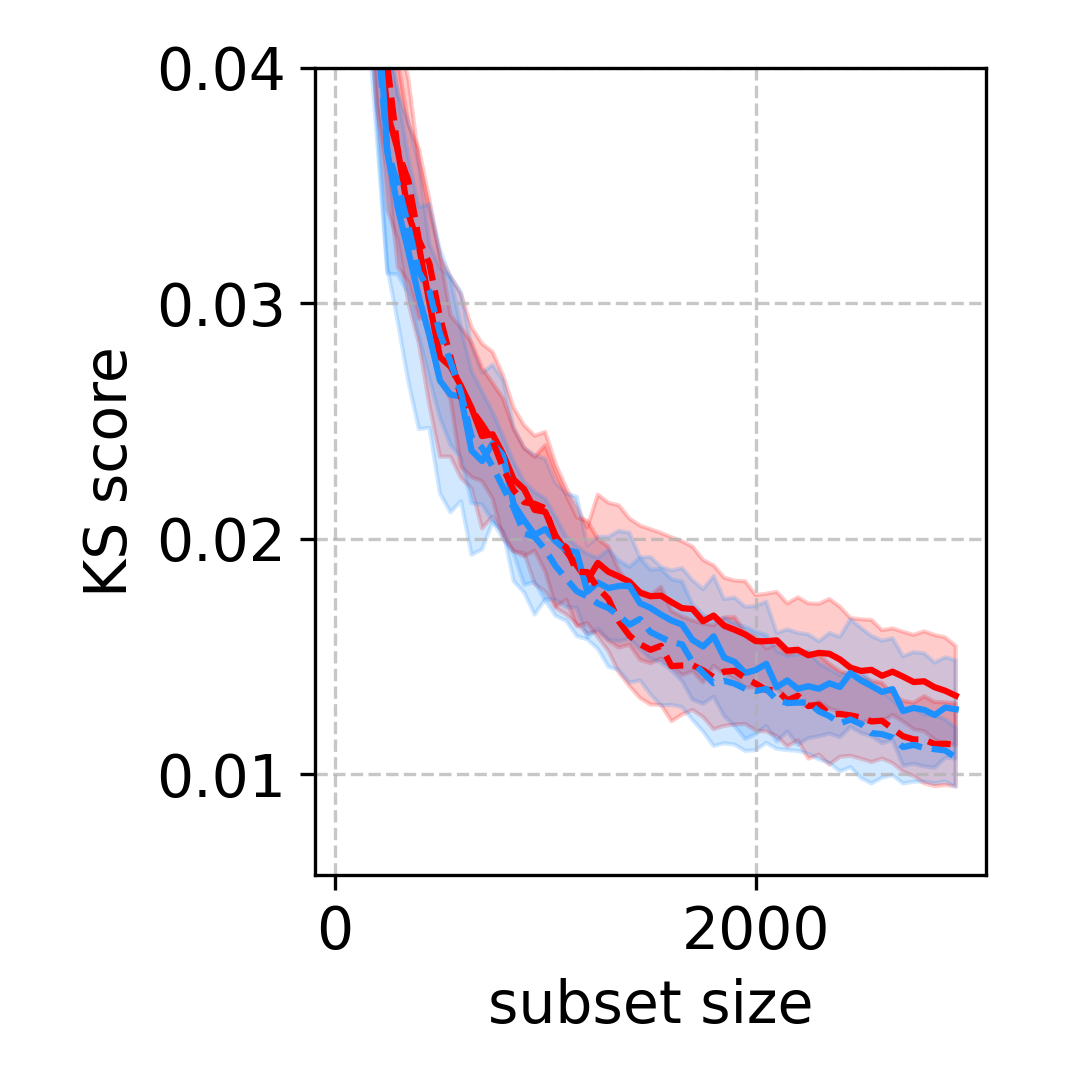}
    \end{subfigure}
    \hfill
    \begin{subfigure}[c]{0.4\linewidth}
        \centering
        \includegraphics[width=\linewidth]{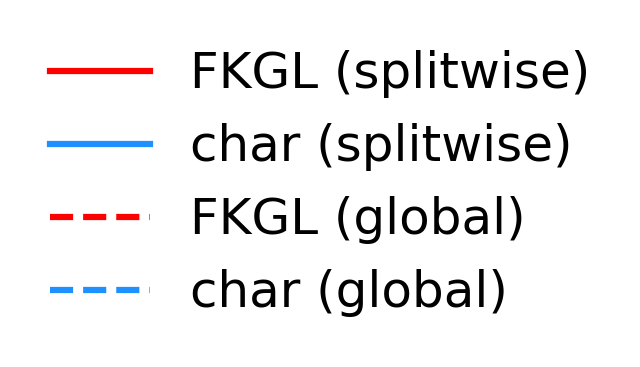}
    \end{subfigure}

    \caption{\textsc{WikiLarge}. A comparison of split-based and split-agnostic sampling shows the latter yields lower divergence across both length and readability, with FKGL and $N$ chars yielding similar results.}
    \label{fig:sampling-wiki-both}

\end{figure}

Given that the \textsc{Wikilarge} corpus contains a native train/dev/test split, we conduct a stratified sampling experiment to determine an optimal strategy to minimize divergence across all five control attributes. We compared two approaches: \textit{global} sampling (from the entire dataset pool) and \textit{split-wise} sampling (from the native train/dev/test partitions), exploring subset sizes in the 100--3,100 range (step=20) over 10 random seeds. The metrics used are the Kolmogorov-Smirnov test (KS), Jensen–Shannon Divergence (JSD), and Earth Mover’s Distance (EMD), computed with \texttt{SciPy} \citep{2020SciPy-NMeth}.

\subsection{Instruction Fine-Tuning Experiments} \label{subsec:ift-experiments}
\paragraph{Main Experiment.}
The primary experiment in this study evaluates the effectiveness of instruction fine-tuning (IFT) for CATS across four distinct domains and five control attributes. We fine-tuned our selection of LLMs on the full training sets of \textsc{Med-EASi}, \textsc{SimPA}, \textsc{WikiLarge}, and \textsc{Newsela}. Each model was trained separately for each of the five control attributes: readability (FKGL, ARI, and Dale-Chall) and length (character and word compression).

\paragraph{Mono-Datasets Ablation.}
We performed a fine-tuning experiment on the filtered subsets of the dataset described in Section~\ref{subsec:data-preprocessing-pipe}. We hypothesized that removing uninformative instances might reduce computational demands and provide a clearer gradient for the model to learn the correlation between control tokens and linguistic outcomes.

\paragraph{Scaling Experiment.}
We investigated the relationship between model size and performance by scaling from 1B to 14B parameters within models of the same family and, if available, same model generation. The scaling experiment was conducted only on the \textsc{Med-EASi} and \textsc{SimPA} datasets using FKGL and char compression as control attributes.

\section{Results} \label{sec:results}

\begin{table}[t]
\centering
\large
\resizebox{\linewidth}{!}{%
\begin{tabular}{ll ll ll}
\toprule

\multicolumn{2}{c}{\textbf{SARI $\uparrow$}} &
\multicolumn{2}{c}{\textbf{LENS $\uparrow$}} &
\multicolumn{2}{c}{\textbf{MAE $\downarrow$}} \\


\multicolumn{1}{c}{1B} & \multicolumn{1}{c}{8B} &
\multicolumn{1}{c}{1B} & \multicolumn{1}{c}{8B} & 
\multicolumn{1}{c}{1B} & \multicolumn{1}{c}{8B}\\
\hline

\rowcolor{green!12}
\multicolumn{6}{c}{\textsc{\textbf{Med-EASi}} FKGL} \\

\textit{43.68} & \textit{36.20} & 
\textit{59.03} & \textit{68.95}&
\textit{3.37} & \textit{4.89 }\\

42.64 \textcolor{BrickRed}{-1.04} & 43.99 \textcolor{ForestGreen}{+7.77} & 
57.33 \textcolor{BrickRed}{-1.70}& 59.83 \textcolor{BrickRed}{-9.12}&
3.96 \textcolor{BrickRed}{+0.51}& 2.91 \textcolor{ForestGreen}{?} \\

\rowcolor{green!12}
\multicolumn{6}{c}{\textsc{\textbf{Med-EASi}} char} \\

\textit{32.38 }& \textit{36.76} &
\textit{21.23} & \textit{62.89}&
\textit{1.23} & \textit{0.37} \\

42.43 \textcolor{ForestGreen}{+10.05} & 45.25 \textcolor{ForestGreen}{+8.49}&
55.74 \textcolor{ForestGreen}{+34.51}& 59.90 \textcolor{BrickRed}{-2.99}&
0.39 \textcolor{ForestGreen}{-0.84}& 0.21 \textcolor{ForestGreen}{-0.16}\\

\midrule

\rowcolor{RoyalBlue!8}
\multicolumn{6}{c}{\textsc{\textbf{SimPA}} FKGL} \\

\textit{42.90} & \textit{23.03 }& 
\textit{55.29} & \textit{69.34}&
\textit{2.92} & \textit{6.46} \\

42.15 \textcolor{BrickRed}{-0.75} & 41.81 \textcolor{ForestGreen}{+18.78} & 
55.05 \textcolor{BrickRed}{-0.24}& 55.42 \textcolor{BrickRed}{-13.92}&
2.56 \textcolor{ForestGreen}{-0.36} & 2.18 \textcolor{ForestGreen}{-4.28} \\

\rowcolor{RoyalBlue!8}
\multicolumn{6}{c}{\textsc{\textbf{SimPA}} char} \\

\textit{33.07} & \textit{27.31} &
\textit{26.22} &  \textit{61.86}&
\textit{0.73} & \textit{0.33}\\

38.95 \textcolor{ForestGreen}{+5.88}& 59.14 \textcolor{ForestGreen}{+31.83} &
53.93 \textcolor{ForestGreen}{+27.21}& 57.54 \textcolor{BrickRed}{-4.32}&
0.18 \textcolor{ForestGreen}{-0.55}& 0.04 \textcolor{ForestGreen}{-0.29}\\

\midrule

\rowcolor{violet!8}
\multicolumn{6}{c}{\textsc{\textbf{WikiLarge}} FKGL} \\

\textit{40.55 }& \textit{25.69 }& 
\textit{50.55 }& \textit{81.37} &
\textit{3.66} & \textit{2.69} \\

38.21 \textcolor{BrickRed}{-2.34} & 37.73 \textcolor{ForestGreen}{+12.04} & 
61.55 \textcolor{ForestGreen}{+11.00}& 60.97 \textcolor{BrickRed}{-20.40}&
4.25 \textcolor{BrickRed}{+0.59} & 2.94 \textcolor{BrickRed}{+0.25} \\

\rowcolor{violet!8}
\multicolumn{6}{c}{\textsc{\textbf{WikiLarge}} char} \\

\textit{34.49} & \textit{37.54} &
\textit{22.53 }& \textit{66.60}&
\textit{1.60} & \textit{0.33}\\

37.90 \textcolor{ForestGreen}{+3.41}& 38.94 \textcolor{ForestGreen}{+1.40}&
60.12 \textcolor{ForestGreen}{+37.59}& 61.55 \textcolor{BrickRed}{-5.05}&
0.40 \textcolor{ForestGreen}{-1.20}& 0.24 \textcolor{ForestGreen}{0.07}\\

\midrule

\rowcolor{orange!6}
\multicolumn{6}{c}{\textsc{\textbf{Newsela}} FKGL} \\

\textit{28.52} &\textit{ 25.69} & 
\textit{61.37} & \textit{81.37}&
\textit{2.89} &\textit{ 2.69} \\

37.60 \textcolor{ForestGreen}{+9.08}& 42.13 \textcolor{ForestGreen}{+16.44}&
40.34 \textcolor{BrickRed}{-21.03}& 50.46 \textcolor{BrickRed}{-30.91}&
2.51 \textcolor{ForestGreen}{-0.38} & 7.78 \textcolor{BrickRed}{+5.09}\\
 
\rowcolor{orange!6}
\multicolumn{6}{c}{\textsc{\textbf{Newsela}} char} \\

\textit{33.25} & \textit{26.88} &
\textit{44.51 }& \textit{70.74}&
\textit{0.35} & \textit{0.56}\\

34.45 \textcolor{ForestGreen}{+1.25} & 38.79 \textcolor{ForestGreen}{+11.91} & 
36.46 \textcolor{BrickRed}{-8.05}& 49.46 \textcolor{BrickRed}{-21.28}&
0.72 \textcolor{BrickRed}{+0.37}& 0.36 \textcolor{ForestGreen}{-0.20}\\


\bottomrule

\end{tabular}
}

\caption{IFT of 1B and 8B Llama models improves SARI and MAE over the non-fine-tuned baseline, with stronger gains for larger models and compression-based controls, whereas LENS does not consistently improve and in some cases decreases after fine-tuning.}
\label{tab:baseline-results}
\end{table}

\begin{table*}[t]
    \centering

    \begin{subtable}[t]{0.33\textwidth}
        \centering
        \LARGE
        \caption{SARI $\uparrow$}
        \renewcommand{\arraystretch}{1.04}
        \resizebox{\linewidth}{!}{
        \begin{tabular}{cccc c cccc}
        \toprule
         \multicolumn{4}{c}{Llama} & & \multicolumn{4}{c}{Qwen} \\
         \cmidrule{1-4}
         \cmidrule{6-9}
        1B & 3B & 8B & 13B & & 1.7B & 4B & 
        8B & 14B \\
        \cmidrule{1-4}
        \cmidrule{6-9}
        \noalign{\vskip -\belowrulesep}
        \rowcolor{green!12}
        42.64 & 43.05 & 43.99 & \textbf{52.02} & & 50.50 & 51.17 & 51.10 & \underline{51.93} \\
        \rowcolor{RoyalBlue!20}
        42.15 & 60.16 & 41.81 & \textbf{67.78} & & 65.52 & 65.07 & \underline{67.65} & 65.90 \\
        \noalign{\vskip -\aboverulesep}
        \bottomrule
        \end{tabular}
        }
        
    \end{subtable}\hfill
    \begin{subtable}[t]{0.33\textwidth}
        \Large
        \centering
        \caption{LENS $\uparrow$}
        \renewcommand{\arraystretch}{1.04}
        \resizebox{\linewidth}{!}{
        \begin{tabular}{cccc c cccc}
        \toprule
        \multicolumn{4}{c}{Llama} & & \multicolumn{4}{c}{Qwen} \\
        \cmidrule{1-4}
        \cmidrule{6-9}
        1B & 3B & 8B & 13B & & 1.7B & 4B & 8B & 14B \\
        \cmidrule{1-4}
        \cmidrule{6-9}
        \noalign{\vskip -\belowrulesep}
        \rowcolor{green!12}
        57.33 & 56.82 & \textbf{59.90} &  57.41  & & 58.18 & 58.96 & \underline{59.27} & 58.88 \\
        \rowcolor{RoyalBlue!20}
        55.05 & \textbf{59.13} & 55.42 & 58.35& & 58.10 & 58.09 & \underline{58.59} & 58.38\\
        \noalign{\vskip -\aboverulesep}
        \bottomrule
        \end{tabular}
        }
    \end{subtable}\hfill
    \begin{subtable}[t]{0.33\textwidth}
        \centering
        \caption{MAE $\downarrow$}
        \resizebox{\linewidth}{!}{
        \begin{tabular}{cccc c cccc}
        \toprule
        \multicolumn{4}{c}{Llama} & & \multicolumn{4}{c}{Qwen} \\
        \cmidrule{1-4}
        \cmidrule{6-9}
        1B & 3B & 8B & 13B & & 1.7B & 4B & 8B & 14B \\
        \cmidrule{1-4}
        \cmidrule{6-9}
        \noalign{\vskip -\belowrulesep}
        \rowcolor{green!12}
        3.96 & 3.49 & 2.91 & 2.99 & & 2.72 & 2.91 & \textbf{2.34} & \underline{2.61} \\
        \rowcolor{RoyalBlue!20}
        2.56 & 1.11 & 2.18 & 1.27 & & 1.22 & 1.08 & \textbf{1.03} & \underline{1.06} \\
        \noalign{\vskip -\aboverulesep}
        \bottomrule
        \end{tabular}
        }
    \end{subtable}\vspace{0.5em}
    \begin{subtable}[t]{0.33\textwidth}
        \centering
        \LARGE
        \caption{SARI $\uparrow$}
        \renewcommand{\arraystretch}{1.04}
        \resizebox{\linewidth}{!}{
        \begin{tabular}{cccc c cccc}
        \toprule
        \multicolumn{4}{c}{Llama} & & \multicolumn{4}{c}{Qwen} \\
         \cmidrule{1-4}
        \cmidrule{6-9}
        1B & 3B & 8B & 13B & & 1.7B & 4B & 8B & 14B \\
        \cmidrule{1-4}
        \cmidrule{6-9}
        \noalign{\vskip -\belowrulesep}
        \rowcolor{green!12}
        42.43 & 41.58 & 45.25 & \underline{52.06} & & \textbf{52.09} & 51.33 & 51.91 & 50.72 \\
        \rowcolor{RoyalBlue!20}
        38.95 & \underline{64.29} & 59.14 & 67.17 & & 62.82 & 62.30 & \textbf{65.96} & 62.94 \\
        \noalign{\vskip -\aboverulesep}
        \bottomrule
        \end{tabular}
        }
        
    \end{subtable}\hfill
    \begin{subtable}[t]{0.33\textwidth}
        \LARGE
        \centering
        \caption{LENS $\uparrow$}
        \renewcommand{\arraystretch}{1.04}
        \resizebox{\linewidth}{!}{
        \begin{tabular}{cccc c cccc}
        \toprule
        \multicolumn{4}{c}{Llama} & & \multicolumn{4}{c}{Qwen} \\
        \cmidrule{1-4}
        \cmidrule{6-9}
        1B & 3B & 8B & 13B & & 1.7B & 4B & 8B & 14B \\
        \cmidrule{1-4}
        \cmidrule{6-9}
        \noalign{\vskip -\belowrulesep}
        \rowcolor{green!12}
        55.74 & 58.45 & 59.83 & 57.47 & & \textbf{60.54} & 58.91 & \underline{59.94} & 59.19 \\
        \rowcolor{RoyalBlue!20}
        53.93 & \underline{58.45} & 57.54 & 58.28 & & 58.29 & 57.55 & \textbf{58.49} & 57.81 \\
        \noalign{\vskip -\aboverulesep}
        \bottomrule
        \end{tabular}
        }

    \end{subtable}\hfill
    \begin{subtable}[t]{0.33\textwidth}
        \centering
        \caption{MAE $\downarrow$}
        \resizebox{\linewidth}{!}{
        \begin{tabular}{cccc c cccc}
        \toprule
        \multicolumn{4}{c}{Llama} & & \multicolumn{4}{c}{Qwen} \\
        \cmidrule{1-4}
        \cmidrule{6-9}
        1B & 3B & 8B & 13B & & 1.7B & 4B & 8B & 14B \\
        \cmidrule{1-4}
        \cmidrule{6-9}
        \noalign{\vskip -\belowrulesep}
        \rowcolor{green!12}
        0.39 & 0.36 & 0.21 & 0.29 & & \underline{0.18} & 0.22 & 0.19 & \textbf{0.15} \\
        \rowcolor{RoyalBlue!20}
        0.18 & \textbf{0.04} & \textbf{0.04} & \underline{0.05} & & 0.07 & \underline{0.05} & \textbf{0.04} & \textbf{0.04} \\
        \noalign{\vskip -\aboverulesep}
        \bottomrule
        \end{tabular}
        }
    \end{subtable}

    \caption{Scaling experiment with a broader parameter size range. Rows in \textcolor{ForestGreen}{green} represent results on \textsc{MED-EASi}, whereas rows in \textcolor{RoyalBlue}{blue} represent results on \textsc{SimPA}. Top value per row and metric is in bold, second-best value is underscored. (a) - (c) FKGL, (d) - (f) char compression.}
    \label{tab:scaling}
        
\end{table*}

\subsection{Data Experiments} \label{subsec:data-experiments}

\paragraph{Stratified Partitioning.}
Using FKGL, ARI, Dale-Chall, character count, and word count as candidate stratification variables, we measured divergence with the KS statistic across multiple bin sizes and random seeds, in order to apply the same partitioning strategy to all datasets. Stratification by FKGL yielded lowest KS divergence between splits (see Fig.~\ref{fig:sampling-wiki}~and~\ref{fig:sampling-wiki-both} for a comparison with sampling from \textsc{WikiLarge}). The results led to our choice of FKGL as the primary stratification variable for split creation. This demonstrates that relying on native dataset splits or random partitioning can introduce substantial distributional mismatch, potentially confounding evaluation but also damaging fine-tuning.

\paragraph{Sampling from \textsc{WikiLarge}.}
Across all bin sizes, \textit{global} sampling resulted in smaller distribution divergence between the original dataset and its subsets. Fig.~\ref{fig:sampling-wiki}~and~\ref{fig:sampling-wiki-both} show a nearly-monotonic decrease in the divergence score (KS) following subset size increase. Using $N$ chars as stratification variable lead to lower divergence scores in both \textit{global} and \textit{split-wise} setups (Fig.~\ref{fig:sampling-wiki-both}).

\subsection{Instruction Fine-Tuning} \label{subsec:results-ift}

\begin{figure*}[t]
  \centering

  \begin{subfigure}[t]{\textwidth}
    \centering
    \includegraphics[width=0.8\linewidth]{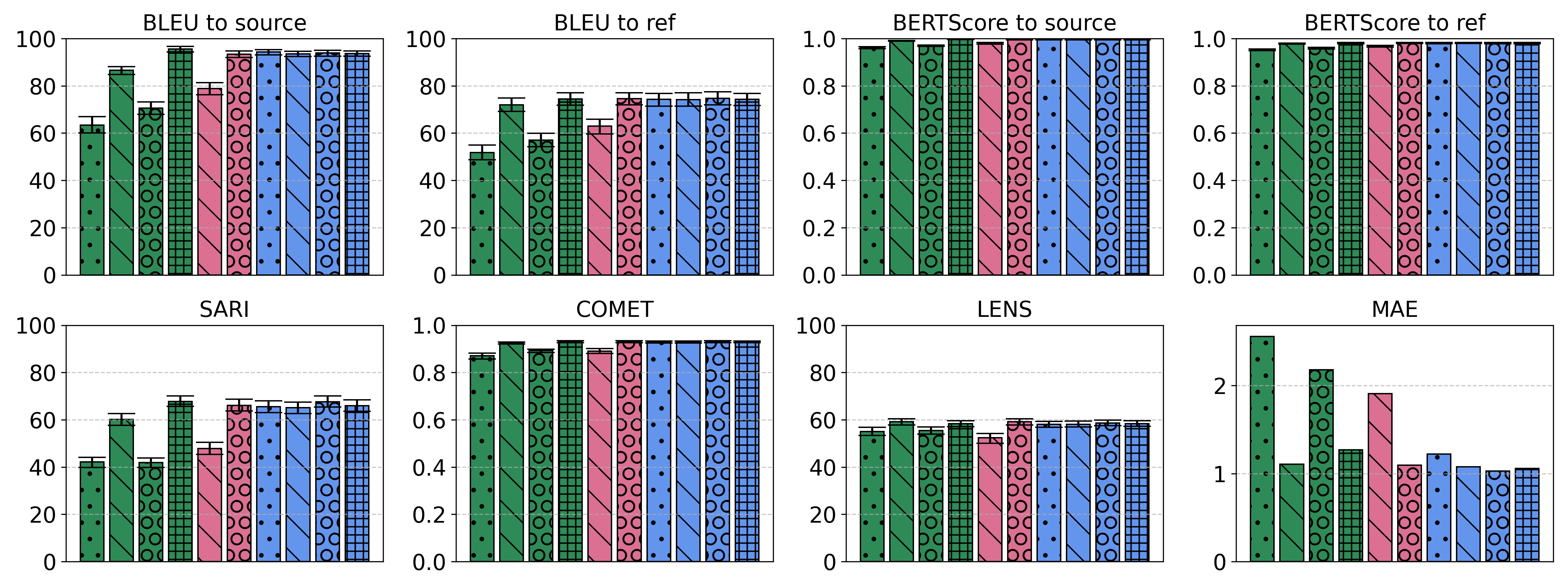}
    \label{fig:fkgl-simpa-main}
  \end{subfigure}


  \begin{subfigure}[t]{\textwidth}
    \centering
    \includegraphics[width=0.8\linewidth]{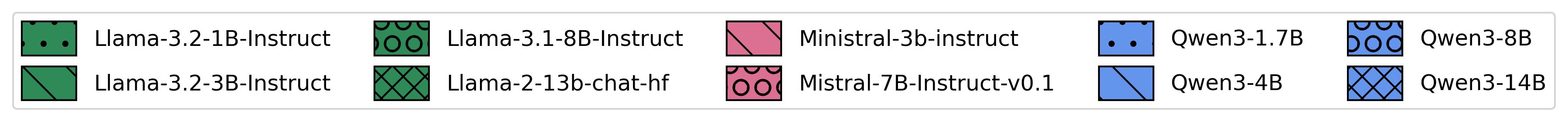}
    \caption*{} 
    \label{fig:fkgl-simpa-legend}
  \end{subfigure}

  \caption{Dataset: \textsc{SimPA}. Control attribute: FKGL. Scaling does not always boost performance. However, we observe strong positive correlation between SARI and COMET, and a strong negative correlation between SARI/COMET and error-based metrics across most datasets and control attributes.}
  \label{fig:results-simpa}
\end{figure*}

\paragraph{Main Experiment.}

Table~\ref{tab:main-results} reports simplification quality (SARI, LENS) and controllability (MAE) across four datasets, three model families, and five control attributes. Performance appears to vary more by \emph{dataset} and \emph{attribute} than by model size.

\textit{Dataset effects.} \textsc{SimPA} yields the strongest results in terms of SARI across most model families and control attributes. This pattern does not hold for LENS, where \textsc{Med-EASi} and \textsc{WikiLarge} sometimes higher scores depending on the model and control attribute. This divergence suggests that SARI and LENS reward different aspects of simplification quality, with their agreement varying greatly across models, control attributes and datasets (see Appendix~\ref{sec:appendix-heatmaps}). \textsc{Newsela} (document-level) shows the lowest quality in terms of both SARI and LENS among all datasets, but not in terms of MAE. See model output examples in Table~\ref{tab:medeasi-fkgl-outputs} and Appendix~\ref{sec:appendix-output-examples}.

\textit{Model-family trends.} Across control attributes, Mistral-7B frequently achieves strong SARI scores on \textsc{Med-EASi} and \textsc{SimPA}. This advantage is less consistent under LENS, where Qwen models often match or outperform while also achieving low control error (MAE). Qwen exhibits most stable performance across datasets and metrics, whereas Mistral and Llama show pronounced dataset-specific peaks. Increasing model size does not lead to consistent improvements: gains are non-monotonic and vary depending on the dataset, control attribute, and evaluation metric.

\textit{Baseline comparison.} 
Under identical inference conditions, only Llama models consistently produce well-formed outputs suitable for evaluation (see Section~\ref{sec:limitations}). Comparing fine-tuned models to their non-fine-tuned counterparts reveals a consistent improvement in SARI and MAE, particularly for larger models and compression-based controls (see Table~\ref{tab:baseline-results} and Appendix~\ref{sec:appendix-scatter}). This trend does not extend to LENS, where fine-tuning leads to drops in LENS despite gains in SARI and MAE. 

\paragraph{Mono-Datasets Ablation.}
Fine-tuning on mono-datasets did not lead to consistent improvements across metrics, but rather introduced a trade-off between simplification quality and controllability (Table~\ref{tab:main-results}). While SARI often decreases or remains comparable, LENS improves on \textsc{Med-EASi} and \textsc{WikiLarge}. Controllability (MAE) does not consistently benefit from mono-dataset training. Qwen models show worse MAE, while Llama and Mistral models show minor fluctuations in both directions. Filtering for monotonic readability reduction seems to either damage training diversity, harmfully shifting the training distribution, or creates dataset that are too small for fine-tuning 
and evaluation.

\begin{table*}[!htb]
\centering

\resizebox{\textwidth}{!}{%
\begin{tabular}{cc cc c cc c cc c cc c cc c cc c cc c cc c cc}
\toprule
\multirow{4}{*}{\textbf{Data}} & \multirow{4}{*}{\textbf{Ctrl.}} 
& \multicolumn{8}{c}{\textbf{SARI $\uparrow$}} &
& \multicolumn{8}{c}{\textbf{LENS $\uparrow$}} &
& \multicolumn{8}{c}{\textbf{MAE $\downarrow$}} \\
\cmidrule{3-10}
\cmidrule{12-19}
\cmidrule{21-28}

& 
& \multicolumn{2}{c}{Llama} &
& \multicolumn{2}{c}{Mistral} & 
& \multicolumn{2}{c}{Qwen} &
& \multicolumn{2}{c}{Llama} & 
& \multicolumn{2}{c}{Mistral} & 
& \multicolumn{2}{c}{Qwen} &
& \multicolumn{2}{c}{Llama} & 
& \multicolumn{2}{c}{Mistral} & 
& \multicolumn{2}{c}{Qwen} \\
\cmidrule{3-4}
\cmidrule{6-7}
\cmidrule{9-10}
\cmidrule{12-13}
\cmidrule{15-16}
\cmidrule{18-19}
\cmidrule{21-22}
\cmidrule{24-25}
\cmidrule{27-28}

& 
& 1B & 8B & & 3B & 7B & & 1.7B & 8B & 
& 1B & 8B & & 3B & 7B & & 1.7B & 8B & 
& 1B & 8B & & 3B & 7B & & 1.7B & 8B \\
\hline

\rowcolor{green!12}
\multicolumn{28}{c}{\textsc{\textbf{Med-EASi}}} \\

base & ARI
& 42.47 & 43.47 & & 44.78 & {\textbf{50.90}} & & 50.32 &  {50.55} &
& 56.89 & \textbf{60.17} & & 50.20 &  59.28 & &  58.74 &  59.33 &
& 4.39 & 3.35 & & 8.19 & 3.30 & & 3.63 & \textbf{2.86} \\

base & DC
& 42.04 & 44.03 & & 43.85 &  {\textbf{52.18}} & & 50.29 &  {50.49} & 
& 55.21 & \textbf{59.64} & & 50.89 &  57.32 & & 57.14 &   57.62 &
& 1.54 & 1.27 & & 1.66 & 1.29 & & 1.39 & \textbf{1.16} \\

base & char
& 42.43 & 45.25 & &  44.86 &  52.03 & &  \textbf{{52.09}} & 51.91 &
& 55.74 & 59.83 & & 51.25 &  57.42 & &  \textbf{60.54}  &  59.94 &
& 0.39 & 0.21 & & 0.70 & 0.28 & & \textbf{0.18} & 0.19 \\

base & word
& 42.59 & 45.31 & & 44.68 &  {51.23} & & 51.13 &  \textbf{{52.00}} &
& 56.75 & 59.97 & & 51.78 &  57.51 & & \textbf{60.62 }&  59.81 &
& 0.37 & \textbf{0.24} & & 0.75 & 0.29 & & \textbf{0.24} & 0.26 \\

base & FKGL
& 42.64 & 43.99 & & 42.39 &  \textbf{{51.87}} & & 50.50 &  {51.10} &
& 57.33 & \textbf{59.90} & & 49.08 &  58.28 & &  58.18 &  59.27 &
& 3.96 & 2.91 & & 6.53 & 2.85 & & 2.72 & \textbf{2.34} \\

mono & FKGL
& 41.95 & 43.43 & & 44.26 & \textbf{50.34} & & 47.81 & 49.53 &
& 59.27 &\textbf{ 61.65} & & 50.09 &  56.45 & &  58.34 &  59.26 &
& 3.61 & \textbf{2.71} & & 4.12 & 3.67 & & 3.44 & 2.93 \\

\hline

\rowcolor{RoyalBlue!8}
\multicolumn{28}{c}{\textsc{\textbf{SimPA}}} \\

base & ARI
& 41.47 & 40.64 & & 52.13 & 64.48 & &  {65.17} &  \textbf{{65.77}} & 
& 55.59 & 53.83 & & 54.25 &  \textbf{58.85} & &  58.44 & 58.48 &
& 3.05 & 2.45 & & 2.53 & 1.24 & & \textbf{1.21} & 1.25 \\

base & DC
& 41.80 & 42.87 & & 47.75 &  \textbf{{65.75}} & & 65.30 &  {65.72} &
& 55.17 & 55.79 & & 51.62 &  \textbf{58.49} & &  57.56 & 58.36  &
& 0.81 & 0.68 & & 0.78 & 0.53 & & 0.51 & \textbf{ 0.47} \\

base & char
& 38.95 & 59.14 & & 50.84 & 65.56 & & 62.82 &  \textbf{{65.96} }&
& 53.93 & 57.54 & & 52.67 & \textbf{ 58.68} & &  58.29 & 58.49 &
& 0.18 & \textbf{0.04} & & 0.23 & 0.05 & & 0.07 & \textbf{0.04} \\

base & word
& 40.42 & 58.49 & & 54.23 &  {62.52} & & 59.09 &  \textbf{{66.50}} &
& 53.49 & 56.80 & & 54.49 &  57.44 & &  55.82 & \textbf{58.16} &
& 0.13 & 0.05 & & 0.06 & \textbf{ 0.04} & & \textbf{0.04 }& 0.08 \\

base & FKGL
& 42.15 & 41.81 & & 47.87 &  {66.10} & & 65.52 & \textbf{ {67.65}} & 
& 55.05 & 55.42 & & 52.35 &  \textbf{59.10} & &  58.10 & 58.59 &
& 2.56 & 2.18 & & 1.91 & 1.10 & & 1.22 & \textbf{1.03} \\

mono & FKGL
& 45.32 & 45.31 & & 50.78 & 57.44 & & 57.26 & \textbf{58.21 }& 
& 51.07 & \textbf{58.06} & & 53.99 &  56.92 & &  57.74 & 57.53 &
& 2.94 & 2.70 & & 2.86 & 2.79 & & \textbf{2.34} & 2.57 \\

\hline

\rowcolor{violet!8}
\multicolumn{28}{c}{\textsc{\textbf{WikiLarge}}} \\

base & ARI
& 38.34 & 37.78 & & 37.89 & 48.18 & &  {49.45} &  \textbf{{63.10}} & 
& 60.64 & \textbf{61.88} & & 48.95 &  57.20 & &  55.24 &  56.87 & 
& 4.69 & 3.78 & & 16.87 & 2.81 & & 2.90 & \textbf{1.44} \\

base & DC
& 38.10 & 37.85 & & 40.23 &  {48.86} & & 48.71 &  \textbf{{49.30}} & 
& 60.91 & \textbf{62.18 }& & 50.02 &  54.74 & &  54.44 &  56.65 &
& 2.33 & 2.40 & & 2.03 & 1.61 & & \textbf{1.56} & 1.63 \\

base & char
& 37.90 & 38.94 & & 39.25 & 48.28 & &  {50.92} &  \textbf{{51.67} }& 
&\textbf{ 60.12 }& 38.94 & & 52.01 &  57.49 & & 53.97 &  55.92 &
& 0.39 & 0.24 & & 0.34 & 0.24 & & 0.14 & \textbf{0.12} \\

base & word
& 37.65 & 40.79 & & 39.24 & 50.10 & &  {50.49} &  \textbf{{51.41} }& 
& 60.10 & \textbf{61.14} & & 52.58 &  55.77 & &  54.76 & 55.00 &
& 0.44 & 0.20 & & 0.45 & 0.22 & & 0.17 & \textbf{0.13} \\

base & FKGL
& 38.21 & 37.73 & & 39.10 &  {49.11} & &  \textbf{{49.47}} & 48.66 & 
& \textbf{61.55} & 60.97 & & 46.89 &  57.56 & & 54.60 &  56.28 &
& 4.25 & 2.94 & & 8.24 & 2.39 & & 2.50 & \textbf{2.24} \\

mono & FKGL
& 40.38 & 41.38 & & 39.64 & 44.59 & & \textbf{44.77} & 44.08 & 
&\textbf{ 65.54} & 63.40 & & 50.18 &  59.17 & & 60.26 & 59.77 &
& 3.46 & \textbf{2.89} & & 7.87 & 3.41 & & 3.43 & 3.08 \\

\hline

\rowcolor{orange!6}
\multicolumn{28}{c}{\textsc{\textbf{Newsela}}} \\

base & FKGL
& 37.60 & 42.13 & & 31.02 &  {43.41} & &  \textbf{{43.51}} & 43.34 &
& 40.34 &  50.46 & & 19.98 &  47.53 & &  48.00 & 49.62 &
& 2.51 & 7.78 & & 26.46 & 2.77 & & 2.43 & \textbf{2.35} \\


base & char
& 34.45 & 38.79 & & 31.49 & 41.62 & &  {43.09} &  \textbf{{43.52}} &
& 36.46 & 49.46 & & 20.97 &  47.61 & & 46.97 &  \textbf{49.82} &
& 0.72 & 0.36 & & 1.39 & 0.31 & & \textbf{0.26} & 0.27 \\

\bottomrule

\end{tabular}
}

\caption{Main experiment results show strong performance of the fine-tuned Qwen models closely followed by the larger Mistral. ``DC'' stands for Dale-Chall. \textit{``Mono''} refers to the monotonically filtered subsets of the respective dataset; \textit{``base''} refers to its full version. ``Word'' and ``char'' denote word- and character-level compression ratio, respectively. Higher is better for SARI and LENS; lower is better for MAE.}
\label{tab:main-results}
\end{table*}

\paragraph{Scaling experiment.}
Scaling does not lead to consistent improvements across evaluation metrics (Table~\ref{tab:scaling}, Appendix~\ref{sec:appendix-barplots}). For FKGL control, SARI generally increases with model size, with larger checkpoints achieving the highest scores. LENS, in contrast, is less affected by model sizes change and remains stable especially across the Qwen family. Controllability (MAE) generally improves with scale. For char-compression control, the effect of scaling is even less pronounced.  
Scaling primarily benefits edit-based metrics such as SARI and, to some extent, MAE, while offering limited gains in LENS.

\section{Discussion} \label{sec:discussion}

Controllable simplification is not only a modeling problem, but is fundamentally constrained by \emph{data and evaluation}. Performance differences are driven by signal availability (attribute variation, distribution match) and by how we measure control. Building CATS systems is largely an exercise in curating \emph{controllable signal} and \emph{measuring target compliance}, with model choice and scale playing a secondary role provided a suitable fine-tuning framework.

\textit{Splits and sampling shape training and evaluation.}
Our stratified partitioning experiments (Section~\ref{subsec:data-experiments}) show that random or native splits risk distributional mismatch in control attributes and potentially confound training and evaluation. Practically, controllable ATS benchmarks should report split creation and verify representativeness with divergence checks before attributing gains to modeling.

\textit{Scale is not a reliable proxy for controllability.}
We observe non-monotonic gains with increasing model size (Table~\ref{tab:scaling}, Fig.~\ref{fig:results-simpa}). Smaller models can be competitive: targeted IFT and data properties dominate raw scale; comparing models without controlling for data signal can prompt wrong conclusions.

\textit{Readability control is learnable, compression control is insufficient in sentence-level datasets.}
Across datasets, readability targets (FKGL/ARI/Dale-Chall) are learned more consistently than length. Our attribute distribution analysis indicates that compression targets provide weak training signal because many sentence-aligned corpora contain minimal complex-simple length variation (Fig.~\ref{fig:mean-ctrl-attr}~and~\ref{fig:simpa-newsela-hist-grid}). Progress on compression-controllable ATS requires dedicated datasets that explicitly encode diverse compression ratios.

\textit{Excessive data cleaning risks drowning signal.}
The mono-datasets ablation shows slightly degraded performance, suggesting that strict filtering might remove diversity and shift the training distribution. Retaining broad coverage of attribute values may be better than enforcing monotonic readability reductions, especially with smaller datasets.

\textit{Measuring control requires integration of error-based metrics.}
Traditional metrics (SARI, LENS, BLEU) do not quantify alignment to the target and can reward copying behavior. Dedicated error-based measures are key for measure of control compliance and are necessary to evaluate CATS systems when the objective is \emph{target matching}. While we observe strong (negative) correlation between SARI/LENS and MAE (see Fig.~\ref{fig:results-simpa}), a holistic approach to CATS evaluation requires multiple dimensions of simplification, including deviation from the target value (controllability), fluency (grammaticality) and meaning preservation (adequacy).

\textit{Robustness is part of evaluation.}
Because LLM outputs vary with decoding and random seeds, multi-seed evaluation is essential for stable comparisons. Our inference protocol prevents over-interpreting single-run results.

\section{Conclusion}
We investigated the efficacy of instruction fine-tuning with discrete control tokens to steer open-source LLMs toward readability and compression targets. 
Our experiments demonstrated that IFT with discrete control tokens is a lightweight and flexible method to transform open-source LLMs into steerable simplification systems. While we observe a positive correlation between model size and performance improvement, some outliers (Qwen3-1.7B) match or outperform larger counterparts.

Our data experiments across common text simplification datasets reveal a crucial limitation: the richness of control-attribute signal in the training data limits how well the model can learn to perform an attribute-specific simplification. With most sentence-level datasets showing minimal compression in the complex-simple pairs, the model largely fails to learn to compress, whereas a pronounced difference in the FKGL value in the complex-simple pairs allows the model to learn to generate predictions approximating a target readability level.

We urge for a thoughtful selection of the stratification variable, in particular in a multi-control-attribute setup. As we demonstrated in the sampling and partitioning experiments, defaulting to the dataset's native splits may lead to an unwelcome divergencies in the attribute distribution between the dataset and its subsets. Running sampling by several control attributes before picking one yielding the lowest divergence provides an effective safeguard.

\section{Limitations} \label{sec:limitations}

\paragraph{Automatic evaluation and metric validity.}
We rely on automatic metrics for (i) simplification quality (SARI, LENS, COMET), (ii) similarity (BLEU, BERTScore), and (iii) controllability (MAE on target attributes). These metrics only partially capture human notions of simplicity, adequacy, and fluency: e.g., SARI is biased toward lexical edits, similarity metrics reward copying, LENS is designed specifically to align text simplification quality with human evaluation, but its agreement with other metrics ranges widely between datasets and control attributes (see Appendix~\ref{sec:appendix-heatmaps}). While COMET is not designed for text simplification, MAE quantifies attribute matching but does not guarantee that outputs are acceptable simplifications: human evaluation would be indispensable to validate whether lower attribute error corresponds to better perceived controllability and readability.

\paragraph{Fine-tuning techniques.}
By using LoRA fine-tuning for models above 3B size, we effectively adopt two distinct fine-tuning approaches for smaller and larger LLMs. This puts a limitation on the comparability of model performance evaluation in the scaling analysis. We also adopt two sets of fine-tuning configurations (with and without LoRA) arrived at through hyperparameter tuning.

\paragraph{Baseline comparison.} We are unable to provide a fully uniform non-fine-tuned baseline across Llama, Qwen, and Mistral. Our IFT pipeline relies on a complex shared prompting format (model-native chat template + dynamically inserted control tokens). Llama models largely follow instructions and produce coherent outputs, but the Qwen and Mistral lineup often seem to be overwhelmed by the complex prompt template and tend to produce degenerate output (empty or malformed output, prompt repetition, output repetition loops), preventing meaningful comparison and automatic evaluation. Using simpler or model-specific prompts could yield stronger prompt-only baselines, but would not be directly comparable to the IFT setting due to the altered conditioning format and effective task definition. Allowing model-specific baseline prompts would risk conflating controllability with prompt engineering.

\paragraph{Representativeness of sampled data.}
We use subsets for \textsc{WikiLarge} and \textsc{Newsela} to keep experiments tractable. While we minimize \emph{distributional divergence} between the original corpora and our subsets using stratified sampling (measured via KS/JSD/EMD), some mismatch in attribute distributions can remain, especially in the tails. As a result, reported controllability and simplification scores may differ when training/evaluating on the full datasets.

\paragraph{Cross-attribute analysis.}
We cannot draw robust comparative observations about controllability among different control attributes, because they have different scales. To make such comparisons possible, it would be necessary to normalize the attribute values and attribute-specific errors.

\paragraph{Language inclusiveness.}
We work exclusively with English-language datasets, which naturally limits the generalization of our findings. Scarcity of expert-generated, well-aligned simplification corpora is even more pronounced in other languages.

\section{Acknowledgments}
 This work was supported by the Swiss Innovation Agency Innosuisse, Flagship Inclusive Information and Communication Technology (IICT), funding no. PFFS-21-47. We sincerely thank Prof. Sarah Ebling for her valuable contributions to the study. 

\section{Plain Language Summary}
Some texts are difficult to read and understand for reasons including age, literacy, or language mastery; texts may also be difficult if they come from a specific knowledge field unfamiliar to their reader, such as medical documents or government decisions. The goal of automatic text simplification is to make texts more accessible. Ideally, such systems should also give their users the power to control how simple they want the text to be in terms of its length, vocabulary and syntactic complexity.

In this work, we explore a method that teaches Large Language Models how to simplify text to a desired reading level or length by showing them examples of good simplifications. We apply this method to texts from the fields of medicine, public administration, news and encyclopedic knowledge. We find that models can learn to control readability reasonably well, which can be measured by comparing predefined gold standard simplification with the text produced by the model and measuring how different they are. Controlling text length is a harder task: because complex and simple texts in the existing datasets tend to have similar length, there is not much length transformation the model can learn to imitate. 

We also show that results depend strongly on how the data is selected, prepared and how performance is measured. Common evaluation metrics do not fully capture whether the model follows the instructions. Our findings suggest that the future of controllable simplification systems depends both on improving model capabilities and on improving data quality and evaluation methods.

\section{Bibliographical References}\label{sec:reference}

\bibliographystyle{lrec2026-natbib}
\bibliography{lrec2026-example}

\label{lr:ref}
\bibliographystylelanguageresource{lrec2026-natbib}
\bibliographylanguageresource{languageresource}

\clearpage
\appendix
\onecolumn

\clearpage
\section{Model Performance Comparison}\label{sec:appendix-barplots}

\begin{figure}[!htbp]
    \centering
    \caption{Model performance on \textsc{Newsela}.}
    \label{fig:appendix-results-main-barplots-newsela}

    \begin{subfigure}[t]{0.8\textwidth}
        \centering
        \caption{FKGL}
        \includegraphics[width=\linewidth]{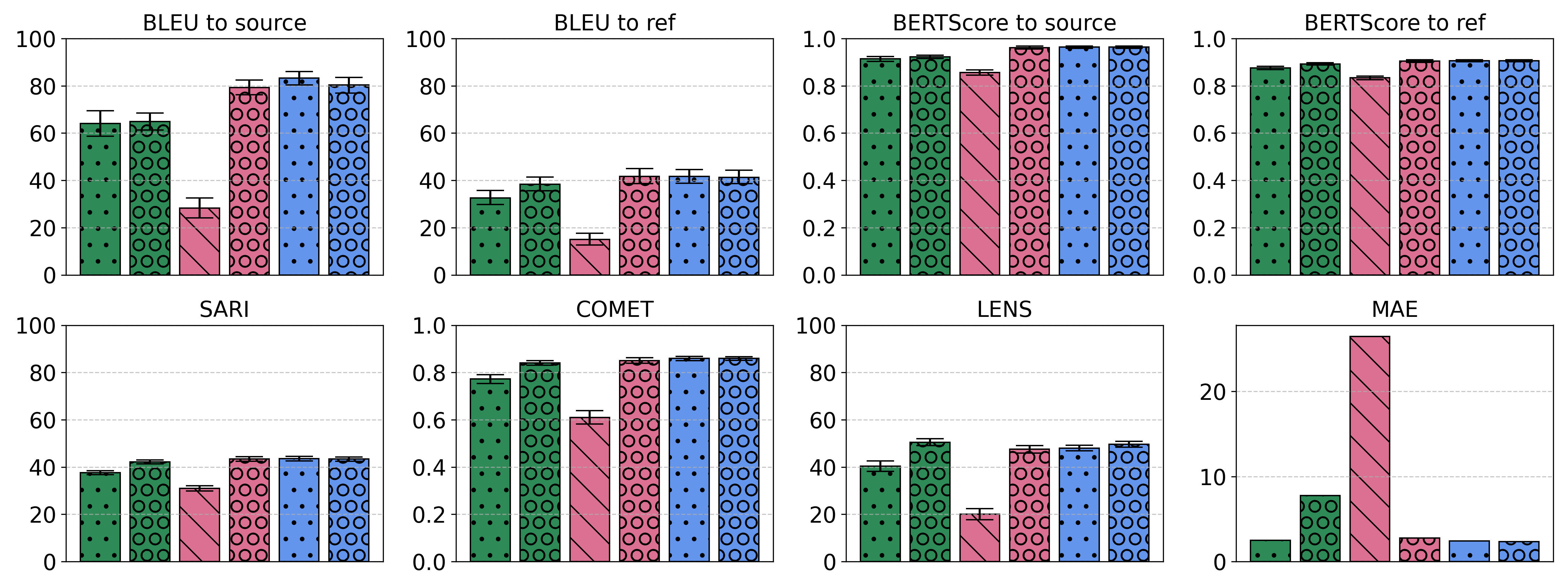}
    \end{subfigure}

    \begin{subfigure}[t]{0.8\textwidth}
        \centering
        \caption{char compression}
        \includegraphics[width=\linewidth]{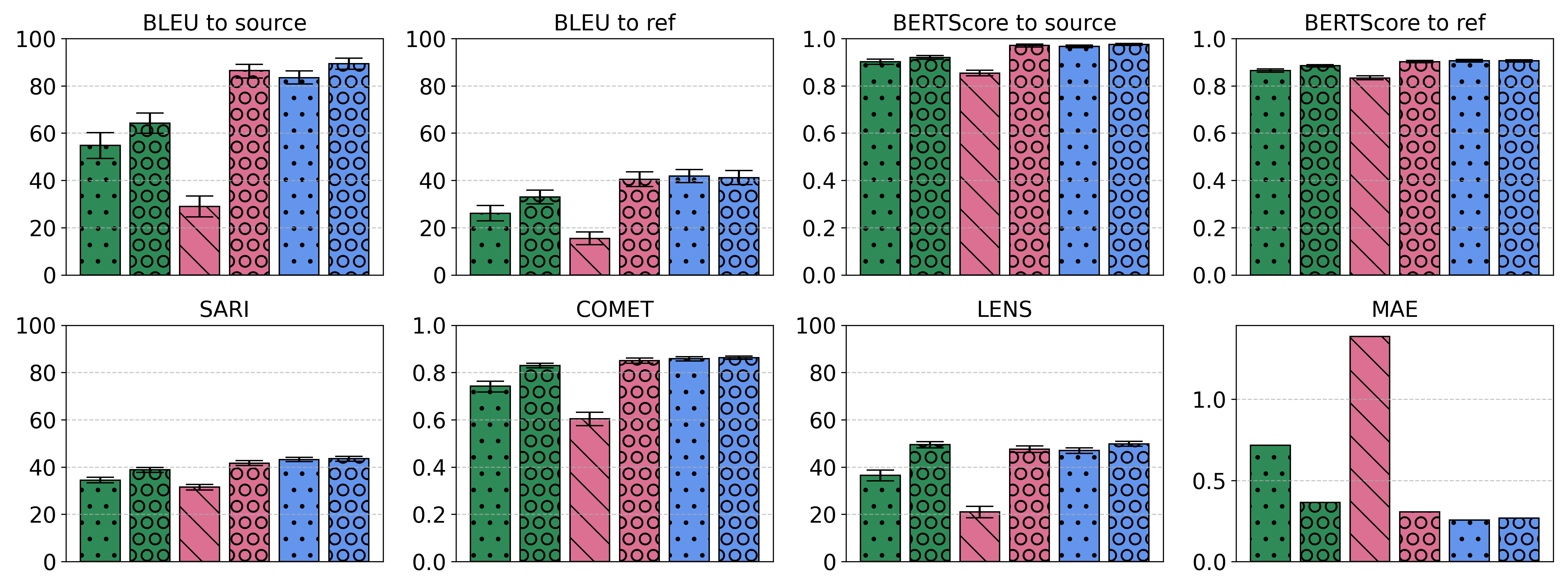}
    \end{subfigure}
    
    \begin{subfigure}[t]{0.8\textwidth}
        \centering
        \includegraphics[width=\linewidth]{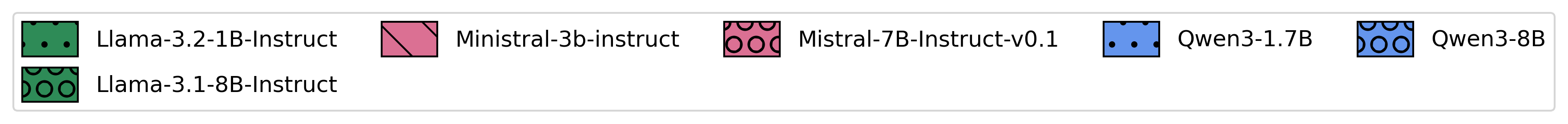}
    \end{subfigure}
    
\end{figure}

\begin{figure}[!htbp]
    \centering
    \caption{Model performance on \textsc{Med-EASi}.}
    \label{fig:appendix-results-main-barplots-medeasi-4}

    \begin{subfigure}[t]{0.8\textwidth}
        \centering
        \caption{ARI}
        \includegraphics[width=\linewidth]{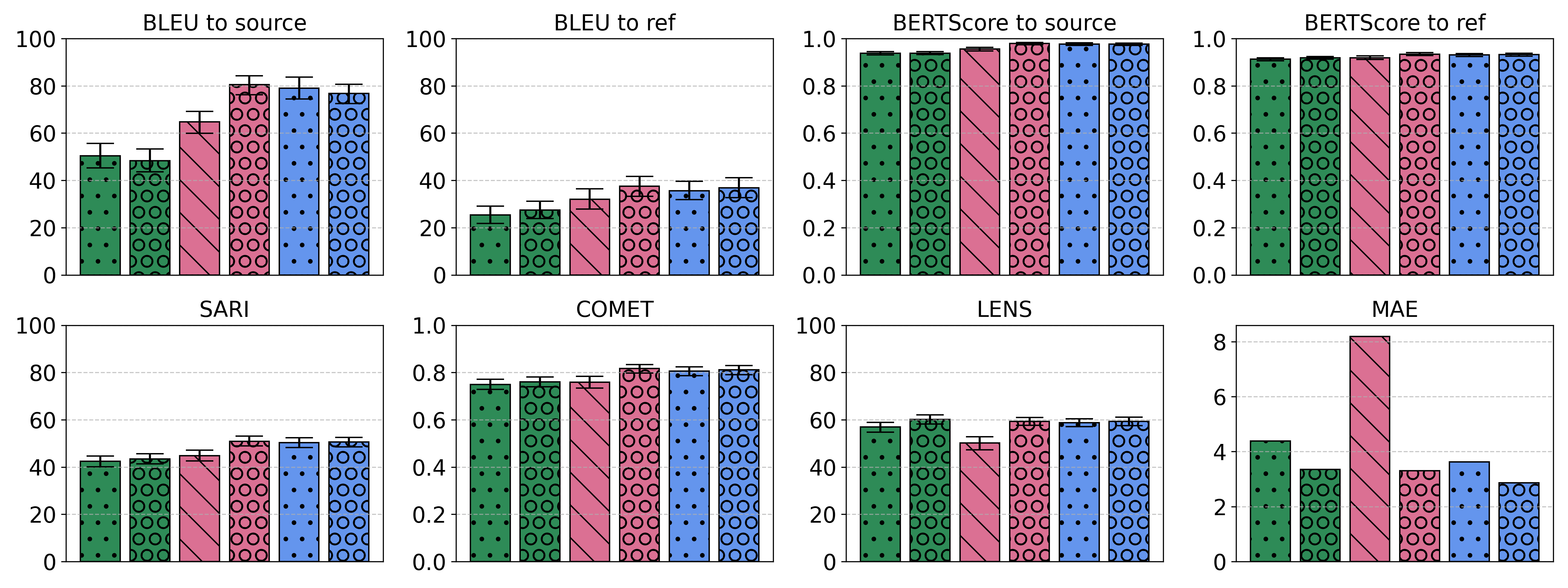}
    \end{subfigure}

    \begin{subfigure}[t]{0.8\textwidth}
        \centering
        \caption{Dale-Chall}
        \includegraphics[width=\linewidth]{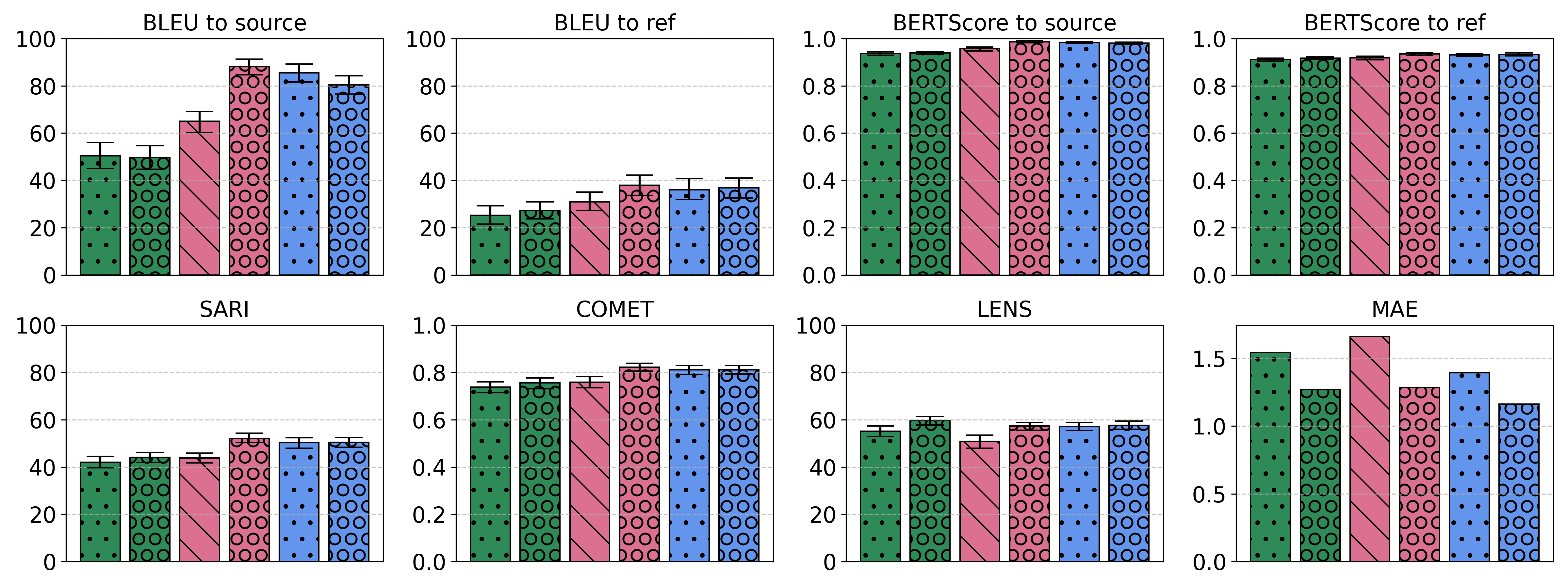}
    \end{subfigure}

    \begin{subfigure}[t]{0.8\textwidth}
        \centering
        \caption{FKGL}
        \includegraphics[width=\linewidth]{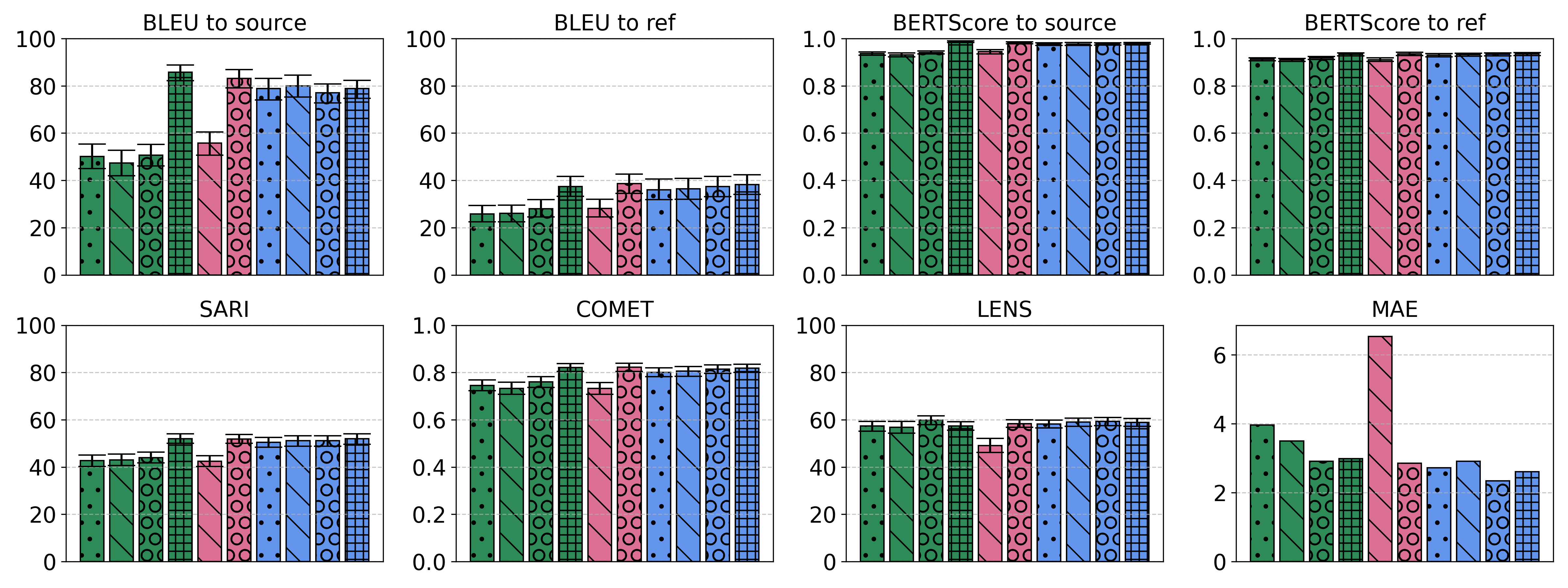}
    \end{subfigure}

    \begin{subfigure}[t]{0.8\textwidth}
        \centering
        \caption{char compression}
        \includegraphics[width=\linewidth]{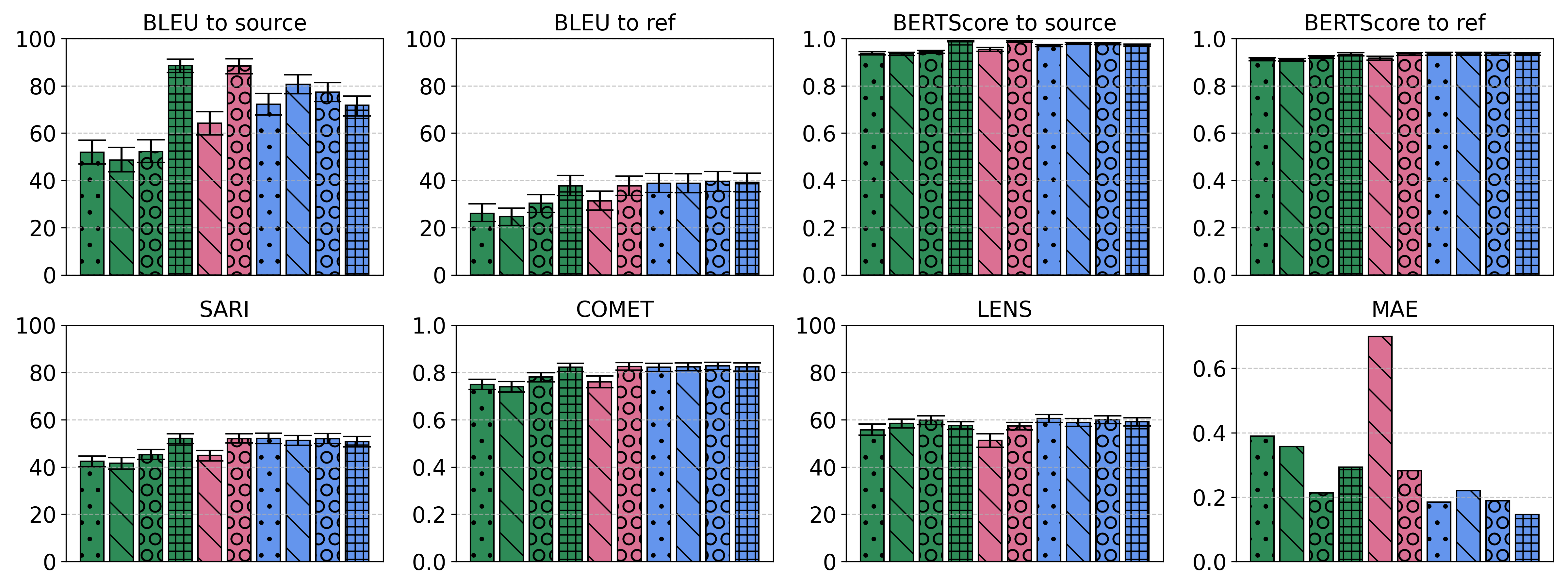}
    \end{subfigure}
    
    \begin{subfigure}[t]{0.8\textwidth}
        \centering
        \includegraphics[width=\linewidth]{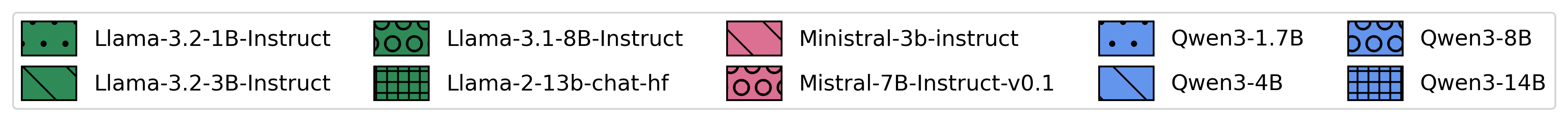}
    \end{subfigure}
    
\end{figure}

\begin{figure}[!htbp]
    \centering
    \caption{Model performance on \textsc{SimPA}.}
    \label{fig:appendix-results-main-barplots-simpa-4}

    \begin{subfigure}[t]{0.8\textwidth}
        \centering
        \caption{ARI}
        \includegraphics[width=\linewidth]{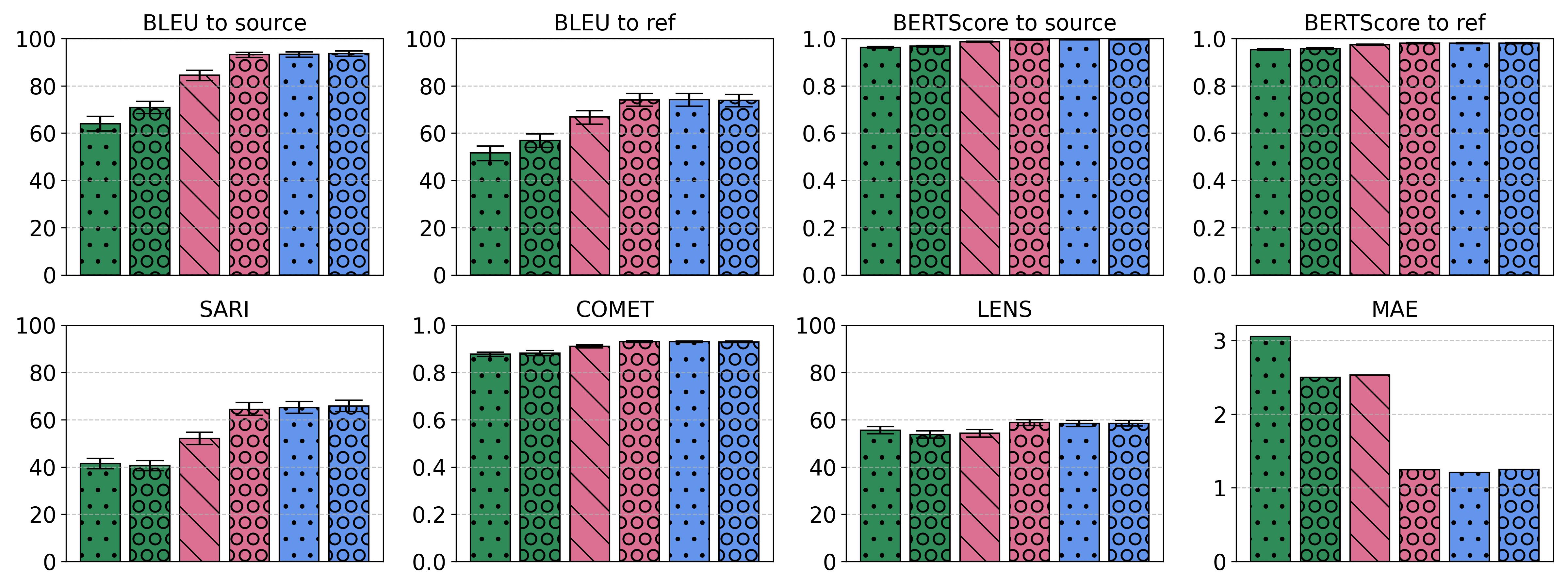}
    \end{subfigure}

    \begin{subfigure}[t]{0.8\textwidth}
        \centering
        \caption{Dale-Chall}
        \includegraphics[width=\linewidth]{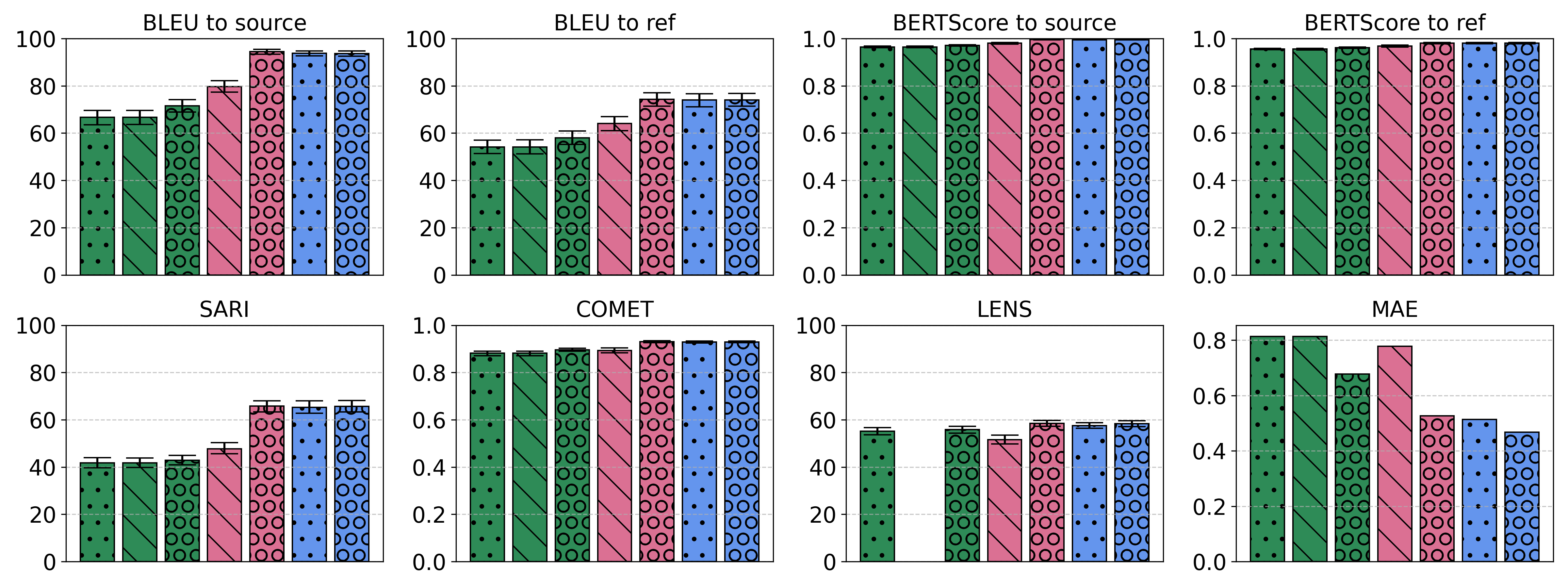}
    \end{subfigure}

    \begin{subfigure}[t]{0.8\textwidth}
        \centering
        \caption{word compression}
        \includegraphics[width=\linewidth]{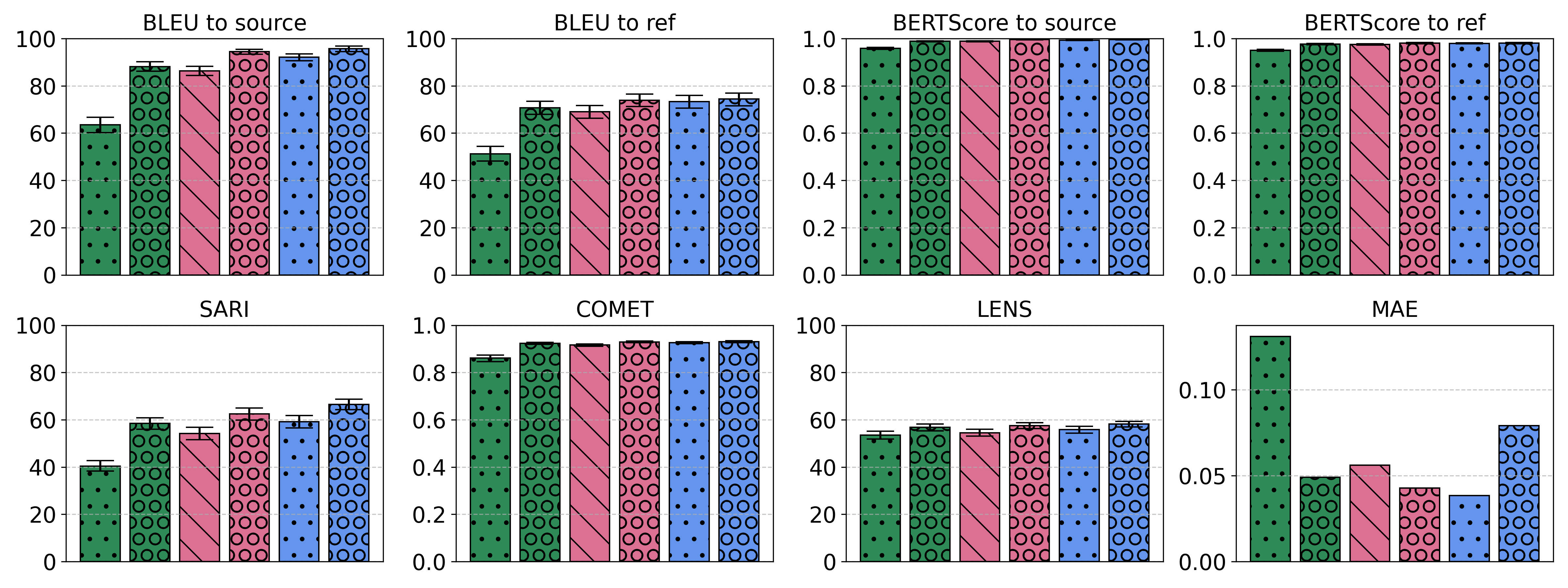}
    \end{subfigure}

    \begin{subfigure}[t]{0.8\textwidth}
        \centering
        \caption{char compression}
        \includegraphics[width=\linewidth]{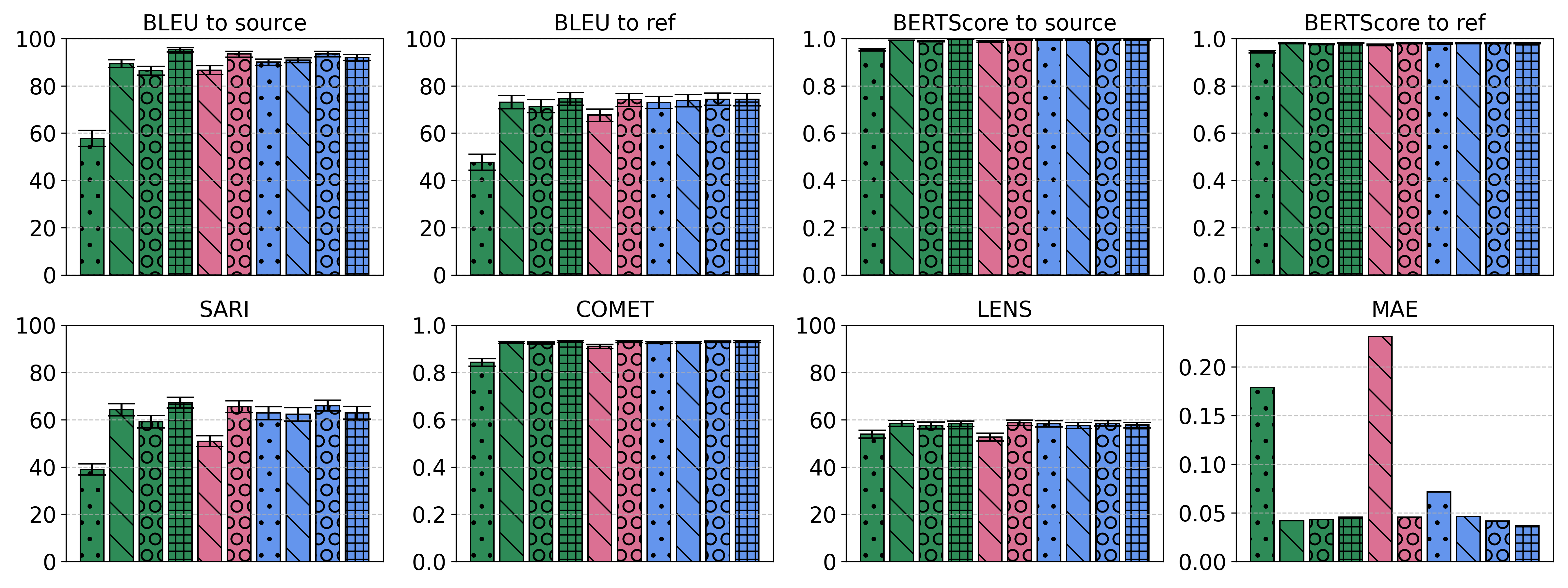}
    \end{subfigure}
    
    \begin{subfigure}[t]{0.8\textwidth}
        \centering
        \includegraphics[width=\linewidth]{assets/performance/CHAR_COMPRESSION_Med-EASi_token_explanation_legend.png}
    \end{subfigure}
    
\end{figure}

\begin{figure}[!htbp]
    \centering
    \caption{Model performance on \textsc{WikiLarge}.}
    \label{fig:appendix-results-main-barplots-wiki-4}

    \begin{subfigure}[t]{0.8\textwidth}
        \centering
        \caption{ARI}
        \includegraphics[width=\linewidth]{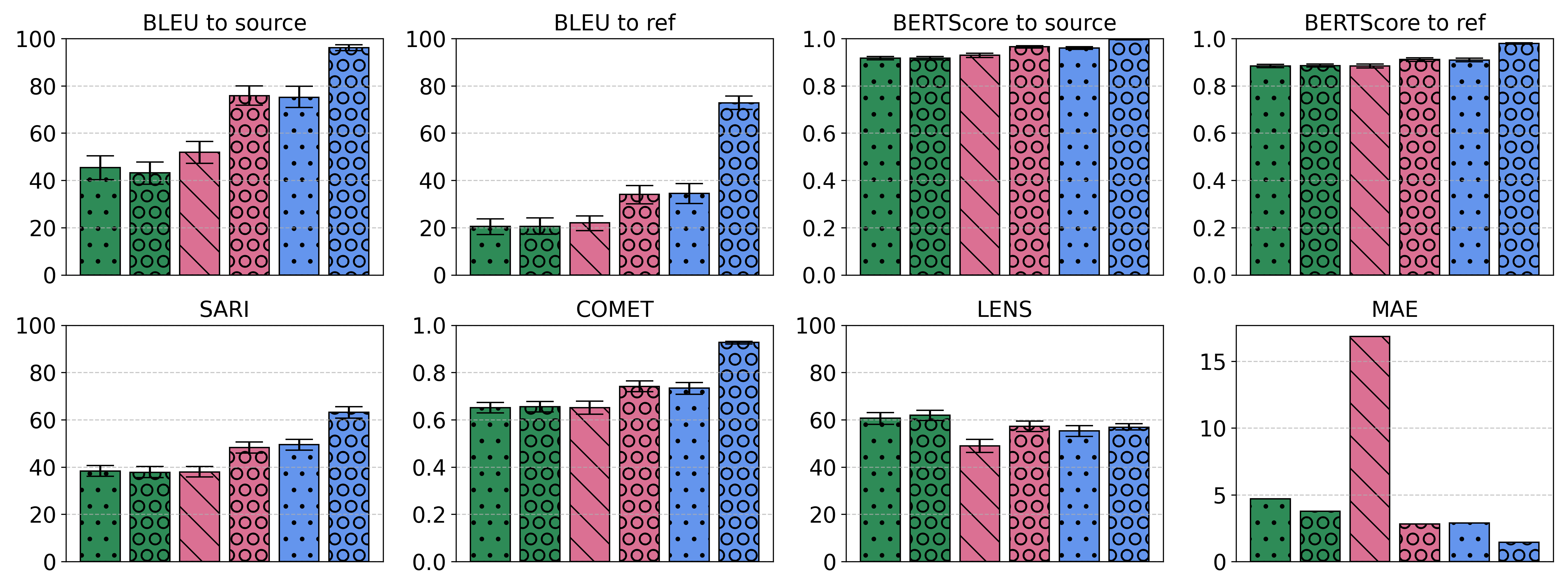}
    \end{subfigure}

    \begin{subfigure}[t]{0.8\textwidth}
        \centering
        \caption{Dale-Chall}
        \includegraphics[width=\linewidth]{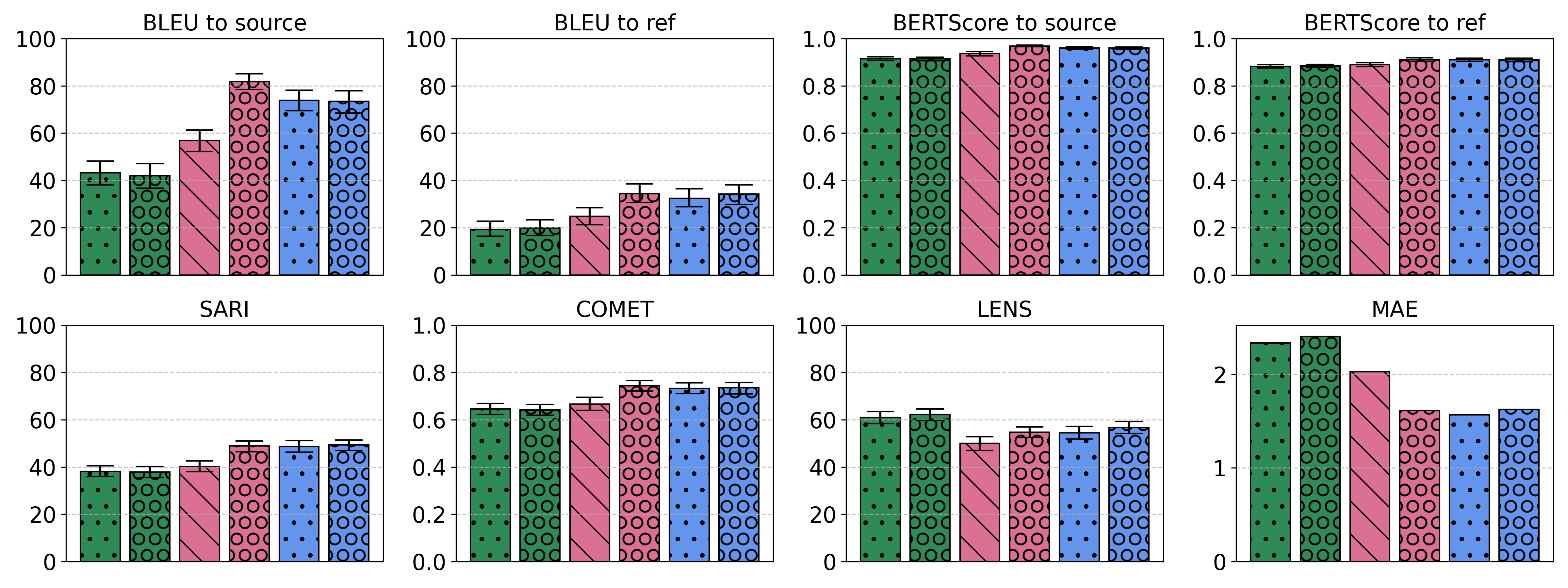}
    \end{subfigure}

    \begin{subfigure}[t]{0.8\textwidth}
        \centering
        \caption{FKGL}
        \includegraphics[width=\linewidth]{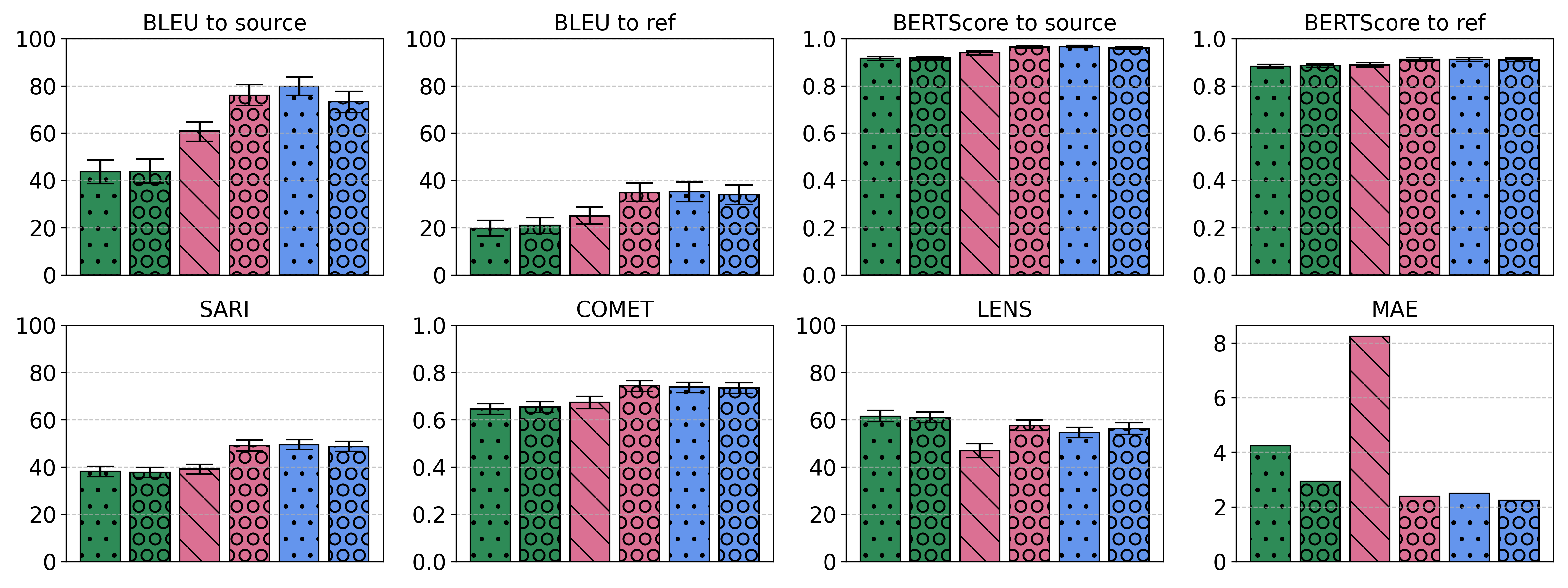}
    \end{subfigure}

    \begin{subfigure}[t]{0.8\textwidth}
        \centering
        \caption{char compression}
        \includegraphics[width=\linewidth]{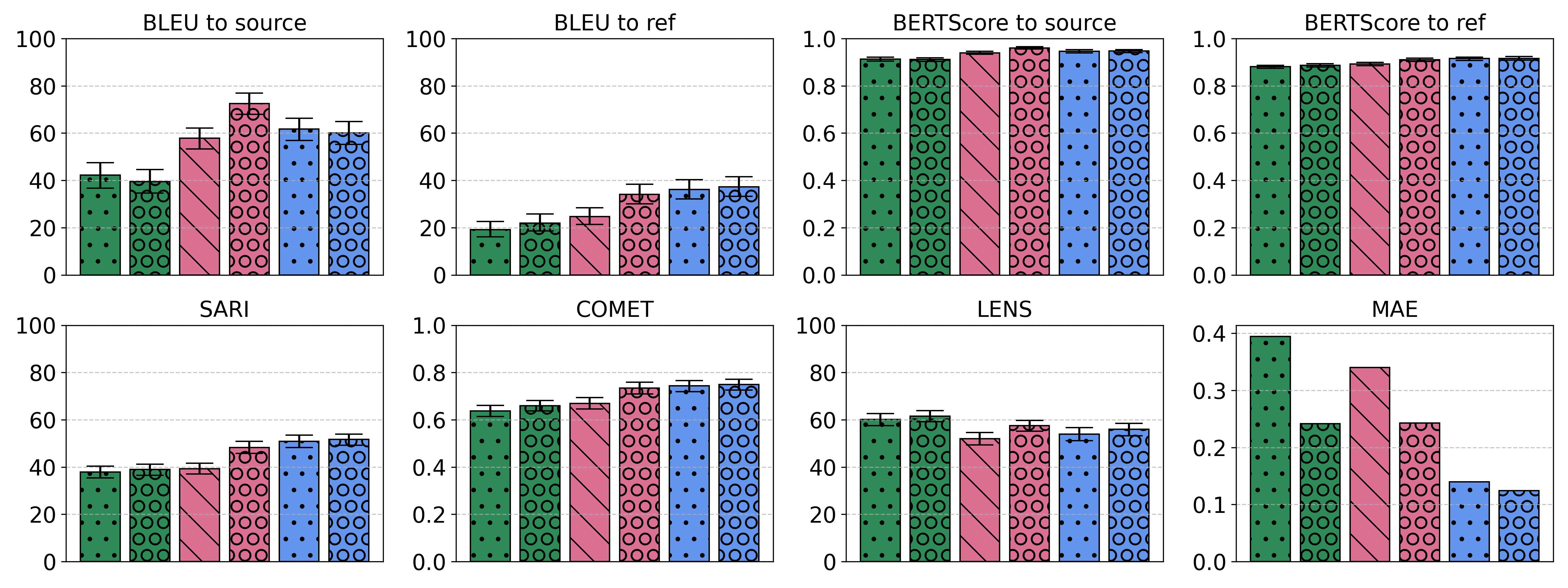}
    \end{subfigure}
    
    \begin{subfigure}[t]{0.8\textwidth}
        \centering
        \includegraphics[width=\linewidth]{assets/performance/CHAR_COMPRESSION_Med-EASi_token_explanation_legend.png}
    \end{subfigure}
    
\end{figure}

\clearpage
\section{Baseline Controllability Analysis}\label{sec:appendix-scatter}

\begin{figure}[!htb]
  \centering
  \caption{Dataset: \textsc{SimPA}}
  \begin{subfigure}[t]{0.48\columnwidth}
    \centering
    \includegraphics[width=\linewidth]{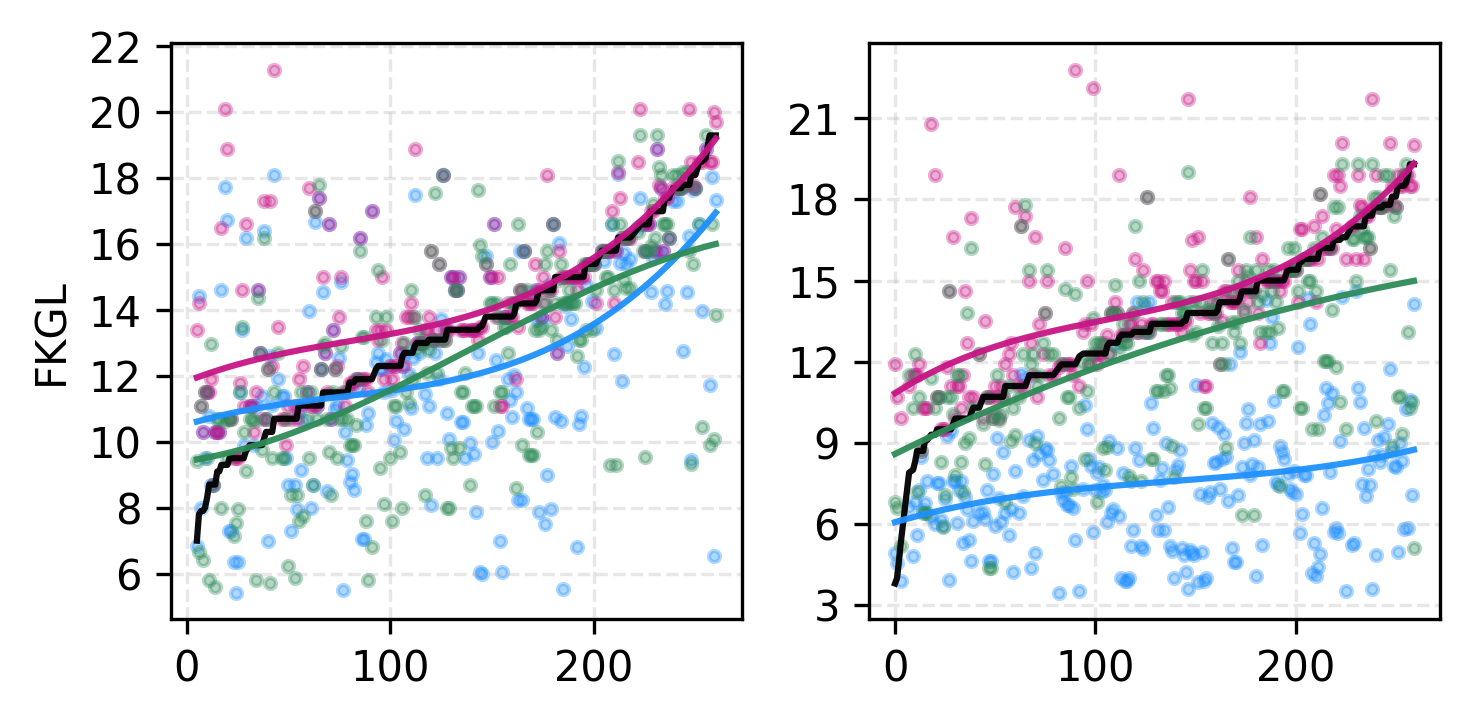}
  \end{subfigure}\hfill
  \begin{subfigure}[t]{0.48\columnwidth}
    \centering
    \includegraphics[width=\linewidth]{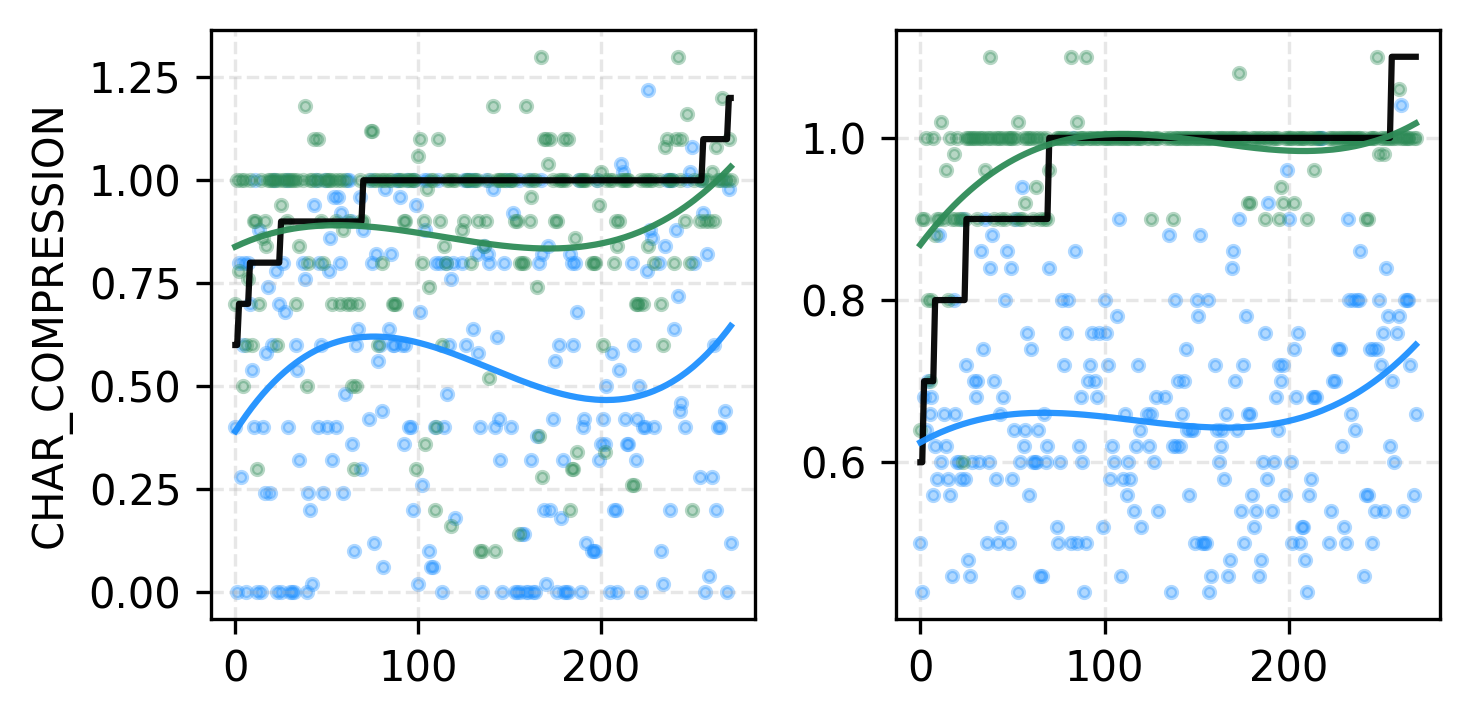}
  \end{subfigure}

    \caption{Dataset: \textsc{MedEASi}}
  \begin{subfigure}[t]{0.48\columnwidth}
    \centering
    \includegraphics[width=\linewidth]{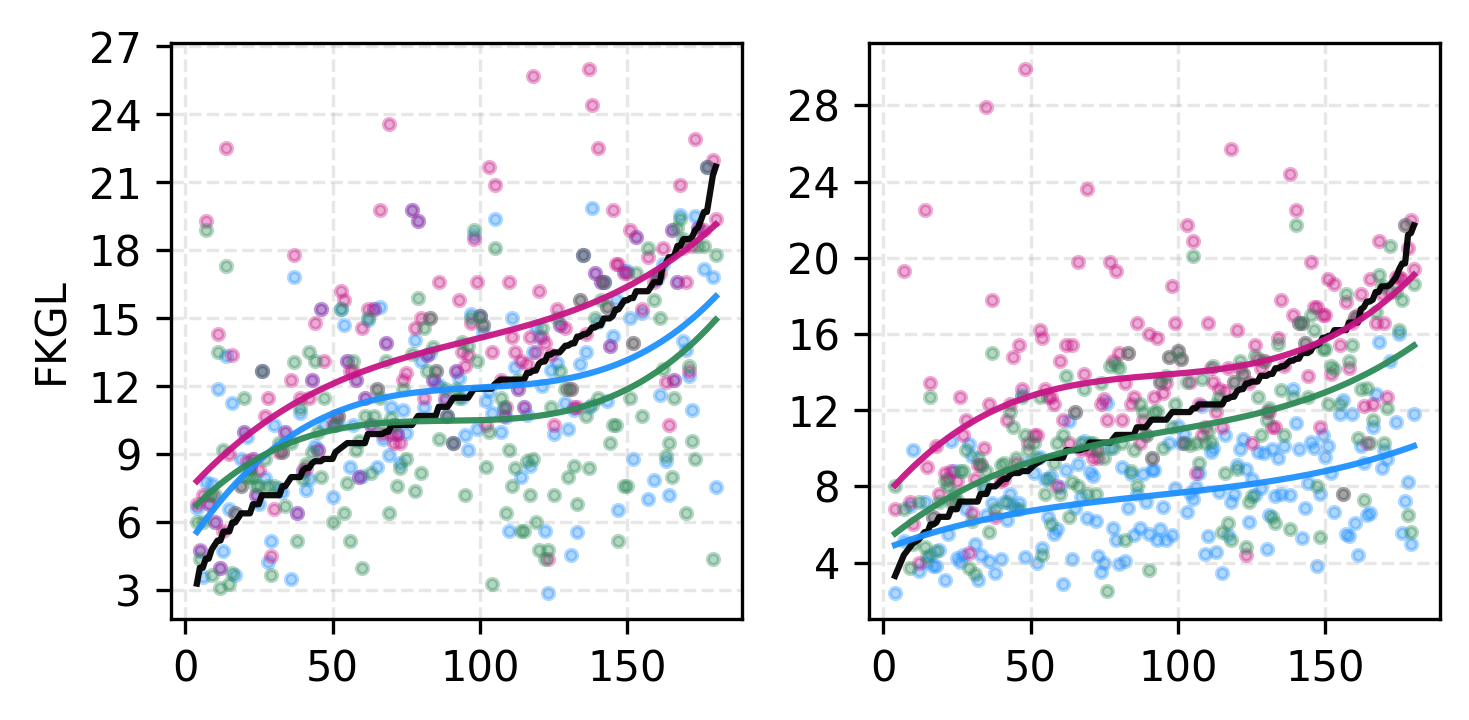}
  \end{subfigure}\hfill
  \begin{subfigure}[t]{0.48\columnwidth}
    \centering
    \includegraphics[width=\linewidth]{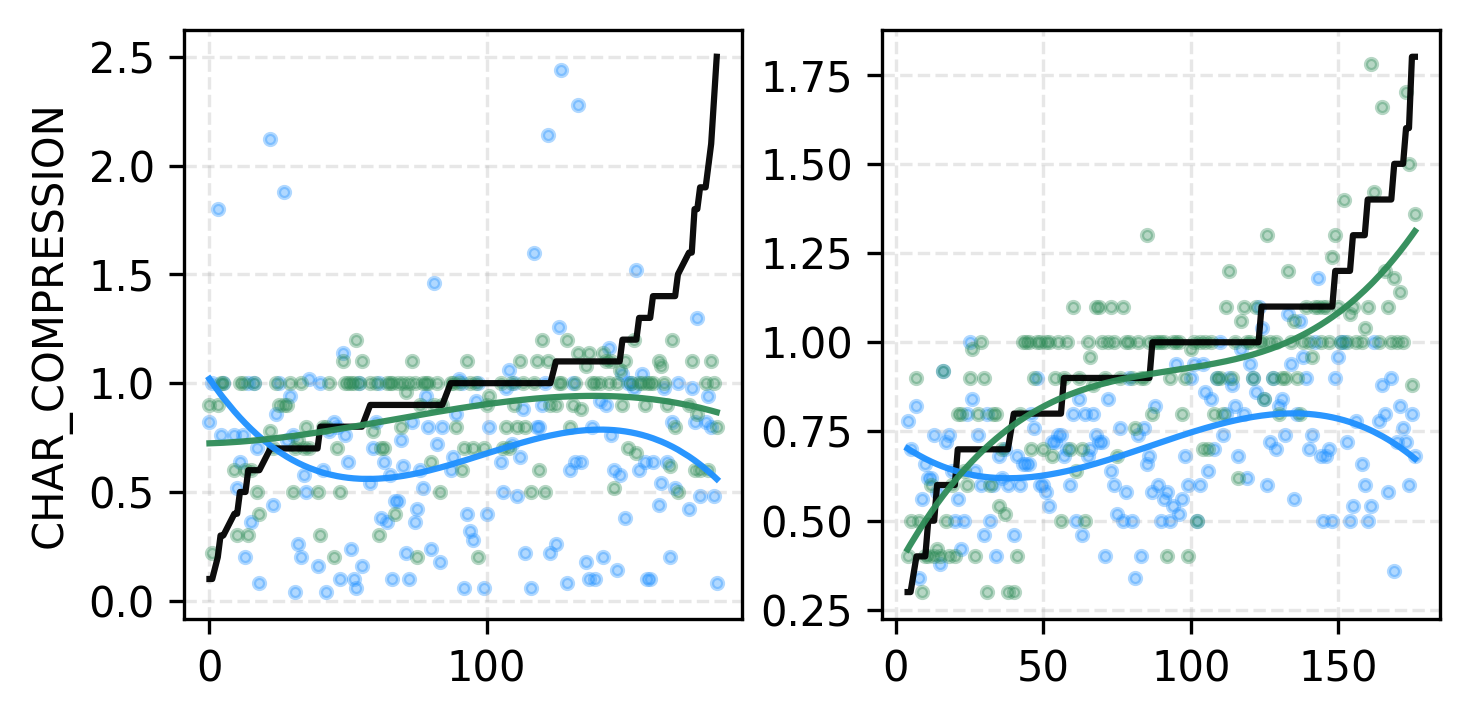}
  \end{subfigure}

  \caption{Dataset: \textsc{WikiLarge}}
  \begin{subfigure}[t]{0.48\columnwidth}
    \centering
    \includegraphics[width=\linewidth]{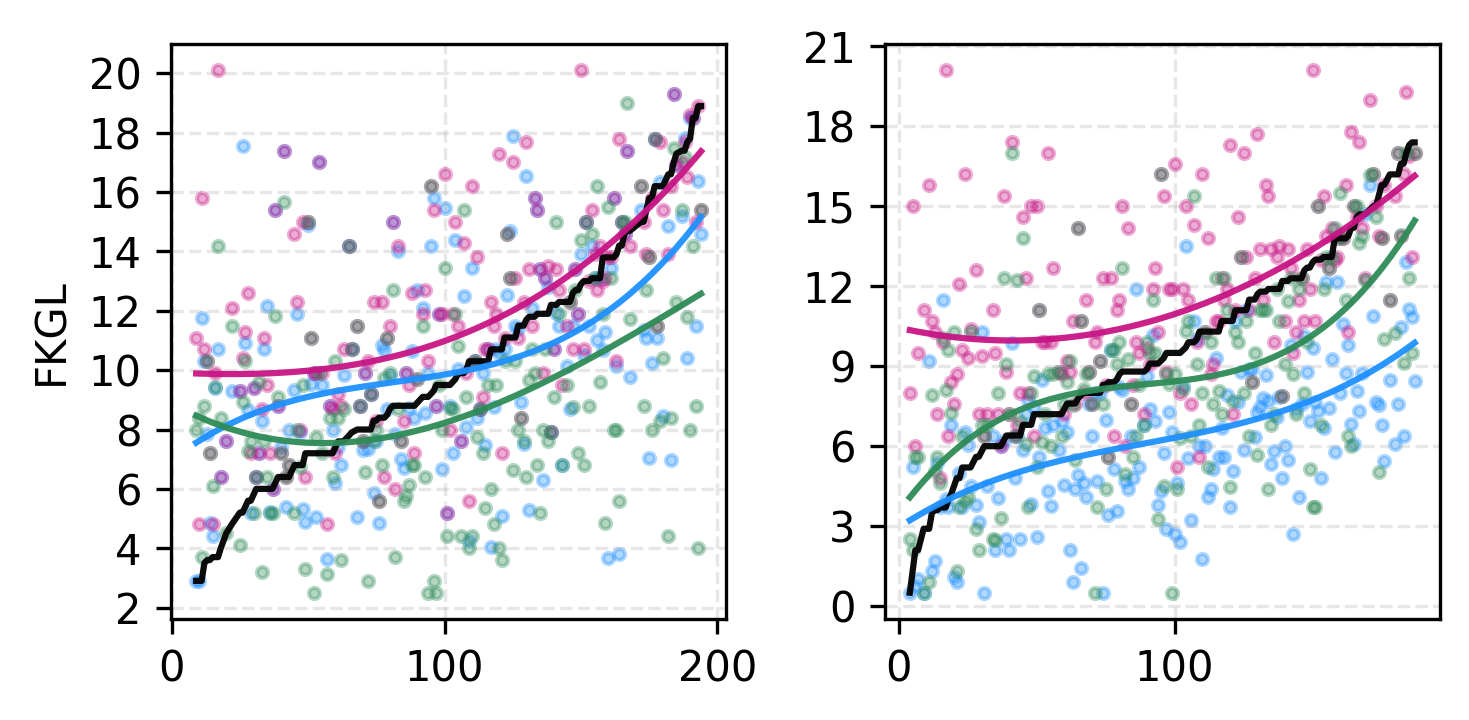}
  \end{subfigure}\hfill
  \begin{subfigure}[t]{0.48\columnwidth}
    \centering
    \includegraphics[width=\linewidth]{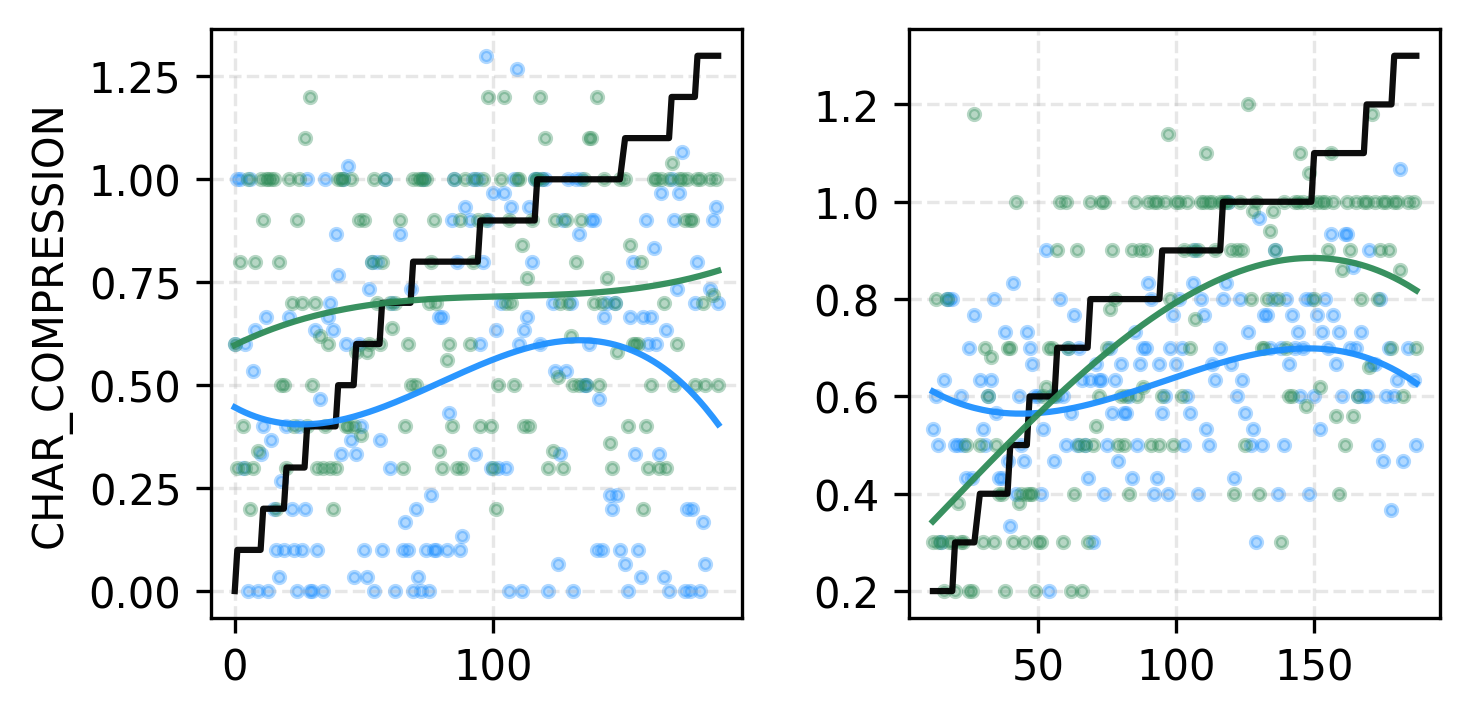}
  \end{subfigure}

  \caption{Dataset: \textsc{Newsela}}
  \begin{subfigure}[t]{0.48\columnwidth}
    \centering
    \includegraphics[width=\linewidth]{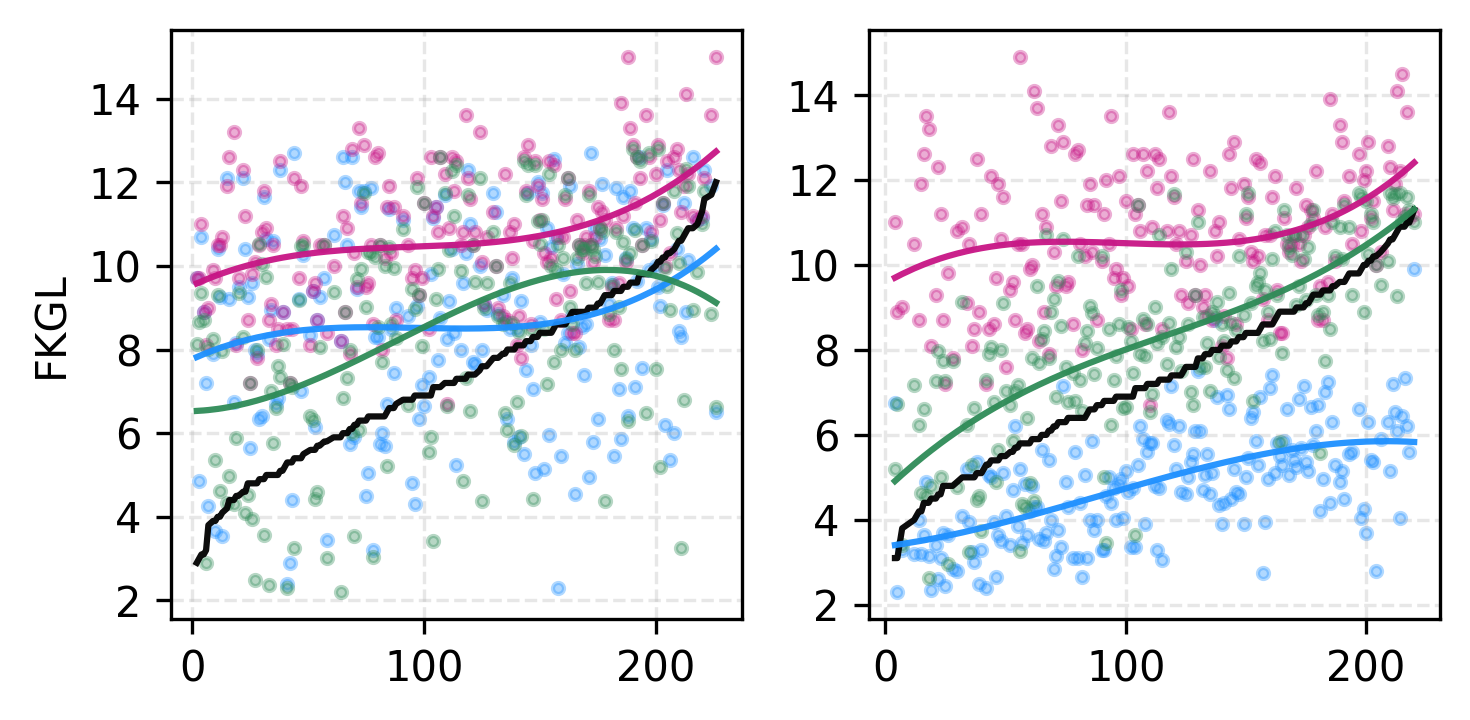}
  \end{subfigure}\hfill
  \begin{subfigure}[t]{0.48\columnwidth}
    \centering
    \includegraphics[width=\linewidth]{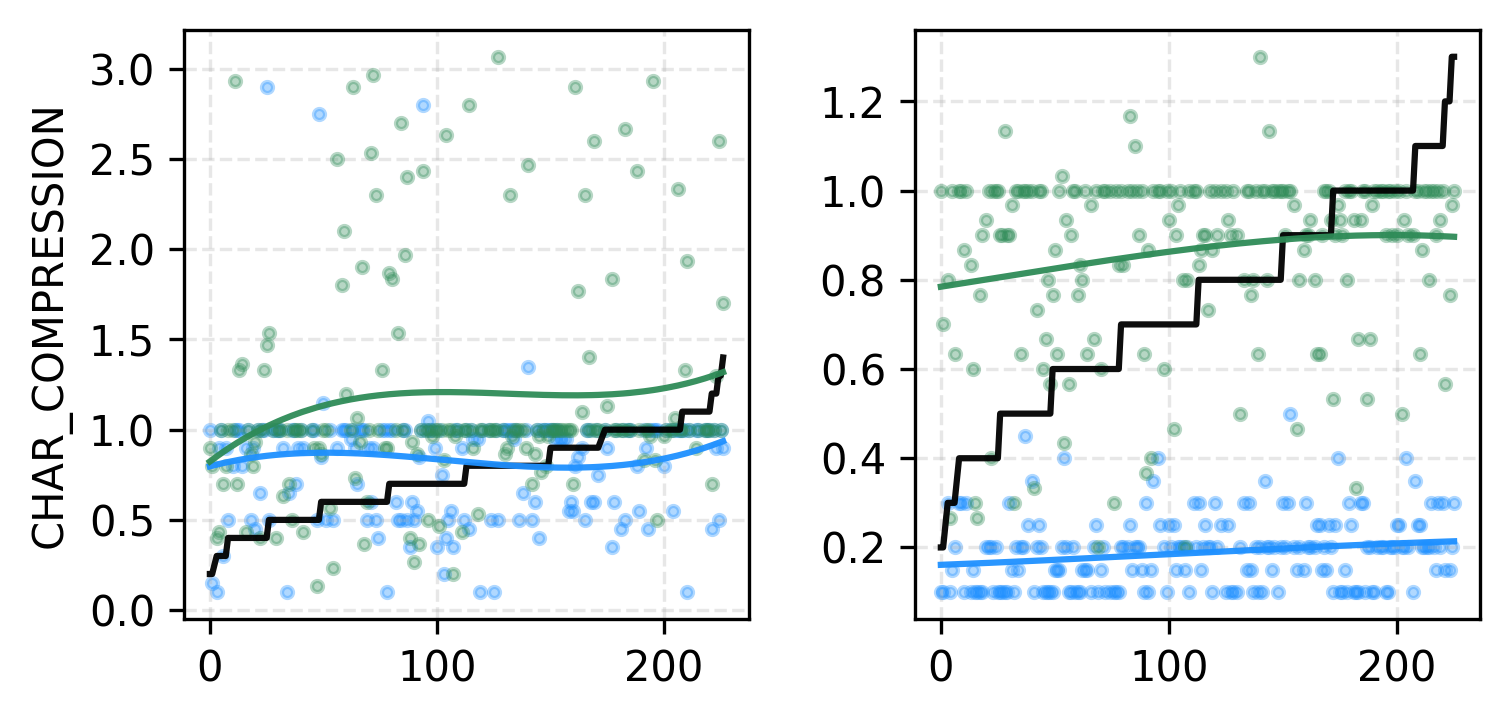}
  \end{subfigure}

  \begin{subfigure}[t]{\columnwidth}
    \centering
    \includegraphics[width=0.6\columnwidth]{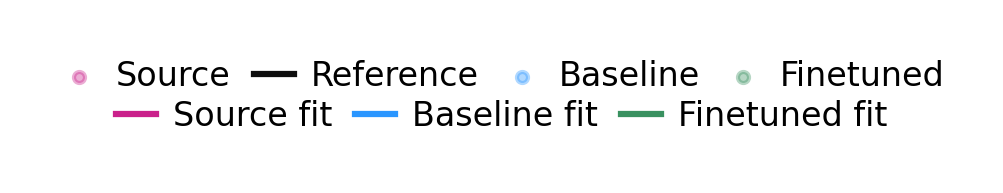}
  \end{subfigure}

  \caption{Comparison of controllability in terms of MAE between the instruction fine-tuned and non-fine-tuned models. For each control attribute: left plot is Llama-3.2-1B-Instruct, right is Llama-3.1-8B-Instruct.}
  \label{fig:control-medeasi}
\end{figure}

\clearpage
\section{Metric Correlation Analysis}\label{sec:appendix-heatmaps}

\begin{figure}[!htbp]
    \centering
    \caption{Pearson correlation heatmaps analyzing metric correlation, aggregated across all models. Dataset: \textsc{Newsela}.}
    \label{fig:correlation-heatmap-newsela}
    
    \begin{subfigure}[t]{0.33\linewidth}
    \centering
    \caption{FKGL}
    \includegraphics[width=\linewidth]{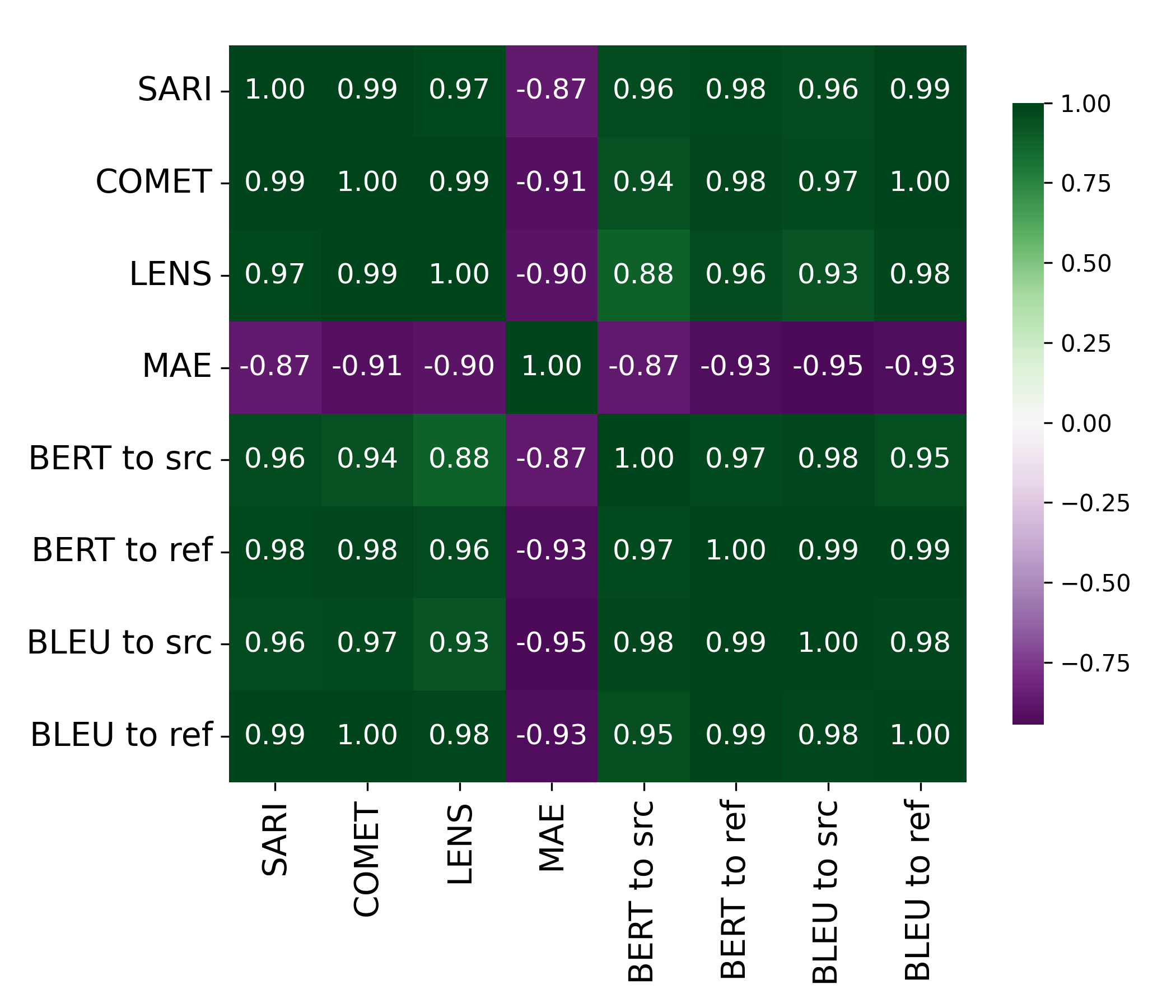}
    \end{subfigure}
    \begin{subfigure}[t]{0.33\linewidth}
    \centering
    \caption{char compression}
    \includegraphics[width=\linewidth]{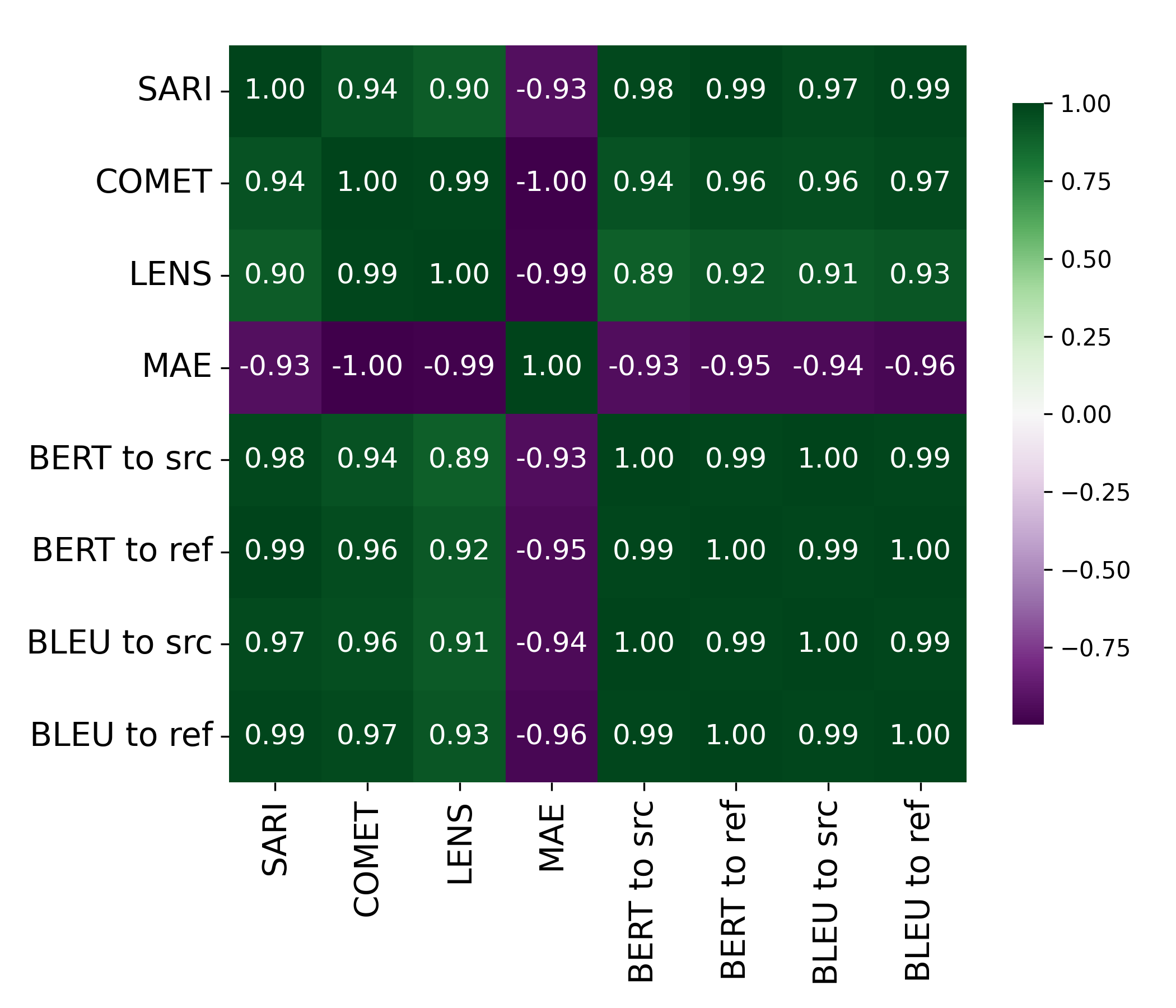} 

    \end{subfigure}

\end{figure}

\begin{figure}[!htbp]
    \centering
    \caption{Pearson correlation heatmaps analyzing metric correlation, aggregated across all models. Dataset: \textsc{Med-EASi}.}
    \label{fig:correlation-heatmap-medeasi}
    
    \begin{subfigure}[t]{0.32\linewidth}
    \centering
    \caption{FKGL}
    \includegraphics[width=\linewidth]{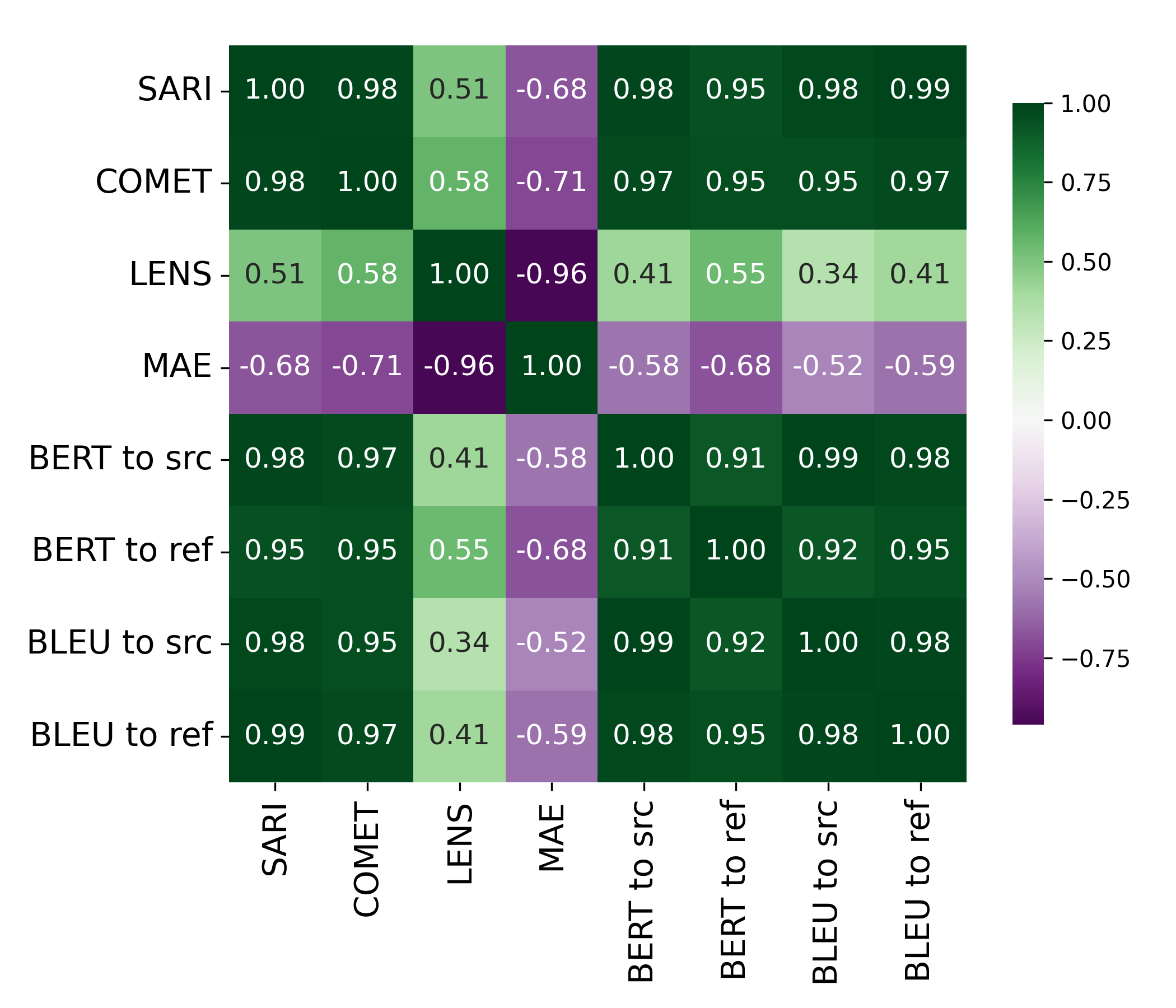}
    \end{subfigure}
    \hfill
    \begin{subfigure}[t]{0.32\textwidth}
    \centering
    \caption{Dale-Chall}
    \includegraphics[width=\linewidth]{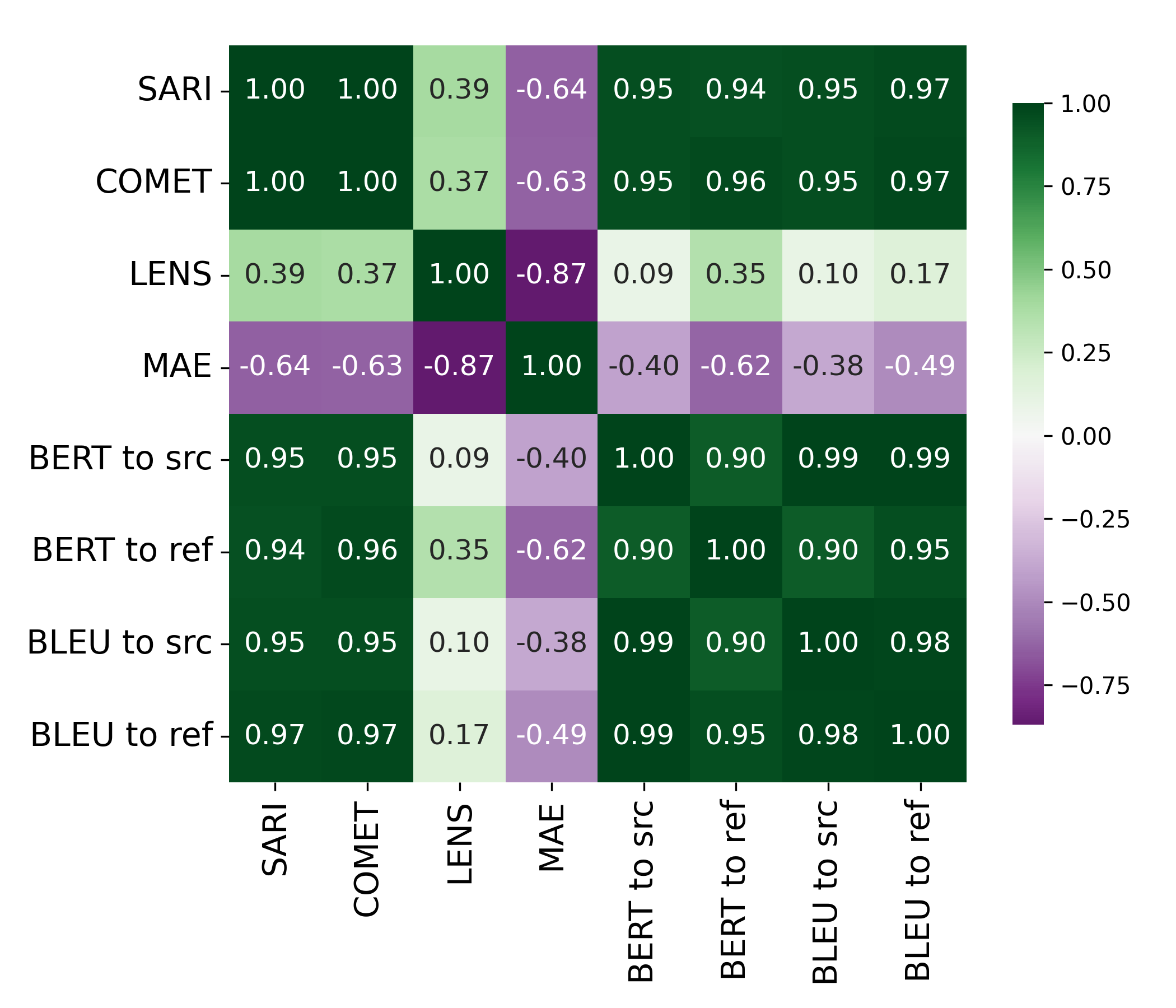} 
    \end{subfigure}
    \hfill
    \begin{subfigure}[t]{0.32\linewidth}
    \centering
    \caption{ARI}
    \includegraphics[width=\linewidth]{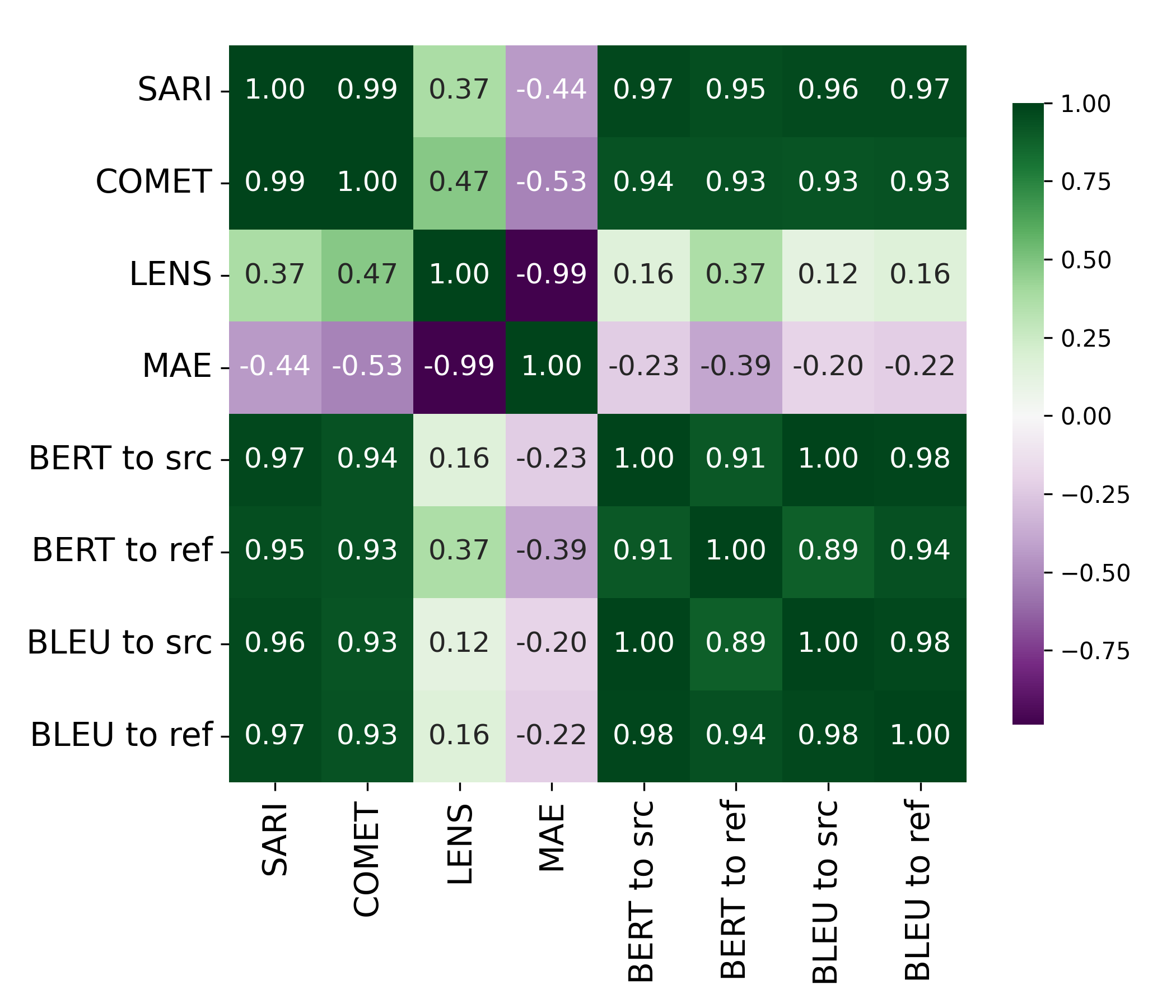}
    \end{subfigure}


    \begin{subfigure}[t]{0.32\linewidth}
    \centering
    \caption{char compression}
    \includegraphics[width=\linewidth]{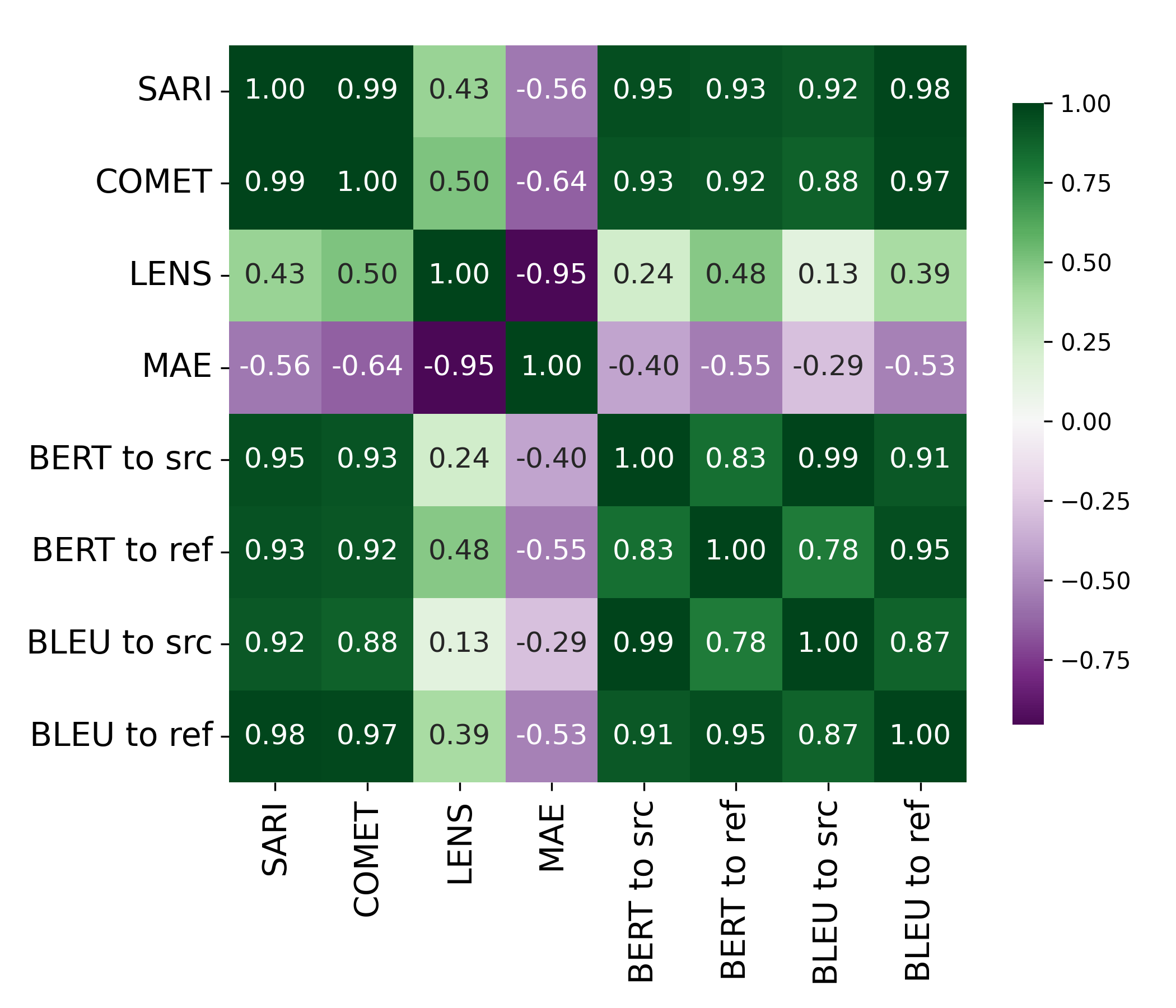} 
    \end{subfigure}
    \begin{subfigure}[t]{0.32\linewidth}
    \centering
    \caption{word compression}
    \includegraphics[width=\linewidth]{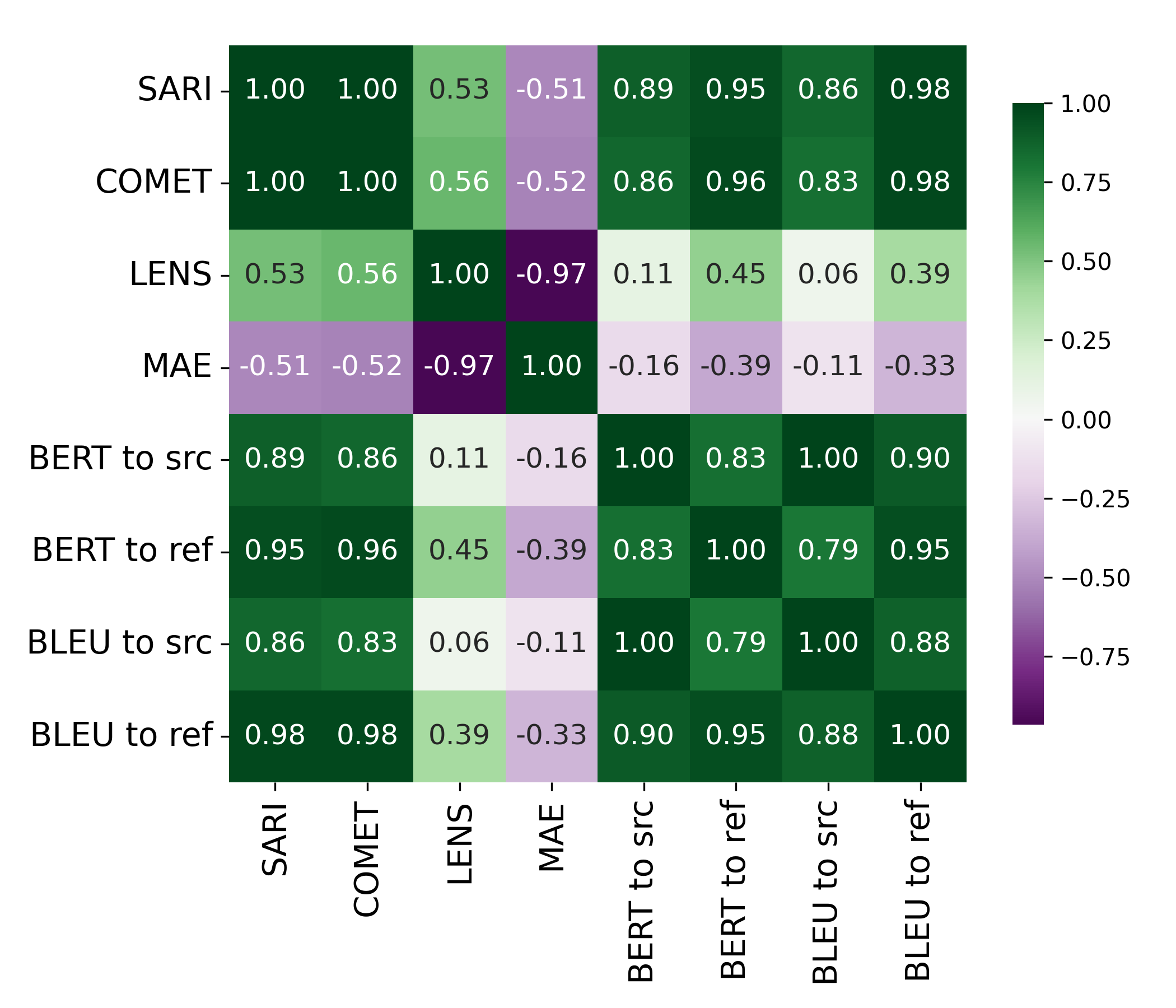}
    \hfill
    \end{subfigure}

\end{figure}

\begin{figure}[!htbp]
    \centering
    \caption{Pearson correlation heatmaps analyzing metric correlation, aggregated across all models. Dataset: \textsc{SimPA}.}
    \label{fig:correlation-heatmap-simpa}
    
    \begin{subfigure}[t]{0.32\linewidth}
    \centering
    \caption{FKGL}
    \includegraphics[width=\linewidth]{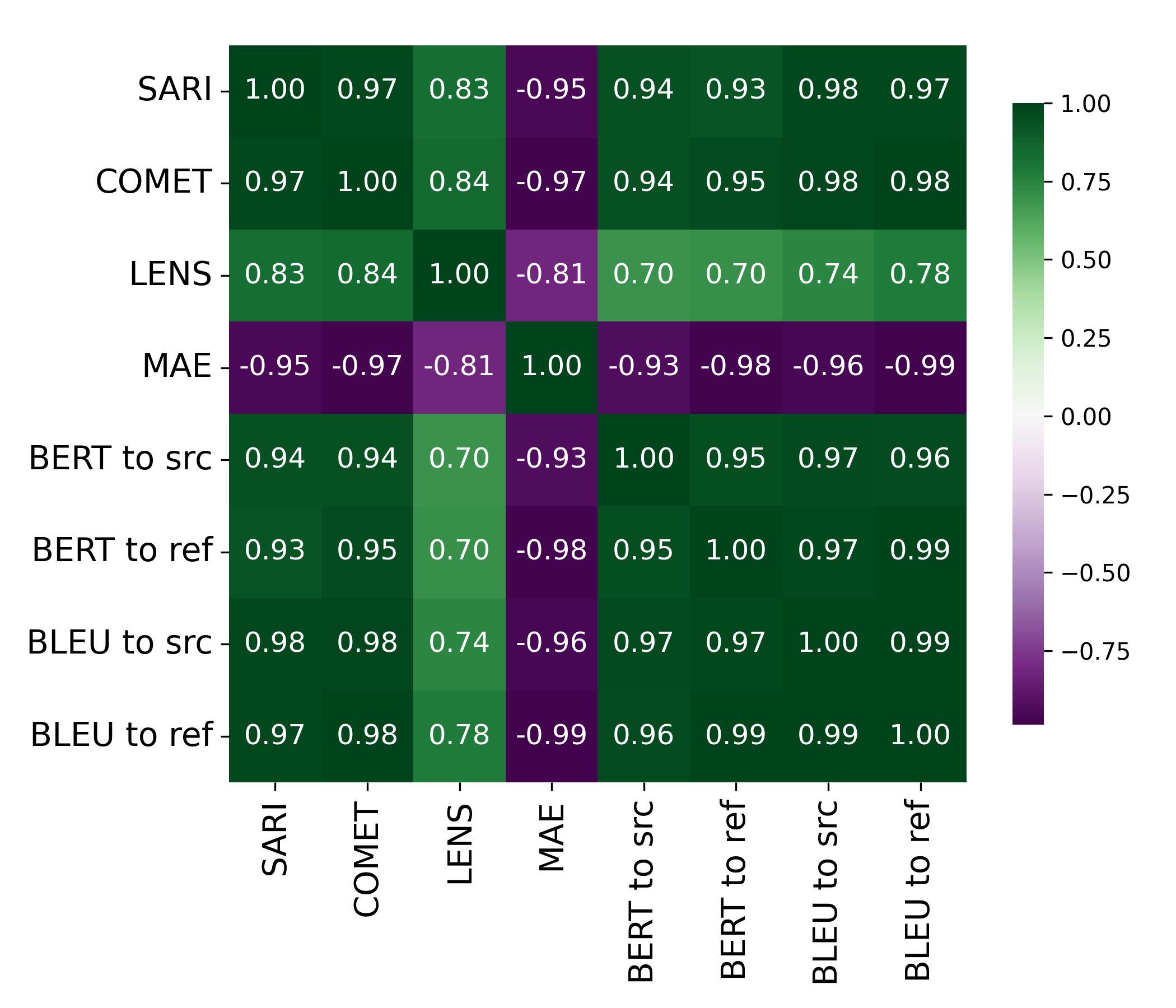}
    \end{subfigure}
    \hfill
    \begin{subfigure}[t]{0.32\textwidth}
    \centering
    \caption{Dale-Chall}
    \includegraphics[width=\linewidth]{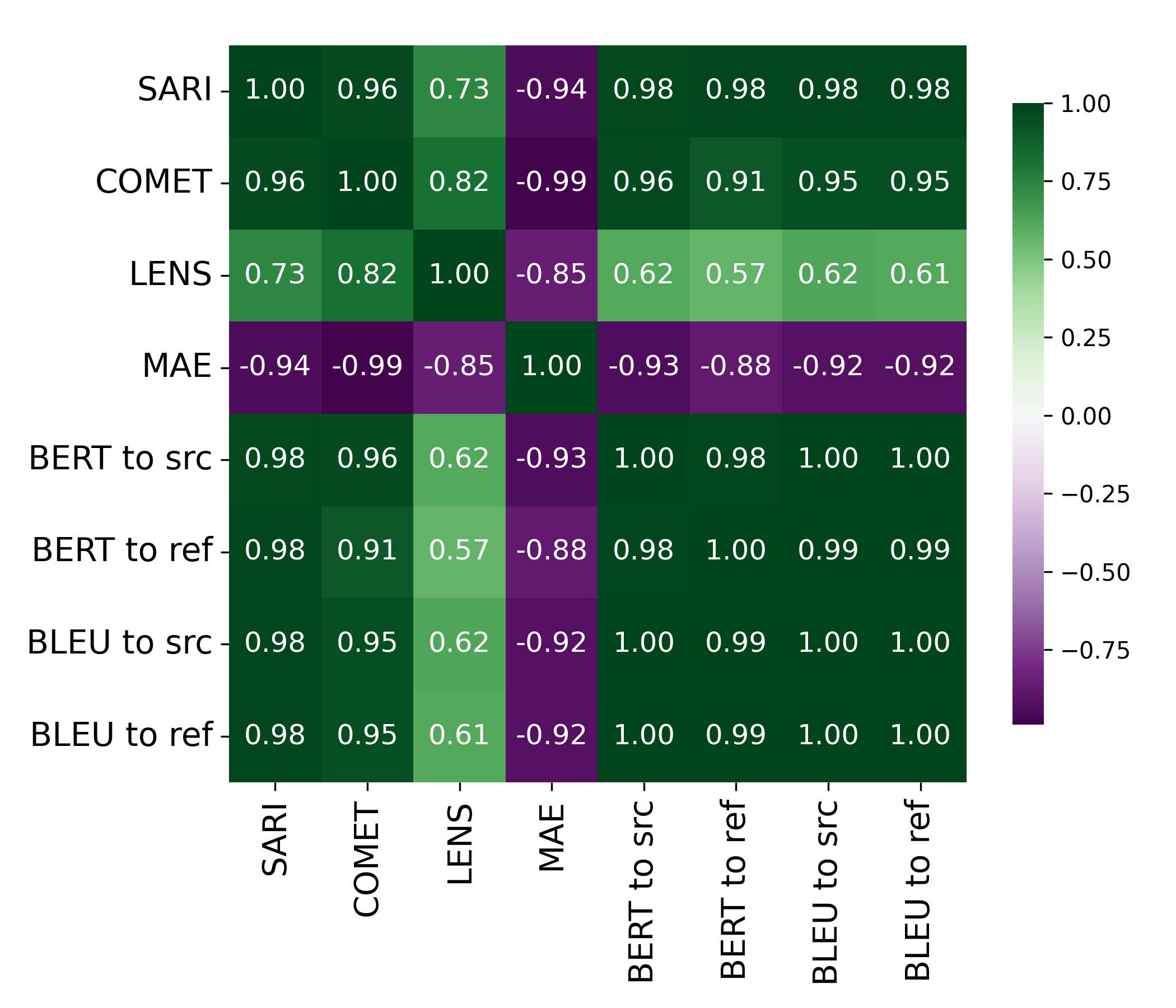} 
    \end{subfigure}
    \hfill
    \begin{subfigure}[t]{0.32\linewidth}
    \centering
    \caption{ARI}
    \includegraphics[width=\linewidth]{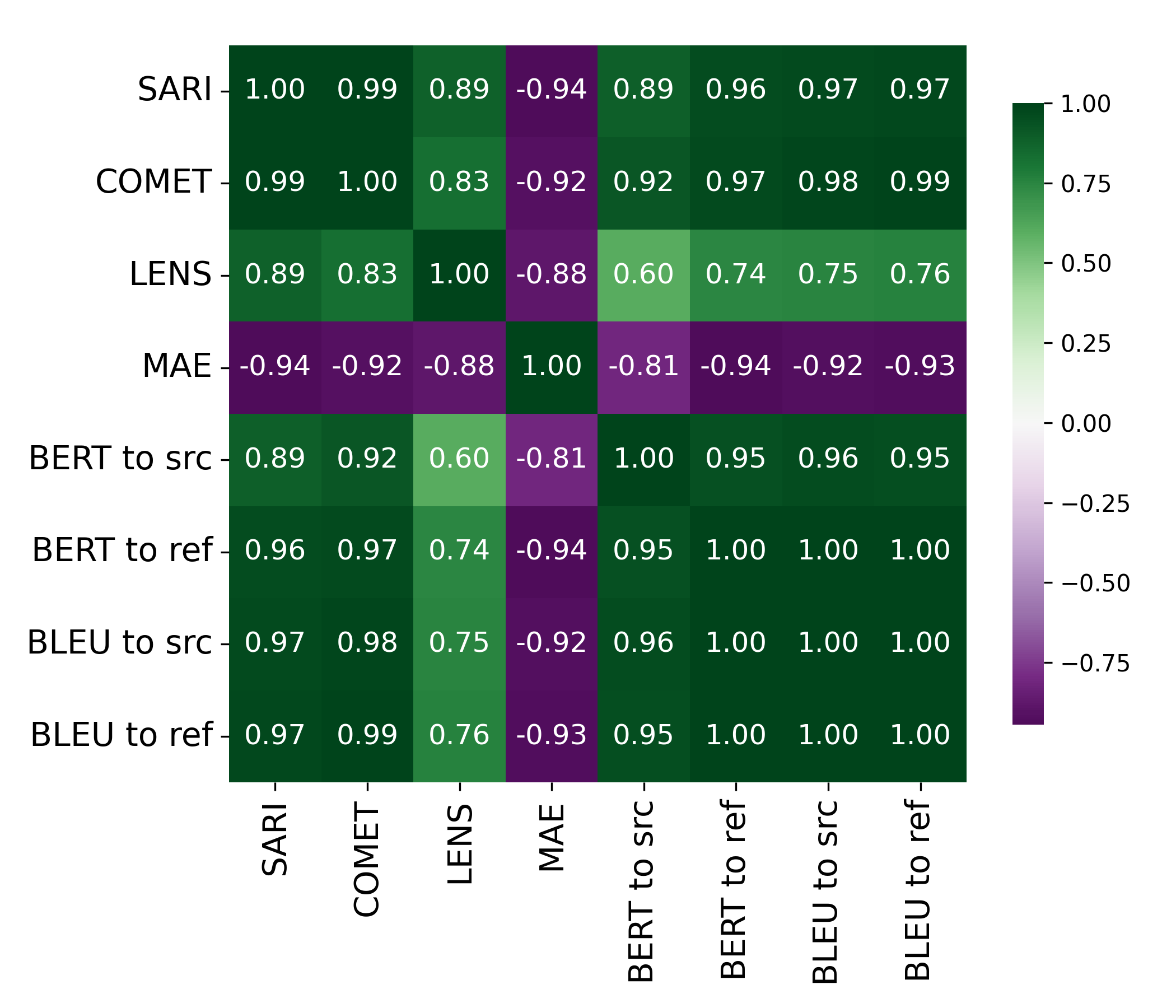}
    \end{subfigure}


    \begin{subfigure}[t]{0.32\linewidth}
    \centering
    \caption{char compression}
    \includegraphics[width=\linewidth]{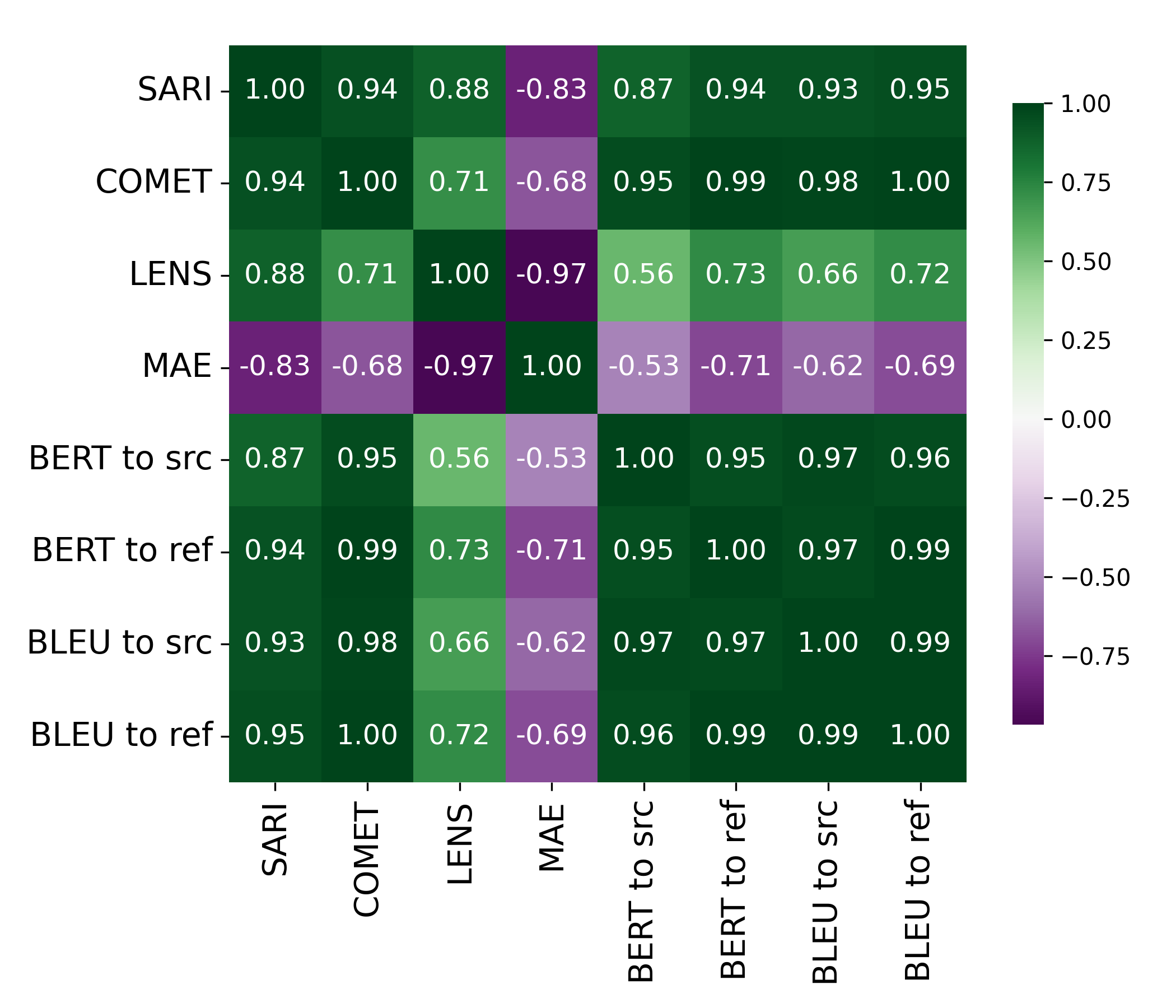} 
    \end{subfigure}
    \begin{subfigure}[t]{0.32\linewidth}
    \centering
    \caption{word compression}
    \includegraphics[width=\linewidth]{assets/heatmaps/WORD_COMPRESSION_Med-EASi_token_explanation_corr_heatmap.png}
    \hfill
    \end{subfigure}

\end{figure}

\begin{figure}[!htbp]
    \centering
    \caption{Pearson correlation heatmaps analyzing metric correlation, aggregated across all models. Dataset: \textsc{WikiLarge}.}
    \label{fig:correlation-heatmap-wikilarge}
    
    \begin{subfigure}[t]{0.32\linewidth}
    \centering
    \caption{FKGL}
    \includegraphics[width=\linewidth]{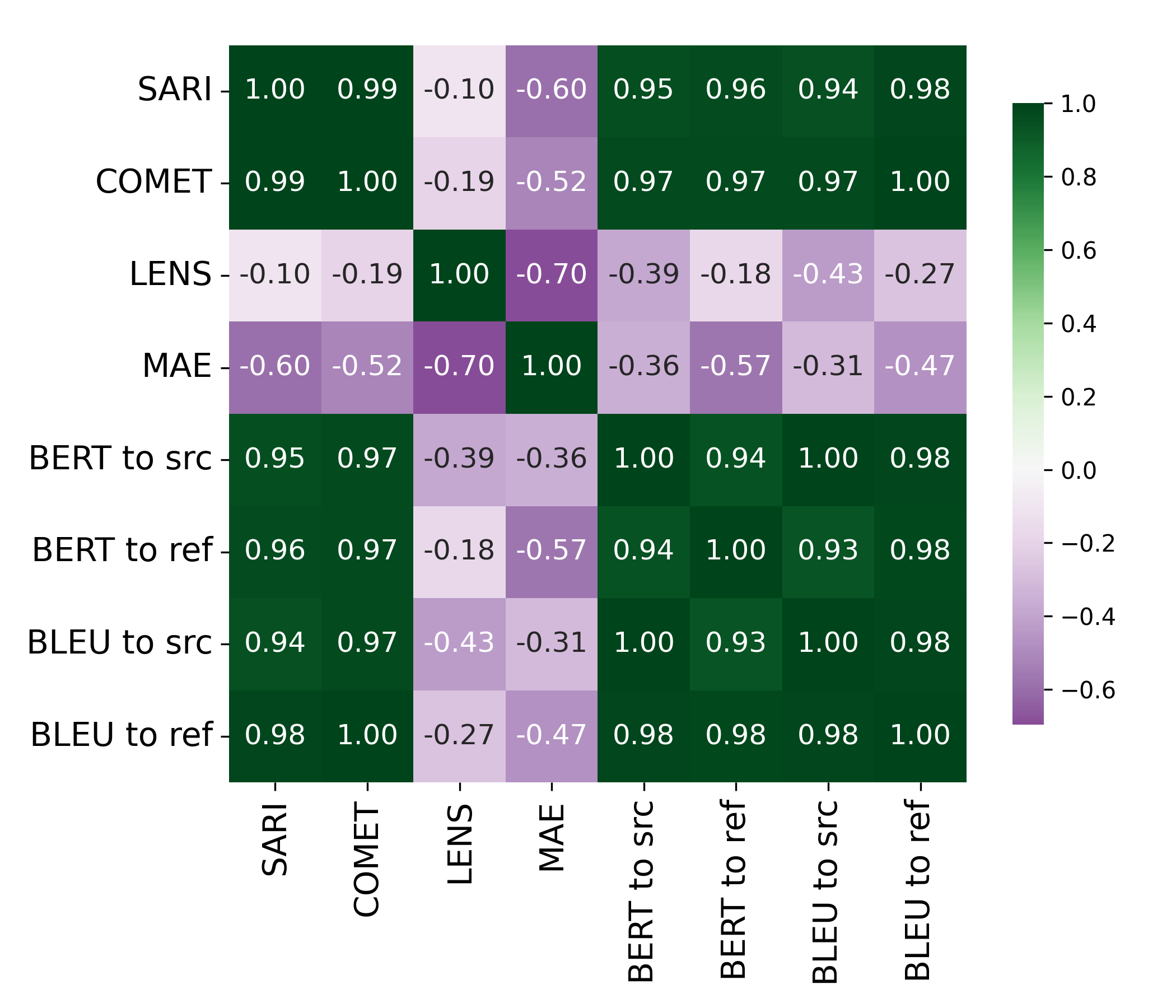}
    \end{subfigure}
    \hfill
    \begin{subfigure}[t]{0.32\textwidth}
    \centering
    \caption{Dale-Chall}
    \includegraphics[width=\linewidth]{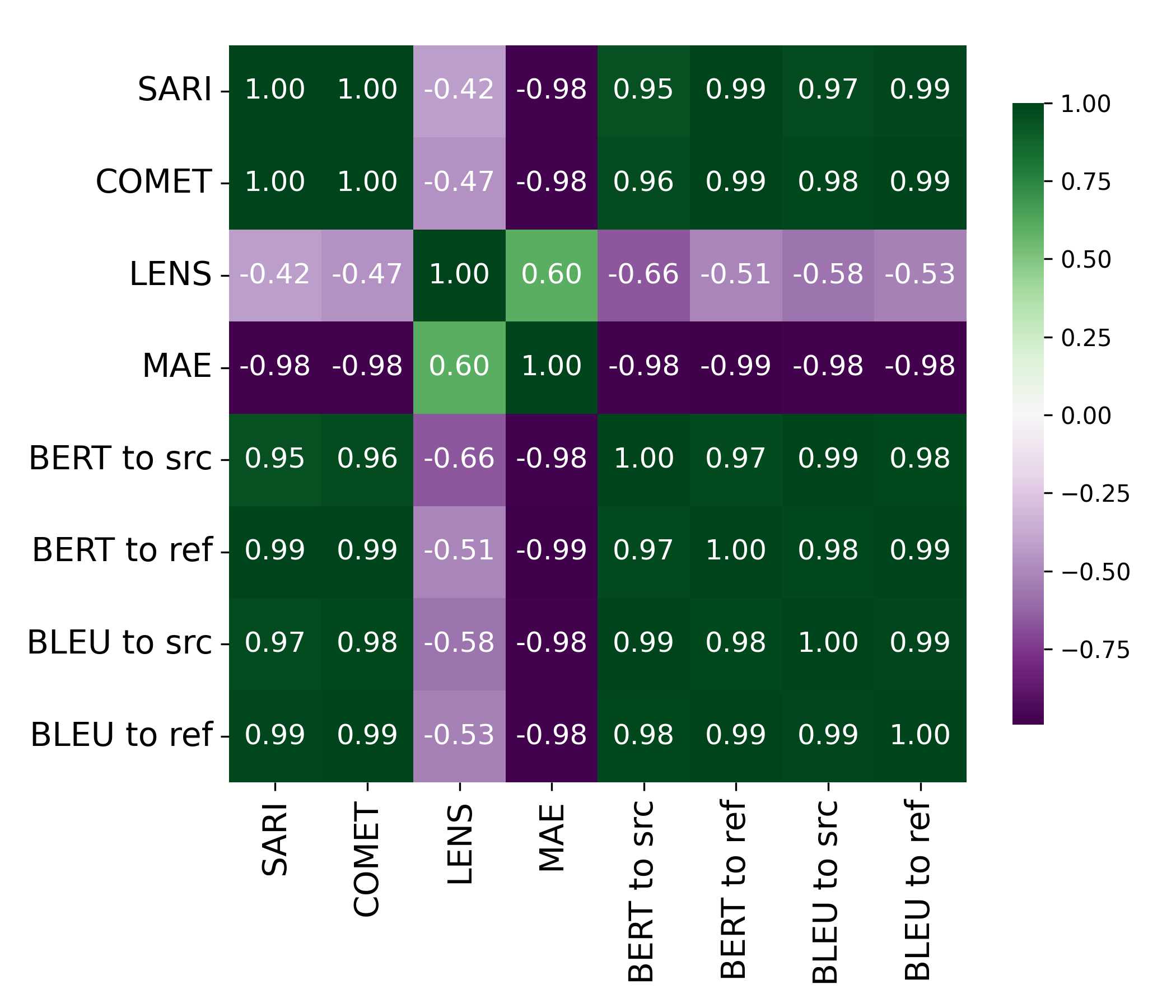} 
    \end{subfigure}
    \hfill
    \begin{subfigure}[t]{0.32\linewidth}
    \centering
    \caption{ARI}
    \includegraphics[width=\linewidth]{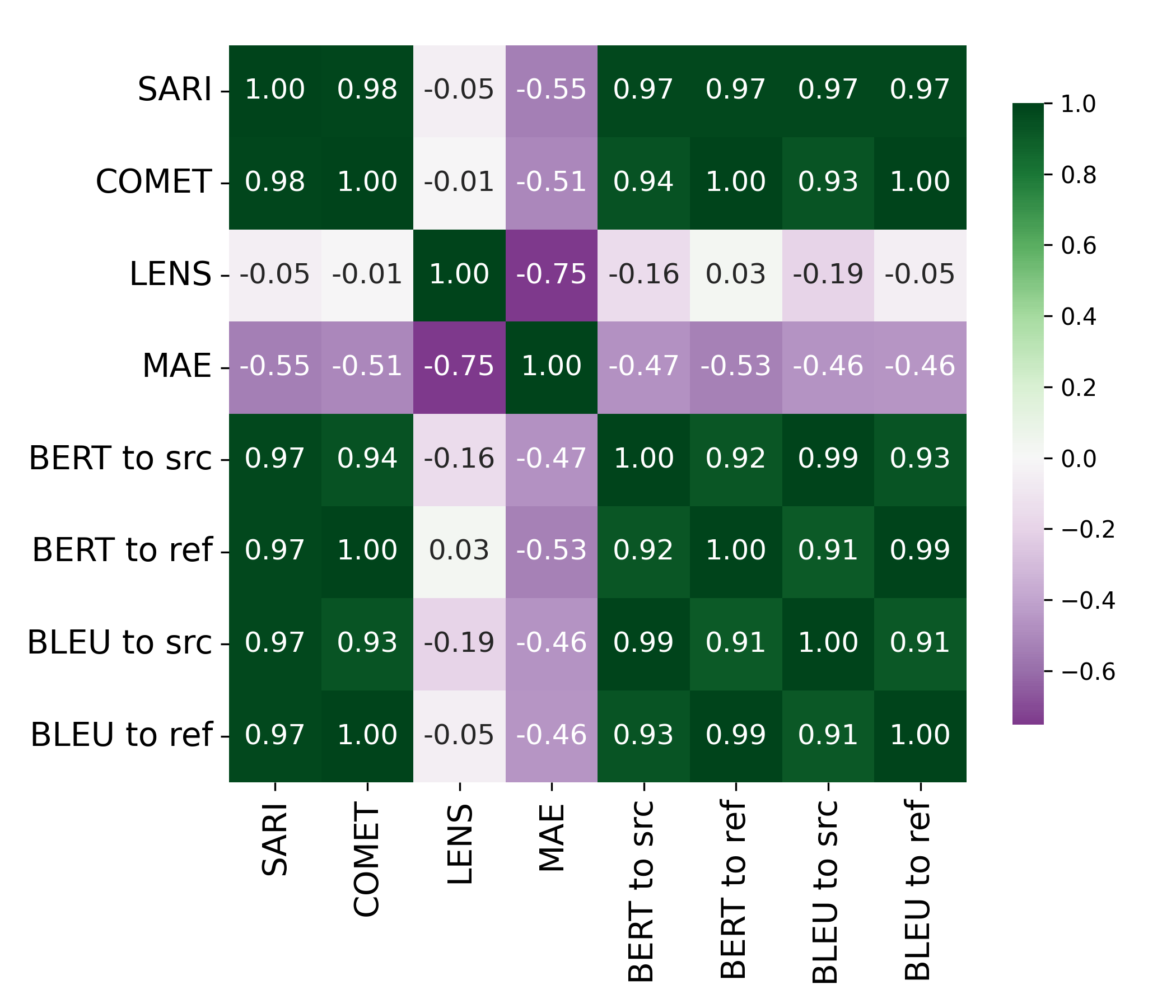}
    \end{subfigure}


    \begin{subfigure}[t]{0.32\linewidth}
    \centering
    \caption{char compression}
    \includegraphics[width=\linewidth]{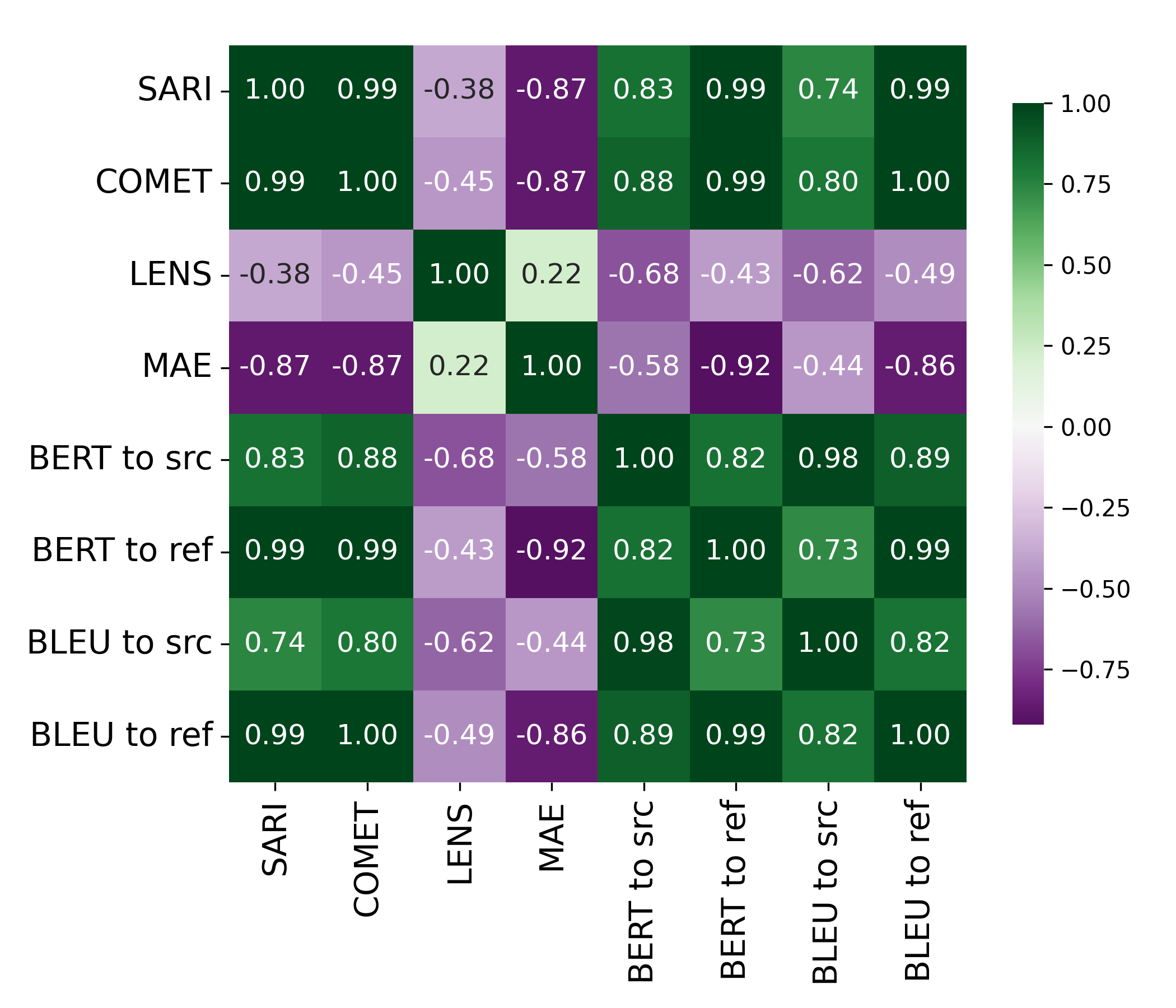} 
    \end{subfigure}
    \begin{subfigure}[t]{0.32\linewidth}
    \centering
    \caption{word compression}
    \includegraphics[width=\linewidth]{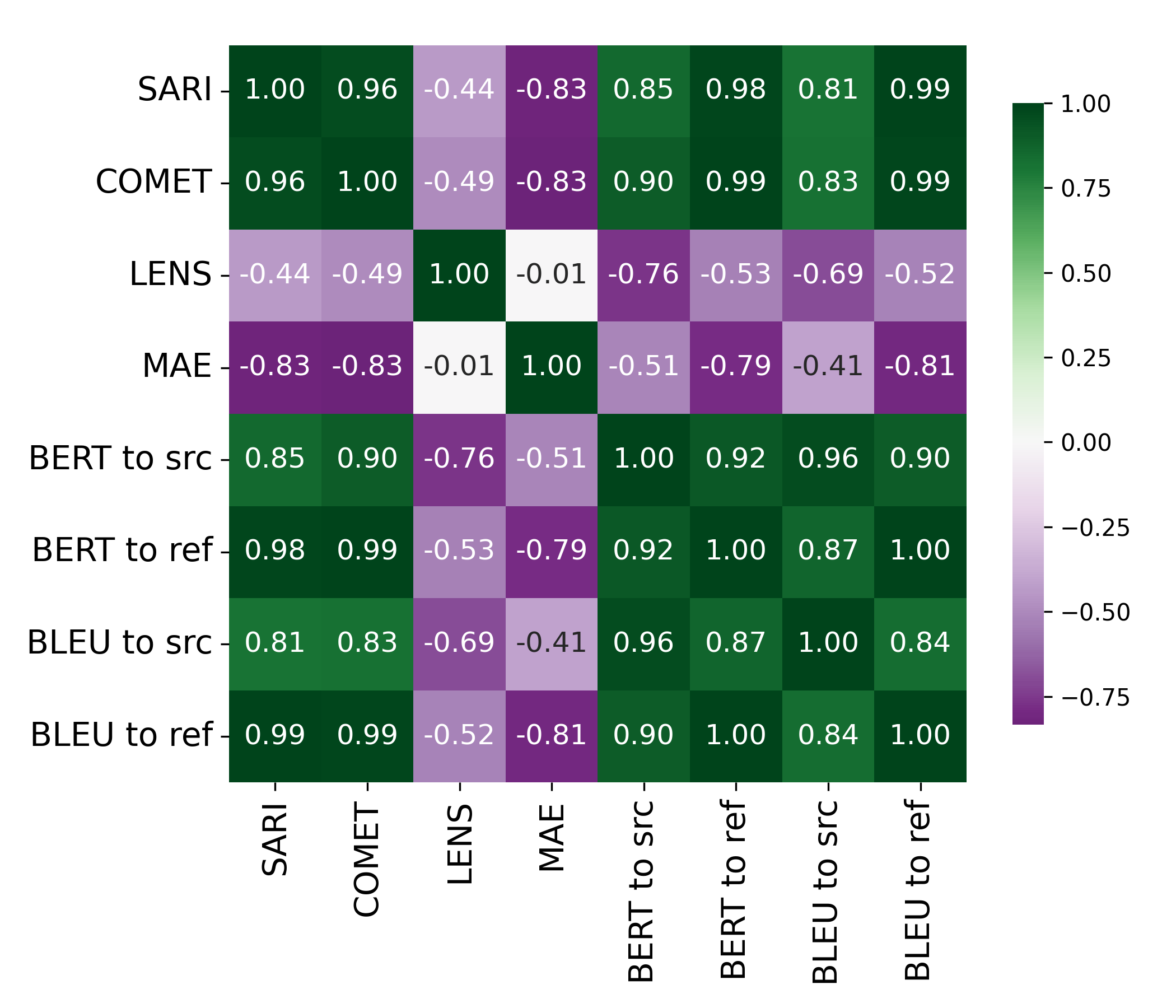}
    \hfill
    \end{subfigure}

\end{figure}

\clearpage
\section{Model Output Examples}\label{sec:appendix-output-examples}

\begin{table}[!htbp]
\footnotesize
\caption{Dataset: \textsc{WikiLarge}. Control Attribute: Dale-Chall. Instance ID: wikilarge\_ori\_252807. In this example, we see how different models focus on different information in the source text, resulting in very different simplifications. Further, reference simplification is much shorter than the source text and omits essential information. This is an example of a poor reference text, with lower adequacy and fluency than the automatically generated simplifications by various models.}
\label{tab:wiki-dc-example}
\begin{tabular}{llp{0.65\textwidth}}
\toprule
& \textbf{DC} & \textbf{Text}           \\ \hline
Source      & 9.5 & ``In public saunas it is more common that men and women go to the sauna separately, although people of both sexes may sometimes bathe together, for example in student clubs. Still, saunas are not associated with sex and sexuality''. \\ 

Reference                & \textbf{5.7} & ``Some saunas are only for men , or only for women''. \\ 
\hline
Llama-3.2-1B-Instruct & 11.1  & ``The sauna is not associated with sex and sexuality''.                   \\ 
Llama-3.1-8B-Instruct &  7.3 & ``The sauna is a place where people go to relax''.                   \\ 
Ministral-3b-instruct & 9.3  & ``In public saunas it is more common that men and women go to the sauna separately, although people of both sexes may occasionally bathe together''. \\
Mistral-7B-Instruct-v0.1 & 9.8 & ``In public saunas it is more common that men and women go to the sauna separately, although people of both sexes may sometimes bathe together, for example in student clubs''. \\
Qwen3-1.7B            & 11.9 & ``Saunas are not associated with sex and sexuality''. \\ 
Qwen3-8B   &  8.4    &  ``In public saunas it is more common that men and women go to the sauna separately''. \\ 
\bottomrule
\end{tabular}
\end{table}

\begin{table}[!htbp]
\footnotesize
\caption{Dataset: \textsc{Med-EASi}. Control Attribute: ARI. Instance ID: medeasi\_000434. Simplifications generated by the model are very similar, further highlighting the effect of sentence splitting on the ARI score. Llama-3.2-1B-Instruct and Ministral-3b-instruct split the long sentence into two and drastically reduce the ARI score without any further transformations. }
\label{tab:medeasi-ari-example}
\begin{tabular}{llp{0.65\textwidth}}
\toprule
& \textbf{ARI} & \textbf{Text}                   \\ \hline
Source                & 20.1 & ``Symptoms include a sense of fullness, heaviness, aching, cramps, pain, tiredness, and paresthesias in the legs; these symptoms worsen with standing or walking and are relieved by rest and elevation''. \\ 
Reference                & \textbf{15.2} & ``The swelling may not cause any symptoms, but some people feel fullness, heaviness, aching, cramps, pain, tiredness, and tingling in the legs''. \\ 
\midrule
Llama-3.2-1B-Instruct &  11.8 & ``The symptoms include a sense of fullness, heaviness, aching, cramps, pain, tiredness, and numbness in the legs. These symptoms worsen with standing or walking and are relieved by rest and elevation''.                   \\ 
Llama-3.1-8B-Instruct &  19.6 & ``The symptoms include a sense of fullness, heaviness, aching, cramps, pain, tiredness, and numbness in the legs; these symptoms worsen with standing or walking and are relieved by rest and elevation''.                   \\ 
Ministral-3b-instruct &  10.7 & ``Symptoms include a sense of fullness, heavyness, aching, cramps, and tiredness. They worsen with standing or walking and are relieved by rest and elevation''. \\
Mistral-7B-Instruct-v0.1 & 19.5 & ``Symptoms include a sense of fullness, heaviness, aching, cramps, pain, tiredness, and tingling in the legs; these symptoms worsen with standing or walking and are relieved by rest and elevation''. \\
Qwen3-1.7B            & 20.1 & ``Symptoms include a sense of fullness, heaviness, aching, cramps, pain, tiredness, and paresthesias in the legs; these symptoms worsen with standing or walking and are relieved by rest and elevation''. \\ 
Qwen3-8B              &   19.5   &  ``Symptoms include a sense of fullness, heaviness, aching, cramps, pain, tiredness, and tingling in the legs; these symptoms worsen with standing or walking and are relieved by rest and elevation''.                                                                                          \\ 
\bottomrule
\end{tabular}
\end{table}

\clearpage
\section{Prompt Template}\label{sec:appendix-prompt}

\begin{lstlisting}[
  basicstyle=\ttfamily\scriptsize,
  breaklines=true,
  columns=fullflexible,
  keepspaces=true,
  caption={Control-token-specific instructions dynamically inserted into the prompt.},
  label={lst:user-instructions}
  ][!htbp]

  "ARI": 
    {"prompt": "INSTRUCTION: Simplify the following text such that its Automated Readability Index (ARI) score is approximately equal to that specified in the control token prepended to your generated simplification. The control token has the following format: <METRIC=VALUE>. The control token prepended to the source text indicates the ARI value of the source text. \nSOURCE TEXT: <ARI={SOURCE_VALUE}> {TEXT}\nEXPLANATION: {EXPLANATION}"}
  "FKGL":
    {"prompt": "INSTRUCTION: Simplify the following text such that its Flesch-Kincaid Grade Level (FKGL) score is approximately equal to that specified in the control token prepended to your generated simplification. The control token has the following format: <METRIC=VALUE>. The control token prepended to the source text indicates the FKGL value of the source text. \nSOURCE TEXT: <FKGL={SOURCE_VALUE}> {TEXT}\nEXPLANATION: {EXPLANATION}"}
  "DALE-CHALL": 
    {"prompt": "INSTRUCTION: Simplify the following text such that its Dale-Chall Readability Score is approximately equal to that specified in the control token prepended to your generated simplification. The control token has the following format: <METRIC=VALUE>. The control token prepended to the source text indicates the Dale-Chall value of the source text. \nSOURCE TEXT: <DALE-CHALL={SOURCE_VALUE}> {TEXT}\nEXPLANATION: {EXPLANATION}"}
  "CHAR_COMPRESSION": 
    {"prompt": "INSTRUCTION: Simplify the following text such that the length of the output text (number of characters) relative to the source text is approximately equal to the ratio specified in the control token prepended to your generated simplification. The control token has the following format: <METRIC=VALUE>. \nSOURCE TEXT: {TEXT}\nEXPLANATION: {EXPLANATION}"}
  "WORD_COMPRESSION": 
    {"prompt": "INSTRUCTION: Simplify the following text such that the length of the output text (number of words) relative to the source text is approximately equal to the ratio specified in the control token prepended to your generated simplification. The control token has the following format: <METRIC=VALUE>. \nSOURCE TEXT: {TEXT}\nEXPLANATION: {EXPLANATION}"}

\end{lstlisting}

\begin{lstlisting}[
  basicstyle=\ttfamily\scriptsize,
  breaklines=true,
  columns=fullflexible,
  keepspaces=true,
  caption={Control-token-specific explanations dynamically inserted into the prompt.},
  label={lst:user-promt-expanation}
  ][!htbp]

    "ARI": {
        "token": "<ARI={ARI_VALUE}>",
        "description": "Automated Readability Index",
        "value type": "float",
        "explanation": "The <ARI={TARGET_VALUE}> token specifies that the target Automated Readability Index (ARI) score should be approximately {TARGET_VALUE}. Lower values indicate simpler text."
    },
    "FKGL": {
        "token": "<FKGL={FKGL_VALUE}>",
        "description": "Flesch-Kincaid Grade Level",
        "value type": "float",
        "explanation": "The <FKGL={TARGET_VALUE}> token specifies that the target Flesch-Kincaid Grade Level should be approximately {TARGET_VALUE}. Lower values indicate simpler text."
    },
    "DALE-CHALL": {
        "token": "<DALE-CHALL={DALE_CHALL_VALUE}>",
        "description": "Dale-Chall Readability Score",
        "value type": "float",
        "explanation": "The <DALE-CHALL={TARGET_VALUE}> token specifies that the target Dale-Chall readability score should be approximately {TARGET_VALUE}. Lower values indicate simpler text."
    },
    "CHAR_COMPRESSION": {
        "token": "<CHAR_COMPRESSION={CHAR_COMPRESSION_VALUE}>",
        "description": "Character-level Compression Rate",
        "value type": "float",
        "explanation": "The <CHAR_COMPRESSION={TARGET_VALUE}> token specifies that the character-level compression rate should be around {TARGET_VALUE}. Values less than 1 indicate text reduction. Values greater than 1 indicate text expansion."
    },
    "WORD_COMPRESSION": {
        "token": "<WORD_COMPRESSION={WORD_COMPRESSION_VALUE}>",
        "description": "Word-level Compression Rate",
        "value type": "float",
        "explanation": "The <WORD_COMPRESSION={TARGET_VALUE}> token specifies that the word-level compression rate should be around {TARGET_VALUE}. Values less than 1 indicate text reduction. Values greater than 1 indicate text expansion."
    }
\end{lstlisting}

\begin{figure}[!htbp]
\centering
    \caption{Formatted prompt examples.}
    \label{fig:prompt-templates}

    \begin{subfigure}[t]{0.49\textwidth}
        \caption{\texttt{Llama-3.2-1B-Instruct}}
        \begin{minipage}[t]{\linewidth}
        \begin{lstlisting}[basicstyle=\ttfamily\scriptsize, breaklines=true]
<|begin_of_text|><|start_header_id|>system<|end_header_id|>

Cutting Knowledge Date: December 2023
Today Date: 19 May 2025

You are a helpful expert in text simplification. You generate a simplified version of the text input by the user. You simplify the text according to the instructions given by the user. When asked to simplify a text, generate only the requested simplification, without any additional comments, notes or explanations.<|eot_id|><|start_header_id|>user<|end_header_id|>

INSTRUCTION: Simplify the following text such that its Flesch-Kincaid Grade Level (FKGL) score is approximately equal to that specified in the control token prepended to your generated simplification. The control token has the following format: <METRIC=VALUE>. The control token prepended to the source text indicates the FKGL value of the source text. 
SOURCE TEXT: <FKGL=4.8> No cure for the common cold exists , but the symptoms can be treated .
EXPLANATION: The <FKGL=4.0> token specifies that the target Flesch-Kincaid Grade Level should be approximately 4.0. Lower values indicate simpler text.<|eot_id|><|start_header_id|>assistant<|end_header_id|>

<FKGL=4.0><|eot_id|>
        \end{lstlisting}
        \end{minipage}
    \end{subfigure}
    \hfill
    \begin{subfigure}[t]{0.49\textwidth}
        \caption{\texttt{Llama-3.1-8B-Instruct}}
        \begin{minipage}[t]{\linewidth}
        \begin{lstlisting}[basicstyle=\ttfamily\scriptsize, breaklines=true]
<|begin_of_text|><|start_header_id|>system<|end_header_id|>

Cutting Knowledge Date: December 2023
Today Date: 19 May 2025

You are a helpful assistant. You are an expert in controlled text simplification. When you receive a text, you simplify it by rewriting it in a manner that is easier to read. Your simplification is guided by the simplification criteria specified by the user. You generate only the simplification result, without any additional comments or explanations.<|eot_id|><|start_header_id|>user<|end_header_id|>

INSTRUCTION: Simplify the following text such that its Flesch-Kincaid Grade Level (FKGL) score is approximately equal to that specified in the control token prepended to your generated simplification. The control token has the following format: <METRIC=VALUE>. The control token prepended to the source text indicates the FKGL value of the source text. 
SOURCE TEXT: <FKGL=4.8> No cure for the common cold exists , but the symptoms can be treated .
EXPLANATION: The <FKGL=4.0> token specifies that the target Flesch-Kincaid Grade Level should be approximately 4.0. Lower values indicate simpler text.<|eot_id|><|start_header_id|>assistant<|end_header_id|>

<FKGL=4.0><|eot_id|>
        \end{lstlisting}
        \end{minipage}
    \end{subfigure}


    \begin{subfigure}[t]{0.49\textwidth}
        \caption{\texttt{Mistral-3B-Instruct}}
        \begin{minipage}[t]{\linewidth}
        \begin{lstlisting}[basicstyle=\ttfamily\scriptsize, breaklines=true]
<s>system
You are a helpful expert in text simplification. You generate a simplified version of the text input by the user. You simplify the text according to the instructions given by the user. When asked to simplify a text, generate only the requested simplification, without any additional comments, notes or explanations.</s>
<s>user
INSTRUCTION: Simplify the following text such that its Flesch-Kincaid Grade Level (FKGL) score is approximately equal to that specified in the control token prepended to your generated simplification. The control token has the following format: <METRIC=VALUE>. The control token prepended to the source text indicates the FKGL value of the source text. 
SOURCE TEXT: <FKGL=4.8> No cure for the common cold exists , but the symptoms can be treated .
EXPLANATION: The <FKGL=4.0> token specifies that the target Flesch-Kincaid Grade Level should be approximately 4.0. Lower values indicate simpler text.</s>
<s>assistant
<FKGL=4.0> </s>
        \end{lstlisting}
        \end{minipage}
    \end{subfigure}
    \hfill
    \begin{subfigure}[t]{0.49\textwidth}
        \caption{\texttt{Mistral-7B-Instruct-v0.1}}
        \begin{minipage}[t]{\linewidth}
        \begin{lstlisting}[basicstyle=\ttfamily\scriptsize, breaklines=true]
<s> [INST] You are a helpful expert in text simplification. You generate a simplified version of the text input by the user. You simplify the text according to the instructions given by the user. When asked to simplify a text, generate only the requested simplification, without any additional comments, notes or explanations.

INSTRUCTION: Simplify the following text such that its Flesch-Kincaid Grade Level (FKGL) score is approximately equal to that specified in the control token prepended to your generated simplification. The control token has the following format: <METRIC=VALUE>. The control token prepended to the source text indicates the FKGL value of the source text. 
SOURCE TEXT: <FKGL=7.2> The common distance of the points of a circle from its center is called its radius .

EXPLANATION: The <FKGL=3.6> token specifies that the target Flesch-Kincaid Grade Level should be approximately 3.6. Lower values indicate simpler text. [/INST] <FKGL=3.6> </s>
        \end{lstlisting}
        \end{minipage}
    \end{subfigure}


    \begin{subfigure}[t]{0.9\textwidth}
        \caption{\texttt{Qwen3-1.7B} \& \texttt{Qwen3-8B}}
        \begin{minipage}[t]{\linewidth}
        \begin{lstlisting}[basicstyle=\ttfamily\scriptsize, breaklines=true]
<|im_start|>system
You are a helpful assistant. You are an expert in controlled text simplification. When you receive a text, you simplify it by rewriting it in a manner that is easier to read. Your simplification is guided by the simplification criteria specified by the user. You generate only the simplification result, without any additional comments or explanations.<|im_end|>
<|im_start|>user
INSTRUCTION: Simplify the following text such that its Dale-Chall Readability Score is approximately equal to that specified in the control token prepended to your generated simplification. The control token has the following format: <METRIC=VALUE>. The control token prepended to the source text indicates the Dale-Chall value of the source text. 
SOURCE TEXT: <DALE-CHALL=8.21> The boundaries have been drawn to take in as much as possible of the course of the brook and linking five distinct zones.
EXPLANATION: The <DALE-CHALL=8.21> token specifies that the target Dale-Chall readability score should be approximately 8.21. Lower values indicate simpler text.<|im_end|>
<|im_start|>assistant
<think>

</think>

<DALE-CHALL=8.21> <|im_end|>
        \end{lstlisting}
        \end{minipage}
    \end{subfigure}

\end{figure}

\clearpage
\section{Training Configuration}
\label{sec:appendix-training}

Fine-tuning and inference experiments were conducted on the University of Zurich Science Cluster, using NVIDIA A100 GPUs with 80\,GB memory. Each model was trained and evaluated on a single GPU, without distributed training. This setup was consistent across all model families and size to ensure comparable experimental conditions for both full and LoRA-based fine-tuning. A subset of inference runs was additionally conducted on the UBELIX cluster (University of Bern), using the same GPU configuration.

\begin{table}[!htbp]
\scriptsize
\centering\caption{Left: training hyperparameters selected for smaller ($\leq$4B) models, fine-tuned without PEFT. Right: training hyperparameters selected for larger ($>$4B) models, fine-tuned with PEFT.}
\label{tab:training-hyperparams}

\begin{tabular*}{0.7\linewidth}{@{\extracolsep{\fill}}lcc}
\toprule
 \textbf{Hyperparameters} & \multicolumn{2}{c}{\textbf{Shared Hyperparams}} \\
\midrule
batch size & \multicolumn{2}{c}{4} \\
grad. accumulation steps & \multicolumn{2}{c}{4} \\
cumulative batch size & \multicolumn{2}{c}{16} \\
weight decay & \multicolumn{2}{c}{0.01} \\
warmup steps & \multicolumn{2}{c}{30} \\
max epochs & \multicolumn{2}{c}{3} \\
scheduler & \multicolumn{2}{c}{cosine} \\
optimizer & \multicolumn{2}{c}{AdamW} \\
max length & \multicolumn{2}{c}{512 (Newsela: 4096)} \\
\midrule
\textbf{Hyperparameters} & \textbf{No PEFT} & \textbf{PEFT} \\
\midrule
learning rate & 5e-6 & 1e-4 \\
max grad norm & 0.5 & 1.0 \\
patience & 4 & 3 \\
LoRA rank & -- & 8 \\
LoRA alpha & -- & 16 \\
LoRA dropout & -- & 0.1 \\
\bottomrule
\end{tabular*}
\end{table}

\begin{table}[!htbp]
    \caption{Random seeds grouped by experiment, with results are aggregated across multiple.}
    \label{tab:seeds}
\centering
\scriptsize

\begin{tabular}{p{0.2\linewidth} p{0.6\linewidth}}
\toprule
\textbf{Experiment} & \textbf{Seeds} \\
      \midrule
        Strat. Partitioning & 2746317213, 478163327, 107420369, 3184935163, 1181241943, 1051802512, 958682846, 599310825, 3163119785, 440213415 \\
        \addlinespace[4pt]
        Downsampling & 69, 1, 40, 7, 29, 48, 78, 34, 67, 84 \\
        \addlinespace[4pt]
        Robust LLM Eval. & 37, 15, 96, 2, 28 \\
        \bottomrule
    \end{tabular}

\end{table}

\begin{table}[!htbp]
  \caption{Hyperparameter optimization on Weights\&Biases platform. Left: sweep configuration for smaller ($\leq$4B) models, finetuned without PEFT. Right: sweep configuration for larger ($>$4B) models, finetuned with PEFT (LoRA).}
  \label{table:sweeps}
\centering
\small
\begin{tabular}{p{0.25\linewidth} p{0.25\linewidth}}
\toprule
\multicolumn{1}{c}{\textbf{No PEFT}} & \multicolumn{1}{c}{\textbf{PEFT}} \\
\hline
\begin{minipage}[t]{\linewidth}
\vspace{1mm}
\centering
\begin{lstlisting}[basicstyle=\ttfamily\scriptsize, breaklines=true]
method: grid
metric:
  goal: minimize
  name: eval_loss
parameters:
  learning_rate:
    values:
      - 1e-06
      - 5e-06
      - 1e-05
      - 5e-05
  max_grad_norm:
    values:
      - 0.5
      - 1
      - 2
  weight_decay:
    values:
      - 0
      - 0.01
      - 0.1
\end{lstlisting}
\vspace{1mm}
\end{minipage}
&
\begin{minipage}[t]{\linewidth}
\vspace{1mm}
\centering
\begin{lstlisting}[basicstyle=\ttfamily\scriptsize, breaklines=true]
method: grid
metric:
  goal: minimize
  name: eval_loss
parameters:
  learning_rate:
    values:
      - 1e-05
      - 5e-05
      - 0.0001
      - 0.0002
  lora_dropout:
    values:
      - 0.05
      - 0.1
  lora_r:
    values:
      - 4
      - 8
      - 16
  max_grad_norm:
    values:
      - 0.5
      - 1 
\end{lstlisting}
\vspace{1mm}
\end{minipage}
\\
\bottomrule
\end{tabular}

\end{table}




\end{document}